%% file: main.tex
\documentclass[letterpaper, 10pt, conference]{ieeeconf}  

\usepackage{svg}
\usepackage{amsmath, amssymb}
\usepackage{mathtools}
\usepackage{enumerate}

\usepackage[hidelinks]{hyperref}
\usepackage{cleveref}
\usepackage[nobiblatex]{xurl}

\usepackage{multirow}
\usepackage{tabularx,booktabs}
\newcolumntype{C}{>{\centering\arraybackslash}X} 
\newcolumntype{R}{>{\raggedleft\arraybackslash}X} 
\newcolumntype{L}{>{\raggedright\arraybackslash}X} 

\usepackage{blindtext}
\usepackage{xcolor}
\usepackage{tikz}
\usetikzlibrary{shapes,
	arrows,
	positioning,
	fit,
	backgrounds,
	calc,
	patterns,
	math,
	external}
\usepackage{pgfplots}
\usetikzlibrary{pgfplots.statistics, pgfplots.colorbrewer, calc, external} 
\pgfplotsset{compat=1.18}

\input{Data/Colors}
\newcommand{\norm}[1]{\left\lVert#1\right\rVert}

\newcommand\copyrighttext{%
    \footnotesize \textcopyright 2025 IEEE.  Personal use of this material is permitted.  Permission from IEEE must be obtained for all other uses, in any current or future media, including reprinting/republishing this material for advertising or promotional purposes, creating new collective works, for resale or redistribution to servers or lists, or reuse of any copyrighted component of this work in other works.
}

\newcommand\copyrightnotice{%
    \begin{tikzpicture}[remember picture,overlay]
        \node[anchor=south,yshift=10pt] at (current page.south) {\fbox{\parbox{\dimexpr\textwidth-\fboxsep-\fboxrule\relax}{\copyrighttext}}};
    \end{tikzpicture}%
}

\IEEEoverridecommandlockouts                              
\title{\LARGE \bf
	Comparison of Localization Algorithms between Reduced-Scale and Real-Sized Vehicles Using Visual and Inertial Sensors
}

\author{Tobias Kern$^{1}$, Leon Tolksdorf$^{1, 2}$, Christian Birkner$^{1}$ 
	\thanks{$^{1}$CARISSMA Institute of Safety in Future Mobility, Technische Hochschule Ingolstadt, Ingolstadt, Germany, e-mail:
		{\tt\small \{tobiasbenjamin.kern, leon.tolksdorf, christian.birkner\}@thi.de}}%
	\thanks{$^{2}$Department of Dynamics and Control, Eindhoven University of Technology, Eindhoven, The Netherlands, e-mail:
		{\tt\small \{l.t.tolksdorf\}@tue.nl}}%
}

\begin{document}
	
	\maketitle
	
	\thispagestyle{empty}
	\pagestyle{empty}
	\copyrightnotice
	\begin{abstract}               
        Physically reduced-scale vehicles are emerging to accelerate the development of advanced automated driving functions. In this paper, we investigate the effects of scaling on self-localization accuracy with visual and visual-inertial algorithms using cameras and an inertial measurement unit (IMU). For this purpose, ROS2-compatible visual and visual-inertial algorithms are selected, and datasets are chosen as a baseline for real-sized vehicles. A test drive is conducted to record data of reduced-scale vehicles. We compare the selected localization algorithms, OpenVINS, VINS-Fusion, and RTAB-Map, in terms of their pose accuracy against the ground-truth and against data from real-sized vehicles. 
        When comparing the implementation of the selected localization algorithms to real-sized vehicles, OpenVINS has the lowest average localization error. Although all selected localization algorithms have overlapping error ranges, OpenVINS also performs best when applied to a reduced-scale vehicle.
        When reduced-scale vehicles were compared to real-sized vehicles, minor differences were found in translational vehicle motion estimation accuracy. However, no significant differences were found when comparing the estimation accuracy of rotational vehicle motion, allowing RSVRs to be used as testing platforms for self-localization algorithms.
	\end{abstract}
	
	
	\section{Introduction} \label{sec:introduction}
	
    Although the development of automated vehicles is ongoing, extensive testing is needed to ensure the safety and reliability of each deployed algorithm as well as the functionality of the entire vehicle. A challenge in testing with real-sized prototype vehicles is that those tests are typically expensive due to the need for proving grounds, dedicated testing hardware, and adherence to safety protocols. Therefore, testing on reduced-scale vehicles or reduced-scale vehicular robots (RSVRs) has emerged as a promising trend. The advantages of this approach are varied: reduced costs, increased accessibility, and enhanced safety due to reduced velocities and weights in comparison to their real-sized counterparts \cite{Park2020springer, Lovato2023springer, 10298759, 10821589, 9444783}. These RSVRs provide an efficient platform for testing and refining perception, localization, and planning algorithms under controlled conditions. For instance, \cite{10298759} and \cite{10821589} explored their use for vehicle dynamic controller development by finding the dynamical similarity of RSVR with their real-sized counterparts. \\
    Since the development of localization systems is particularly crucial for autonomous navigation \cite{8025618}, the impact of scaling on localization accuracy is of significant interest. \\
    Using simultaneous localization and mapping algorithms enables precise positioning, with odometry estimation as one of its core functionalities \cite{8025618}. Different types of sensors can be used to estimate odometry. The simplest is wheel odometry, which uses the tire radius and wheel speed to calculate the vehicle's velocity, from which the position can be estimated. Other typical sensors used for odometry estimation are GNSS, Lidar, camera, and inertial measurement unit (IMU). The latter two are used for visual odometry (VO) and visual-inertial odometry (VIO), respectively. 
    We group the VO and VIO algorithms into three classes:
    \begin{enumerate}
        \item \textit{Filter-based methods:} Usage of probabilistic models, such as Kalman filters or particle filters, to estimate the state of the system over time. For example, OpenVINS \cite{Geneva2020ICRA}, ROVIO \cite{Bloesch2017IteratedEK}, and  MSCKF \cite{4209642},
        \item \textit{Deep learning-based methods:} Leveraging of deep learning techniques like convolutional neural networks and recurrent neural networks for feature extraction and sequence modeling. For example, DVSO \cite{yang2018dvso},  DeepVO \cite{7989236}, and DeepAVO \cite{ZHU202222},
        \item \textit{Optimization-based methods:} Formulating localization as an optimization problem and solving it during runtime. For example, VINS-FUSION \cite{qin2017vins,qin2018online, qin2019a, qin2019b}, Kimera \cite{Rosinol20icra-Kimera}, ORB-SLAM3 \cite{9440682}, Basalt \cite{8938825}.
    \end{enumerate}
    VO and VIO algorithms are typically tested on datasets that capture diverse scenarios and environmental conditions to evaluate the accuracy of self-localization. Notable examples include Kitti Odometry by KIT \cite{Geiger2012CVPR}, Complex Urban by KAIST \cite{jjeong-2019-ijrr}, and EuRoC MAV by ETH Zürich \cite{Burri25012016}. Each dataset offers unique features:
    \begin{enumerate}
        \item \textit{Kitti Odometry:} offers stereo camera images, $360\text{\,°}$-lidar point clouds, and GNSS data, recorded in Germany, in rural, and urban environments.
	    \item \textit{Complex Urban:} offers stereo camera images, $360\text{\,°}$-lidar point clouds from twisted 3D- and 2D-lidars, IMU, GNSS, and wheel odometry data, recorded in Korean urban, suburban and highway environments. 
	    \item \textit{EuRoC MAV:} offers images of a global shutter stereo camera and IMU data, recorded indoors with a micro aerial vehicle (MAV).
    \end{enumerate}
    With the above in mind, we ask whether VO and VIO algorithms perform similarly in estimation accuracy when applied to RSVRs compared to real-sized vehicles. Therefore, this paper contributes:
    \begin{enumerate}
        \item An experimental comparison of the self-localization accuracy of the selected VO and VIO algorithms, VINS-Fusion, OpenVINS, and RTAB-Map, with each algorithm being tested on an RSVR. Precisely, we evaluate the localization accuracy of each algorithm with respect to the ground-truth pose. 
        \item A comparison of the selected VO and VIO algorithms on open-source datasets for real-sized vehicles. Here, we evaluate the self-localization error of selected VO and VIO algorithms with respect to the corresponding dataset's ground-truth data. 
        \item An analysis of the scaling effects on self-localization accuracy of the selected VO and VIO algorithms. This analysis provides insights into the impact of the scaling factor and sensor position on the accuracy of odometry estimation.
    \end{enumerate}
    The remainder of this article is structured as follows. Section \ref{sec:problem_statement} details the problem statement. The selection criteria, tests, and evaluation metrics of the VO or VIO algorithms are given in Section \ref{sec:methodology}. The results are presented in Section \ref{sec:results} and discussed in Section \ref{sec:discussion}. Lastly, we conclude this article in Section \ref{sec:conclusion}.

	\section{Problem Statement} \label{sec:problem_statement}
\begin{figure}
		\centering
		\includesvg[width=\linewidth]{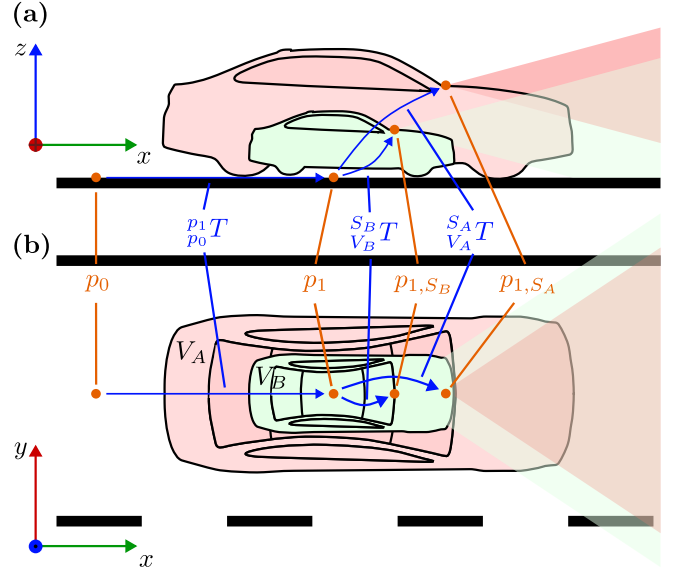}
		\caption{Visualization of the problem statement. \textbf{(a)} side view to visualize different $z$-positions of the sensor systems \textbf{(b)} top-down view to visualize same $x$-$y$-position.}
		\label{fig:ProbStatement}
	\end{figure}

	Let any pose $p$ of a vehicle be characterized by the position $(x, y, z) \in \mathbb{R}^3$ and orientation $(\phi, \psi, \theta) \in [0, 2\pi)^3$, defined at the vehicle's geometric center, where the roll-angle $\phi$, the pitch-angle $\psi$, and the yaw-angle $\theta$ are identified around each respective coordinate axis, see Figure \ref{fig:ProbStatement}. Consider a real-sized vehicle, denoted $V_A$, of which the pose $p_{k,A}$ is measured at discrete-time instance $k$. Vehicle $V_A$ is equipped with a visual-inertial sensor system $S_A$, i.e., camera, accelerometer, and gyroscope, for odometry calculation. However, the pose of the sensor system $p_{k, S_A}$ is offset from the pose of the vehicle $p_{k,A}$, as it is not mounted in the geometric center of the vehicle. Let the translation vector $X \in \mathbb{R}^{3\times1}$ and rotation matrix $R \in \mathbb{SO}(3)$ constitute a transformation matrix $T \in \mathbb{SE}(3)$ between any two poses, with $\mathbb{SO}(3)$ denoting the Lie group of all rotations and $\mathbb{SE}(3)$ the Lie group of all rigid transformations in the three-dimensional space. For instance, let $\prescript{p_1}{p_0}{T}$ consisting of the translation $\prescript{p_1}{p_0}{X}$ and Rotation $\prescript{p_1}{p_0}{R}$, define a transformation from $p_0$ to $p_1$, such that
	\begin{equation}
		p_1 = \prescript{p_1}{p_0}{T}\,p_0.
	\end{equation}
	As vehicle $V_A$ moves from $p_{0,A}=p_0$ at time-instance $k=0$ and arrives at $p_1$ by time-instance $k=1$, the sensor system $S_A$ reports the measured pose $p_{1,A}$.\\
    Consider an RSVR $V_B$, a reduced-scale version of $V_A$, equipped with a similar sensor system $S_B$. The pose of the sensor system $S_B$ has been scaled accordingly. Therefore, $S_B$ is measuring the pose $p_{k,B}$ of RSVR $V_B$ at time-instance $k$. 
	As RSVR $V_B$ moves from $p_{2,B}=p_0$ at time-instance $k=2$ and, driving the same path as $V_A$, arrives at $p_1$ by time-instance $k=3$, the sensor system $S_B$ reports the measured pose $p_{3,B}$.
	Let both sensor systems $S_A$ and $S_B$ be distinctly offset, such that
	\begin{equation}
		\prescript{p_{k,S_A}}{p_{k,A}}{T} \neq \prescript{p_{k,S_B}}{p_{k,B}}{T},
	\end{equation}
	and, therefore, the point of view of the visual sensor differs for each sensor system, as do the inertial reference point of the respective accelerometer and gyroscope.
    \paragraph*{Error definition} 
        While both sensor systems, $S_A$ and $S_B$, report a different pose, as they are displaced from each other due to the scaling, they follow the same path. Hence, the ground-truth change in pose is identical. In a real-world application, however, due to the different poses of the sensor systems, different measurement data is passed on to the VO or VIO algorithms, causing errors in odometry estimation. 
        Thus, the measured change in pose may differ from the ground-truth. 
        To determine the odometry estimation error, the metrics proposed by the evo tool \cite{grupp2017evo} are used, that is:
        \begin{subequations}
        \begin{alignat}{4}
             &APE_\mathrm{full}  &=& \sqrt{\frac{1}{k_f}\sum_{k=0}^{k_f}\norm{\prescript{p_k}{p_0}{T}^{-1}\,\prescript{p_{k,A}}{p_{0,A}}{T} - I}_F^2}\text{\ ,}       \label{eq:APE_full} \\
             &APE_\mathrm{trans} &=& \sqrt{\frac{1}{k_f}\sum_{k=0}^{k_f}\norm{\prescript{p_k}{p_0}{X} - \prescript{p_{k,A}}{p_{0,A}}{X}}_F^2}\text{\ ,}              \label{eq:APE_trans}\\
             &APE_\mathrm{rot}   &=& \sqrt{\frac{1}{k_f}\sum_{k=0}^{k_f}\norm{\prescript{p_k}{p_0}{R}^{-1}\,\prescript{p_{k,A}}{p_{0,A}}{R} - I}_F^2}\text{\ ,}       \label{eq:APE_rot}  \\
             &RPE_\mathrm{full}  &=& \sqrt{\frac{1}{k_f}\sum_{k=1}^{k_f}\norm{\prescript{p_{k}}{p_{k-1}}{T}^{-1}\,\prescript{p_{k,A}}{p_{k-1,A}}{T} - I}_F^2}\text{\ ,} \label{eq:RPE_full} \\
             &RPE_\mathrm{trans} &=& \sqrt{\frac{1}{k_f}\sum_{k=1}^{k_f}\norm{\prescript{p_{k}}{p_{k-1}}{X} - \prescript{p_{k,A}}{p_{k-1,A}}{X}}_F^2}\text{\ ,}         \label{eq:RPE_trans}\\
             &RPE_\mathrm{rot}   &=& \sqrt{\frac{1}{k_f}\sum_{k=1}^{k_f}\norm{\prescript{p_{k}}{p_{k-1}}{R}^{-1}\,\prescript{p_{k,A}}{p_{k-1,A}}{R} - I}_F^2}\text{\ ,} \label{eq:RPE_rot}
        \end{alignat}
        \end{subequations}
        where $k_f$ describes the final discrete-time instance of the measurement, $I$ symbolizes the unit matrix, and $\norm{A}_F$ denotes the Frobenius norm of matrix $A$. Note that for the specific vehicle $V_A$ and $V_B$, the respective poses are substituted in (\ref{eq:APE_full})--(\ref{eq:RPE_rot}), giving the errors for both vehicles. 
        The APE is important to observe drift. Further, RPE is used to assess the accuracy of relative movement between two consecutive points in time. It is worth mentioning that the RPE is not normalized over time. 
            
    \paragraph*{Objective} 
        Experimentally determine (\ref{eq:APE_full})--(\ref{eq:RPE_rot}) for vehicles $V_A$ and $V_B$, respectively.\\
        To approach this objective, first, we record a measurement sequence with $k_f$ samples on the RSVR $V_B$. Second, we use the vehicles out of the datasets Kitti Odometry, and Complex Urban as the real-sized vehicle $V_A$. Third, we process the recorded and retrieved data with the odometry estimation algorithms to, finally, evaluate the odometry estimation error metrics (\ref{eq:APE_full})--(\ref{eq:RPE_rot}).
        
	\section{Methodology} \label{sec:methodology}
	We select the VO and VIO algorithms OpenVINS \cite{Geneva2020ICRA}, VINS-Fusion \cite{qin2017vins,qin2018online,qin2019a,qin2019b}, and RTAB-Map \cite{Labb__2018} due to their widespread usage and ROS2 \cite{doi:10.1126/scirobotics.abm6074} compatibility, as the RSVR utilizes a ROS2 interface, allowing for swift interoperability.
	
    \subsection{Experimental Estimation of Self-localization Accuracy}
\begin{figure}
		\centering
		\includesvg[width=\linewidth]{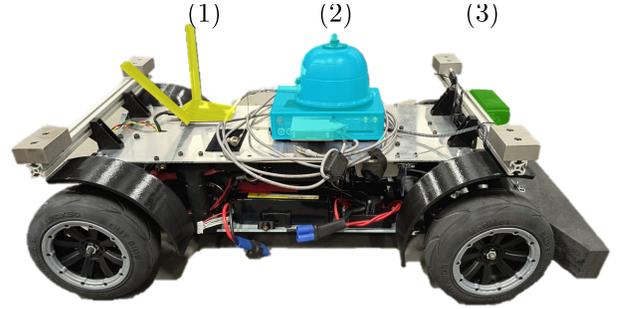}
		\caption{RSVR used for recording the experimental data. $(1)$ (yellow): Wifi-adapter; $(2)$ (cyan): IPS beacon; $(3)$ (green): camera.}
		\label{fig:meas_setup}
	\end{figure}

	For testing with the RSVR, shown in Figure \ref{fig:meas_setup}, we use the RSVR of \cite{Summerer2024} with a scaling factor of $1\!:\!5$, representing a Tesla Model S. The RSVR is remotely driven at the CARISSMA \cite{carissma} indoor facility on an arbitrary path with a velocity of maximum $1.7\text{\,ms}^{-1}$ representing slow urban traffic with a scaled up velocity of $30\text{\,kmh}^{-1}$. To collect ground-truth position, orientation, velocity, and angular rate, the CARISSMA indoor facility is equipped with the indoor positioning system (IPS) of Racelogic \cite{vips}. This IPS has a static positional accuracy of $\!<\!0.1\text{\,m}$ and a velocity accuracy of $\!<\!0.1\text{\,kmh}^{-1}$. During the test drive, the RSVR collects data from a global shutter stereo camera, the Intel Realsense D435i, at $30\text{\,Hz}$ with an image resolution of $848\!\times\!480$ pixels. Inertial data is collected at $200\text{\,Hz}$ by the IMU, which is integrated into the D435i camera. The camera is positioned in the front, and the beacon for the IPS reference position is located above the center of gravity, see Figure \ref{fig:meas_setup}. 
    
    \noindent For each VO or VIO algorithm, the recorded data is replayed on the main computer of the RSVR, an Nvidia Jetson Orin NX $16\text{\,GB}$, while the respective algorithm is run, and the pose estimates are saved. The base parameters used for OpenVINS are given by their "rs\_t265" configuration\footnote{\url{https://github.com/rpng/open_vins/tree/3ae57920799b4d997584c442ad1448869db5659f/config/rs_t265}; Accessed: 2025-04-02}, for VINS-Fusion given by their "realsense\_d435i" configuration\footnote{\url{https://github.com/zinuok/VINS-Fusion-ROS2/tree/main/config/realsense_d435i}; Accessed: 2025-04-02}, and for RTAB-Map given by their configuration "realsense\_d435i\_stereo.launch.py"\footnote{\url{https://github.com/introlab/rtabmap_ros/blob/iron-devel/rtabmap_examples/launch/realsense_d435i_stereo.launch.py}; Accessed: 2025-04-02}. The parameters we adjusted are listed in Table \ref{tab:rsvr_odom_params}. 
    Finally, the pose estimates of each algorithm are evaluated with (\ref{eq:APE_full})--(\ref{eq:RPE_rot}).
    $APE_\mathrm{full}$ (\ref{eq:APE_full}) and $RPE_\mathrm{full}$ (\ref{eq:RPE_full}) are calculated based on the transformation matrices $\prescript{p_{k,B}}{p_{0,B}}{T}$ and $\prescript{p_{k,B}}{p_{k-1,B}}{T}$ with respect to the ground-truth's transformation matrices $\prescript{p_{k}}{p_{0}}{T}$ and $\prescript{p_{k}}{p_{k-1}}{T}$.
    $APE_\mathrm{trans}$ (\ref{eq:APE_trans}) and $RPE_\mathrm{trans}$ (\ref{eq:RPE_trans}) are calculated with the translation vectors $\prescript{p_{k,B}}{p_{0,B}}{X}$ and $\prescript{p_{k,B}}{p_{k-1,B}}{X}$ in relation to the ground-truth's translation vectors $\prescript{p_{k}}{p_{0}}{X}$ and $\prescript{p_{k}}{p_{k-1}}{X}$. 
    $APE_\mathrm{rot}$ (\ref{eq:APE_rot}) and $RPE_\mathrm{rot}$ (\ref{eq:RPE_rot}) are calculated with the rotation matrices $\prescript{p_{k,B}}{p_{0,B}}{R}$ and $\prescript{p_{k,B}}{p_{k-1,B}}{R}$ in combination with the ground-truth's rotation matrices $\prescript{p_{k}}{p_{0}}{R}$ and $\prescript{p_{k}}{p_{k-1}}{R}$.

    \begin{table}
        \centering
        \caption{Parameters adjustments of self-localization algorithms on the RSVR.}
        \label{tab:rsvr_odom_params}
        \begin{tabularx}{\linewidth}{@{}L | R}
            \toprule
            \multicolumn{2}{c}{OpenVINS} \\
            \toprule
            \texttt{calib\_cam\_extrinsics} &  \texttt{false} \\
            \texttt{calib\_cam\_intrinsics} &  \texttt{false} \\
            \texttt{dt\_slam\_delay} & $0$ \\
            \texttt{init\_window\_time} & $1.5$ \\
            \texttt{init\_imu\_thresh} & $0.7$ \\
            \texttt{init\_max\_disparity} & $20.0$ \\
            \texttt{init\_max\_features} & $100$ \\
            \texttt{grid\_x} & $16$ \\
            \texttt{grid\_y} & $16$ \\
            \texttt{track\_frequency} & $20.0$ \\
            \texttt{histogram\_method} & "NONE" \\
            \texttt{use\_mask} & \texttt{false} \\
            \toprule
            \multicolumn{2}{c}{VINS-Fusion} \\
            \toprule
            \texttt{image\_width} &  $848$ \\
            \bottomrule
        \end{tabularx}
    \end{table}

    \subsection{Self-localization Accuracy on Datasets}
    
    To obtain a baseline for the APE and RPE estimates of the RSVR, we also evaluate VINS-Fusion, OpenVINS, and RTAB-Map on the EuRoC-MAV, KITTI Odometry, and Complex Urban datasets. All three VO and VIO algorithms are also processed on a Jetson Orin NX with $16\text{\,GB}$ RAM to exclude the influence of different processing power and compiler optimizations. All three datasets are loaded and replayed on this computer with a set playback speed multiplier of $1.0$ to simulate real-time behavior. Each VO or VIO algorithm runs separately on each dataset to prevent overlapping ROS2-topics and CPU overload. The resulting APE and RPE are only calculated based on succeeded sequences. We define a succeeded sequence as: the standard deviation of the algorithm's velocity estimation is not exceeding $20\text{\,ms}^{-1}$, key points could be tracked consecutively during the sequence, and the algorithm did not fail to initialize. Such reflects real-world usage where, typically, if the estimation fails, appropriate countermeasures are taken, e.g., fall back to redundant algorithm layers. 
    
    \subsection{EuRoC-MAV}
    All three VO and VIO algorithms have an example configuration for the EuRoC-MAV dataset\footnote{\url{https://github.com/introlab/rtabmap_ros/blob/c0e4927f8839d4d9d0b8b8379c6eb001d39b0bd7/rtabmap_examples/launch/euroc_datasets.launch.py}; Accessed: 2025-04-02}\textsuperscript{,}\footnote{\url{https://github.com/HKUST-Aerial-Robotics/VINS-Fusion/tree/1c2949048f79d89abcb836e826a24fe2ff99e2e4/config/euroc}; Accessed: 2025-04-02}\textsuperscript{,}\footnote{\url{https://github.com/rpng/open_vins/tree/3ae57920799b4d997584c442ad1448869db5659f/config/euroc_mav}; Accessed: 2025-04-02}. The results of OpenVINS with the given configuration, as stated in \cite{Geneva2020ICRA}, are not reproducible with the Jetson Orin NX. To compensate for short acceleration pulses, the time window to initialize the algorithm (\texttt{init\_window\_time}) is reduced from $2.0$ to $1.0$, as well as the maximum stereo disparity allowed for initialization (\texttt{init\_max\_disparity}) is relaxed from $10.0$ to $15.0$. Additionally, the zero velocity update (\texttt{try\_zupt}) is enabled.

    \subsection{Kitti Odometry}
    OpenVins was not tested for the Kitti Odometry dataset, as it is a VIO algorithm, and the Kitti Odometry dataset has no inertial measurement data available.
    For VINS-Fusion, the available example configuration\footnote{\url{https://github.com/HKUST-Aerial-Robotics/VINS-Fusion/tree/b00c8b34e1e5d998b456bdc8746f554c1e4fe717/config/kitti_odom}; Accessed: 2025-04-02} was used. This configuration is adopted for RTAB-Map and modified as described in \cite{Labb__2018}. The parameters for RTAB-Map are listed in Table \ref{tab:rtab_kitti_params}. 

    \begin{table}
        \centering
        \caption{Parameters used to evaluate the RTAB-Map algorithm on the Kitti Odometry dataset. Boolean values are written in typewriter font. Numerical values are written as numbers and string or text-type values are written in between quotation marks.}
        \label{tab:rtab_kitti_params}
        \begin{tabularx}{\linewidth}{ @{}l | R}
            \toprule
            Parameter & Value \\
            \midrule
            \texttt{subscribe\_stereo} & \texttt{True} \\
            \texttt{subscribe\_odom\_info} & \texttt{True}  \\
            \texttt{approx\_sync} & \texttt{False} \\
            \texttt{sync\_queue\_size} & $100$ \\
            \texttt{RGBD/CreateOccupancyGrid} & "false" \\
            \texttt{Rtabmap/ImageBufferSize} & "0" \\
            \texttt{Odom/ImageBufferSize} & "0" \\
            \texttt{Rtabmap/DetectionRate} & "2.0" \\
            \texttt{Kp/RoiRatios} & "0.001 0.0 0.0 0.0" \\
            \texttt{Kp/MaxDepth} & "30.0" \\
            \texttt{RGBD/LinearUpdate} & "0" \\
            \texttt{GFTT/QualityLevel} & "0.01" \\
            \texttt{GFTT/MinDistance} & "7" \\
            \texttt{OdomF2M/MaxSize} & "3000" \\
            \texttt{Mem/STMSize} & "30" \\
            \texttt{Kp/MaxFeatures} & "750" \\
            \texttt{Vis/MaxFeatures} & "1500" \\
            \texttt{Rtabmap/CreateIntermediateNodes} & "true" \\
            \bottomrule
        \end{tabularx}
    \end{table}

    \subsection{Complex Urban}
    The Complex Urban dataset was tested on all three VO and VIO algorithms. The dataset provides calibration files for the cameras and the IMU. These are used to construct the needed camera calibration for RTAB-Map. All other parameters of RTAB-Map are set according to the values with EuRoC-MAV. RPNG provides a configuration file for OpenVINS\footnote{\url{https://github.com/rpng/open_vins/tree/3ae57920799b4d997584c442ad1448869db5659f/config/kaist}; Accessed: 2025-04-02}, which is used to evaluate OpenVINS and as a base configuration for VINS-Fusion.

	\section{Results} \label{sec:results}

    \subsection{Experimental Self-localization Accuracy}
\begin{figure}
        \textbf{(a)}\\
		\input{figures/Results_OdomPaths.tikz}
        \textbf{(b)}\\
        \input{figures/Result_v_omega.tikz}
		\caption{\textbf{(a)} The tracked visual and visual-inertial odometry trajectories of the RSVR. \textbf{(b)} ground-truth velocity and yaw rate of the RSVR during experimental measurements.}
		\label{fig:odom paths}
	\end{figure}
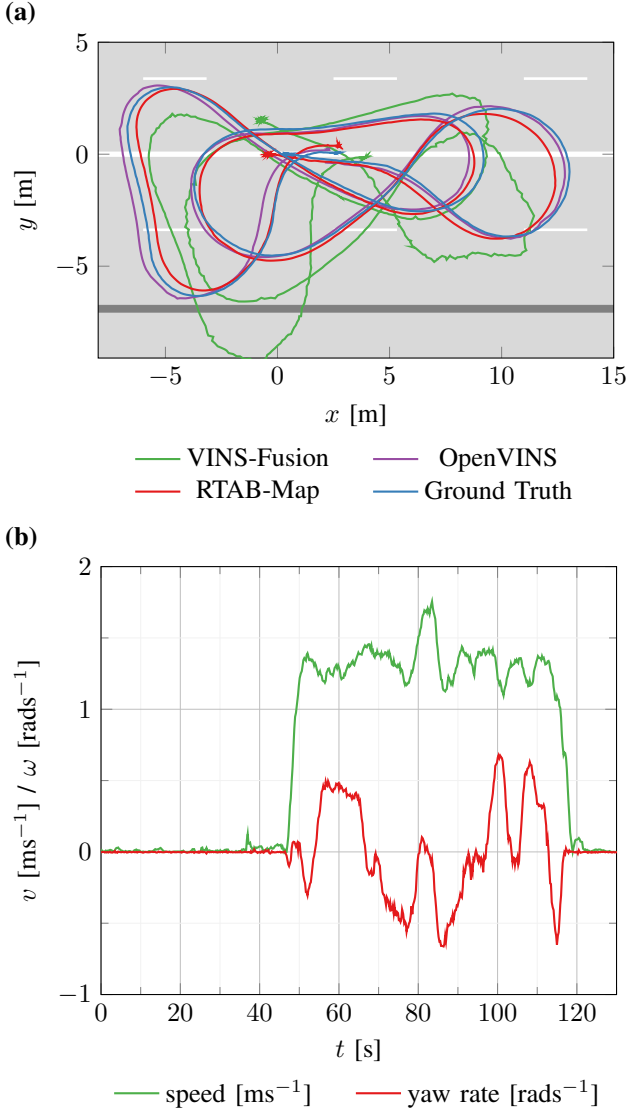

    Figure \ref{fig:odom paths}~(a) shows the recorded ground-truth trajectory (dark blue), driven with the RSVR, and the trajectories, i.e., collections of poses, estimated by the different VO and VIO algorithms. We aligned the estimated trajectories to the ground-truth trajectory using the evo tool~\cite{grupp2017evo}. The estimated paths were aligned to the ground-truth to eliminate the initial offset in translation and rotation between each algorithm's local coordinate system and the ground-truth's ENU coordinate system and to minimize the $APE_\mathrm{full}$.
    Figure \ref{fig:odom paths}~(b) shows the recorded ground-truth speed and yaw rate. 
    The metrics (\ref{eq:APE_full})--(\ref{eq:RPE_rot}) are used to evaluate the recorded trajectory of the RSVR and are given in Table \ref{tab:APE_RPE_reduced_scale}. The best-performing values are written in bold font. The background of Figure \ref{fig:odom paths}~(a) shows the indoor testing area of CARISSMA \cite{carissma} schematically. The dark grey area represents concrete road barriers. As OpenVINS does not publish odometry information until motion is detected and its internal filtering algorithm is initialized, an offset from the origin can be observed in Figure \ref{fig:odom paths}~(a). 
    Also, VINS-Fusion's optimizer failed to converge during the first turns in our self-recorded indoor track. 

	\begin{table}
		\centering
		\caption{Absolute and relative pose error of the tested algorithms on the reduced-scale vehicular robot. The best value is marked in bold font.}
        \label{tab:APE_RPE_reduced_scale}
        \begin{tabularx}{\linewidth}{@{} l | *{1}R | *{1}R | *{1}R}
			\toprule
            \multicolumn{4}{c}{ Absolute Pose Error } \\
            \toprule
			Algorithm    & \multicolumn{1}{c|}{$APE_\mathrm{full}$ [-]} & \multicolumn{1}{c|}{$APE_\mathrm{trans}$ [m]} & \multicolumn{1}{c|}{$APE_\mathrm{rot}$ [-]} \\
            \midrule
			OpenVINS     & \textbf{0.491}           & \textbf{0.374}                 &          0.317                  \\
			VINS-Fusion  &          3.001           &          2.960                 &          0.530                  \\
			RTAB-Map     &          0.574           &          0.496                 & \textbf{0.290}         \\
			\toprule
            \multicolumn{4}{c}{Relative Pose Error } \\
            \toprule
            Algorithm    & \multicolumn{1}{c|}{$RPE_\mathrm{full}$ [-]} & \multicolumn{1}{c|}{$RPE_\mathrm{trans}$ [m]} & \multicolumn{1}{c|}{$RPE_\mathrm{rot}$ [-]}  \\
            \midrule
            OpenVINS     & \textbf{0.005}          &            0.003                &   \textbf{0.004}    \\
			VINS-Fusion  &         0.010           &            0.004                &            0.009    \\
			RTAB-Map     &         0.009           &   \textbf{0.003}                &            0.009    \\
            \bottomrule
		\end{tabularx}
	\end{table}
    
    \subsection{Self-localization Accuracy on Datasets}
    Table \ref{tab:Dataset_sequences} lists the succeeded sequences for each VO and VIO algorithm and dataset.
    OpenVINS failed on MH\_02, MH\_04, and V2\_02 sequences due to bad initialization. On V1\_03 and V2\_03, OpenVINS lost track because of motion blur. RTAB-Map did not initialize in V1\_01, V1\_02. On MH\_05 and V2\_03, RTAB-Map lost track due to motion blur caused by the fast motion of these sequences. In Complex Urbans's sequence 32, RTAB-Map and VINS-Fusion failed to initialize due to the initial movement in this sequence.

     \begin{table*}[h]
        \centering
        \caption{Succeeded sequences of the datasets by odometry algorithm. Succeeded odometry estimations are marked by \checkmark. Failed estimations are marked with X. nc marks a not conducted test due to incompatibility with the dataset.}
        \label{tab:Dataset_sequences}
        \begin{tabularx}{\textwidth}{ @{} l| *{11}C}
            \toprule
            \multicolumn{12}{c}{EuRoC-MAV sequences} \\
            \toprule
            Algorithm & MH\_01 & MH\_02 & MH\_03 & MH\_04 & MH\_05 & V1\_01 & V1\_02 & V1\_03 & V2\_01 & V2\_02 & V2\_03 \\
            \midrule
            VINS-Fusion & \checkmark & \checkmark & \checkmark & \checkmark & \checkmark & \checkmark & \checkmark & \checkmark & \checkmark & \checkmark & \checkmark \\  
            OpenVINS    & \checkmark & X & \checkmark & X & \checkmark & \checkmark & \checkmark & X & \checkmark & X & X \\ 
            RTAB-Map    & \checkmark & \checkmark & \checkmark & \checkmark & X & X & X & \checkmark & \checkmark & \checkmark & X \\ 
            \midrule
            \multicolumn{12}{c}{Kitti Odometry sequences} \\
            \toprule
            Algorithm   & 00 & 01 & 02 & 03 & 04 & 05 & 06 & 07 & 08 & 09 & 10 \\
            \midrule
            VINS-Fusion & \checkmark  & \checkmark  & \checkmark  & \checkmark  & \checkmark  & \checkmark & \checkmark  & \checkmark  & \checkmark  & \checkmark  & \checkmark  \\  
            OpenVINS    & nc & nc & nc & nc & nc & nc & nc & nc & nc & nc & nc \\ 
            RTAB-Map    & \checkmark  & \checkmark  & \checkmark  & \checkmark  & \checkmark  & \checkmark  & \checkmark  & \checkmark  & \checkmark  & \checkmark  & \checkmark  \\ 
            \midrule
             \multicolumn{12}{c}{Complex Urban sequences} \\
            \toprule
            Algorithm   & 28 & 32 & 38 & 39 & & & & & & & \\
            \midrule
            VINS-Fusion &  \checkmark & X & \checkmark & \checkmark & & & & & & & \\  
            OpenVINS    & \checkmark  & \checkmark & \checkmark & \checkmark & & & & & & & \\ 
            RTAB-Map    & \checkmark  & X & \checkmark & \checkmark & & & & & & & \\ 
            \bottomrule
        \end{tabularx}
    \end{table*}
    
    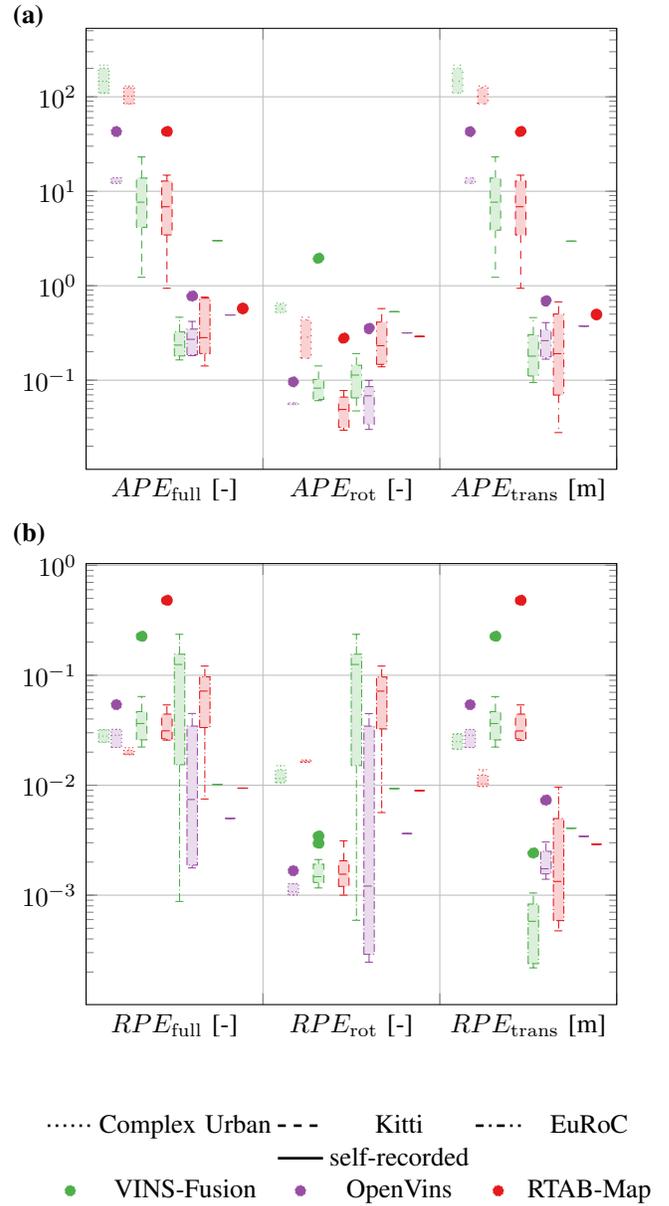
\begin{figure}
        \input{figures/new_figures.tex}
        \caption{Spread of absolute \textbf{(a)} and relative pose error \textbf{(b)} on each dataset and algorithm.}
        \label{fig:Boxiplotti}
    \end{figure}
    
    \noindent Table \ref{tab:APE_RPE_full_scale} lists the APE and RPE results of the three VO and VIO algorithms on each of the three datasets. Best performing values are written in bold font. Figure \ref{fig:Boxiplotti} shows the spread of the APE and RPE results of the algorithms on each dataset and algorithm. OpenVINS performs the best on the Complex Urban dataset with an $APE_\mathrm{full}$ of $20.374$. However, OpenVINS has a worse accuracy in the translational positional change $RPE_\mathrm{trans}$ than RTAB-Map, which has an $APE_\mathrm{full}$ of $110.971$. This is explained by the rotational error ($APE_\mathrm{rot}$), as of which RTAB-Maps is five times higher than OpenVINS's and drifts away from the ground-truth. Contrarily, the opposite is observable when VINS-Fusion is compared with RTAB-Map on the Kitti Odometry dataset. Here, the $APE_\mathrm{rot}$ is better for RTAB-map, and the $RPE_\mathrm{trans}$ is worse than VINS-Fusion, resulting in a better $APE_\mathrm{full}$ for VINS-Fusion due to the shorter sequence durations and lengths of Kitti Odometry than Complex Urban.

     \begin{table*}
		\centering
		\caption{Absolute and relative pose error over all sequences of the respective dataset by tested algorithms on the device under test, with which the respective dataset was recorded. The best-performing value is bold. }
        \label{tab:APE_RPE_full_scale}
		\begin{tabularx}{\textwidth}{@{} l | *{3}R | *{3}R | *{3}R }
			\toprule
            \multicolumn{10}{c}{Absolute Pose Error } \\
            \toprule
			\multirow{2}{*}{Algorithm} & \multicolumn{3}{c|}{$APE_\mathrm{full}$ [-]}           &       \multicolumn{3}{c|}{$APE_\mathrm{trans}$ [m]}              &      \multicolumn{3}{c}{$APE_\mathrm{rot}$ [-]} \\ 
            \cmidrule{2-10}
			                   & Complex Urban   & Kitti Odometry & EuRoC-MAV      & Complex Urban   & Kitti Odometry &  EuRoC-MAV     & Complex Urban  & Kitti Odometry  & EuRoC-MAV \\
			\midrule
			OpenVINS          & \textbf{20.374} &     -          & 0.353          & \textbf{20.374} &     -          & 0.332          & \textbf{0.066} &     -          & \textbf{0.110} \\
			VINS-Fusion       & 169.591         & \textbf{9.885} & \textbf{0.265} & 169.590         & \textbf{9.834} & \textbf{0.222} & 0.597          & 0.256          & 0.115 \\
			RTAB-Map (F2M)    & 110.971         & 10.879         & 0.454          & 110.971         & 10.878         & 0.314          & 0.345          & \textbf{0.070} & 0.308 \\
			\toprule
            \multicolumn{10}{c}{Relative Pose Error } \\
            \toprule
            \multirow{2}{*}{Algorithm} & \multicolumn{3}{c|}{$RPE_\mathrm{full}$ [-]}                &       \multicolumn{3}{c|}{$RPE_\mathrm{trans}$ [m]}              &      \multicolumn{3}{c}{$RPE_\mathrm{rot}$ [-]}    \\
            \cmidrule{2-10}
                OpenVINS          & 0.034           &     -          & \textbf{0.021} & 0.034           &     -          & 0.003          & \textbf{0.001} &     -          & \textbf{0.019} \\
			VINS-Fusion       & 0.029           & \textbf{0.055} & 0.110          & 0.026           & \textbf{0.055} & \textbf{0.001} & 0.013          & 0.002          & 0.110 \\
			RTAB-Map          & \textbf{0.020}  & 0.076          & 0.075          & \textbf{0.011}  & 0.076          & 0.003          & 0.016          & \textbf{0.002} & 0.075 \\
            \bottomrule
		\end{tabularx}
	\end{table*}
     
    \subsection{Comparison of Results}
    In comparison, the datasets for RSVR (self-recorded and EuRoC-MAV) and real-sized vehicles (Complex Urban and Kitti Odometry) show minor differences in the accuracy of $RPE_\mathrm{full}$, as the median for the RSVRs is $0.059$ with an interquartile range (IQR) of $0.008$--$0.12$ with a mean value of $0.07$. For the $RPE_\mathrm{full}$ of the real-sized vehicles, the median is $0.031$ (IQR $0.026$--$0.044$) with a mean value of $0.054$. Differences between real-sized vehicles and RSVRs were observed in the $RPE_\mathrm{rot}$ and $RPE_\mathrm{trans}$. For the experimental study with RSVRs (see Table \ref{tab:APE_RPE_reduced_scale} and Figure \ref{fig:Boxiplotti}), the $RPE_\mathrm{trans}$ is distributed at a median value of $0.0014\text{\,m}$ (IQR $0.0006\text{\,m}$--$0.003\text{\,m}$) with a mean value of $0.0022\text{\,m}$. For the real-sized vehicles, the $RPE_\mathrm{trans}$ is distributed at a median value of $0.03\text{\,m}$ (IQR $0.026\text{\,m}$--$0.044\text{\,m}$) with a mean value of $0.052\text{\,m}$. On the contrary, $RPE_\mathrm{rot}$ at the RSVRs is distributed at a median value of $0.059$ (IQR $0.007$--$0.12$) and a mean value of $0.069$. For real-sized vehicles, the $RPE_\mathrm{rot}$ is distributed at a median value of $0.0016$ (IQR $0.0013$--$0.003$) and a mean value of $0.004$. To summarize, the localization algorithms estimated the translational change in position worse for real-sized vehicles than for RSVR. However, the change in orientation was estimated worse for the RSVR than for the real-sized vehicles.

	\section{Discussion} \label{sec:discussion}

    By comparing the VO and VIO algorithms, we confirm, as discussed in \cite{li2014high,civera2009camera}, that the calibration of the sensor system is essential to estimate the driven trajectory. When a suitable calibration is provided, the accuracy of odometry estimation differs only slightly between the VO and VIO algorithms, as it is evident from the RPE values listed in Table \ref{tab:APE_RPE_full_scale} for each dataset, i.e., on the Kitti Odometry dataset or on our self-recorded data. OpenVINS is here at an advantage as it can, if enabled in the configuration, optimize the intrinsic calibration, e.g., the focal point and focal length, and extrinsic calibration, e.g., the offset of the camera to the IMU, during runtime \cite{Geneva2020ICRA}. VINS-Fusion is only able to optimize the extrinsic calibration. Contrarily, RTAB-Map is unable to optimize calibrations. However, RTAB-Map is more robust in case of a worse initial calibration than VINS-Fusion, as RTAB-Map utilizes more features than VINS-Fusion. As shown in Figure \ref{fig:odom paths}.a), even given the same initial calibration, VINS-Fusion is challenged in the first curve of our self-recorded data to find a good match, whereas later on, it follows the shape of the trajectory. However, if VINS-Fusion starts with a well-fitted calibration, we speculate that it could outperform OpenVINS and RTAB-Map like in Table \ref{tab:APE_RPE_full_scale} at Kitti-Odometry and EuRoC-MAV on $APE_\mathrm{trans}$ and $RPE_\mathrm{trans}$. Thus, a better calibration could improve the result of VINS-Fusion on our experimental data. \\
    In Section \ref{sec:results}, differences in the average accuracy of $RPE_\mathrm{trans}$ of $0.022\text{\,m}$ for RSVR versus $0.052\text{\,m}$ for the real-sized vehicles were found, showing that the algorithms have higher accuracy in estimating translational motion when applied to RSVR.
    However, the algorithms perform better estimating rotational motion when applied to real-sized vehicles, with a mean accuracy of $RPE_\mathrm{rot}$ of $0.069$ for RSVR against $0.004$ for real-sized vehicles. These differences in accuracy between $RPE_\mathrm{trans}$ and $RPE_\mathrm{rot}$ for RSVRs and real-sized vehicles, may stem from variations in the translational and rotational velocities across the datasets. The $95\,\%$-quantiles are considered as maximum values in the following. For instance, the EuRoC-MAV dataset features rotational velocities of ${|\omega|\leq\!2.1\text{\,rads}^{-1}}$ with translational velocities ${|v|\leq\!1.7\text{\,ms}^{-1}}$. In contrast, our self-recorded data exhibits lower maxima, with translational velocities ${|v|\leq\!1.5\text{\,ms}^{-1}}$ and rotational velocities ${|\omega|\leq\!0.7\text{\,rads}^{-1}}$. The Kitti Odometry dataset, on the contrary, includes higher speeds with translational velocities at ${|v|\leq\!15\text{\,ms}^{-1}}$, while rotational velocities remain at ${|\omega|\leq\!1\text{\,rads}^{-1}}$. Lastly, the Complex Urban dataset reaches a maximum translational velocity of $13\text{\,ms}^{-1}$ and a maximum rotational velocity of $0.3\text{\,rads}^{-1}$. Hence, the self-recorded data, Kitti Odometry, and Complex Urban have equivalent rotational velocities; comparing the $RPE_\mathrm{rot}$ of only the self-recorded sequence with the $RPE_\mathrm{rot}$ of Kitti Odometry and Complex Urban is more conclusive. Resulting in a mean $RPE_\mathrm{rot}$ of $0.0073$ for the self-recorded RSVR data compared to $0.0068$ for the datasets Kitti Odometry and Complex Urban, which feature real-sized vehicles. The error in $RPE_\mathrm{trans}$, measured on the mean $RPE_\mathrm{trans}$, is by a factor of $14.25$ lower for the RSVR than the real-sized vehicles at $9$ times lower maximum velocity. If $RPE_\mathrm{trans}$ is only calculated at speeds lower than the maximum velocity of the RSVR (i.e., $1.7\text{\,ms}^{-1}$), the results change to a median $RPE_\mathrm{trans}$ of $0.016\text{\,m}$ for real-sized vehicles and $0.003\text{\,m}$ for RSVRs. Thus, if only the data points on the datasets representing real-sized vehicles, where the respective vehicle drives up to the maximum velocity of the RSVR, are used, the difference of $RPE_\mathrm{trans}$ is significantly reduced by $267\text{\,\%}$. It should be noted that the results in Tables \ref{tab:APE_RPE_reduced_scale} and \ref{tab:APE_RPE_full_scale} do not show the result calculated with reduced velocities.
    
	\section{Conclusion and Future Work} \label{sec:conclusion}    
    In this study, we analyze different ROS2-compatible visual and visual-inertial self-localization algorithms to investigate whether these algorithms behave alike when applied to reduced-scale vehicular robots (RSVRs) and real-sized vehicles. Therefore, two standard datasets, Kitti Odometry and Complex Urban, provided a baseline for real-sized vehicles against EuRoC-MAV and our experimentally recorded data on an RSRV. We selected the self-localization algorithms VINS-Fusion, OpenVINS, and RTAB-Map for our evaluation. We use the evo tool \cite{grupp2017evo} to process the pose estimations of each algorithm to calculate the absolute pose error (APE) and relative pose error (RPE), respectively. Using these metrics, we find that the three chosen algorithms perform similarly on each scaling factor, i.e., for the real-sized vehicles on the datasets and the RSVR on our experimental data.  OpenVINS performs the best in four of the six RPE metrics on the experimental data and EuRoC-MAV dataset. However, for the real-sized vehicles, RTAB-Map and VINS-Fusion display the lowest error in three of the six RPE metrics. For the EuRoC-MAV dataset, we observe higher rotational RPE values due to its higher rotational velocities. Due to the higher translational velocities on the datasets for real-sized vehicles, we also observe higher translational RPE values on the datasets for real-sized vehicles, Complex Urban and Kitti Odometry, when compared against EuRoC-MAV and experimentally recorded data. Considering the fact that only lower velocities were tested on the RSVRs, we find a minor impact of the scaling factor for the translational RPE if only velocities similar to the RSVRs are considered for the real-sized vehicles. In contrast, the scaling does not impact the rotational estimation accuracy of the visual or visual-inertial odometry estimation with cameras and IMU. We, therefore, conclude that the localization algorithms tend to perform similarly independently of the scaling factor up to the tested $1.7\text{\,ms}^{-1}$ velocity of the RSVR. Hence, $1.7\text{\,ms}^{-1}$ corresponds to $8.5\text{\,ms}^{-1}$ using our scaling factor of $1\!:\!5$, representing slow urban traffic; this allows the localization algorithms to be tested on RSVR, aiding the development of automated driving for real-sized vehicles. \\
    Given that our tests with the RSVR were limited to velocities up to $1.7\text{\,ms}^{-1}$, experiments at higher velocities are of interest to determine if the self-localization algorithms behave generally similar. Furthermore, tests with higher dynamics are needed to test the reliability of the self-localization algorithms in critical traffic scenarios. Another possible research direction is the influence of the error of the ground-truth source on the accuracy estimation. Hence, if scaling is considered, the error of the ground-truth measurement could be scaled alongside. Lastly, collecting larger datasets are generally desirable to reduce the effects of, e.g., measurement noise on the presented results.
	\bibliography{references}         
	\bibliographystyle{ieeetr}                                               
	
\end{document}

%% file: Data/Colors.tex
\definecolor{THICOLOR}{rgb}{0,0.112,0.47}

\definecolor{BLAU}{RGB}{0,0,255}

\definecolor{YlGn31}{RGB}{247,252,185}

\definecolor{YlGn3C}{RGB}{247,252,185}

\definecolor{YlGn32}{RGB}{173,221,142}

\definecolor{YlGn3F}{RGB}{173,221,142}

\definecolor{YlGn33}{RGB}{49,163,84}

\definecolor{YlGn3I}{RGB}{49,163,84}

\definecolor{YlGn41}{RGB}{255,255,204}

\definecolor{YlGn4B}{RGB}{255,255,204}

\definecolor{YlGn42}{RGB}{194,230,153}

\definecolor{YlGn4E}{RGB}{194,230,153}

\definecolor{YlGn43}{RGB}{120,198,121}

\definecolor{YlGn4G}{RGB}{120,198,121}

\definecolor{YlGn44}{RGB}{35,132,67}

\definecolor{YlGn4J}{RGB}{35,132,67}

\definecolor{YlGn51}{RGB}{255,255,204}

\definecolor{YlGn5B}{RGB}{255,255,204}

\definecolor{YlGn52}{RGB}{194,230,153}

\definecolor{YlGn5E}{RGB}{194,230,153}

\definecolor{YlGn53}{RGB}{120,198,121}

\definecolor{YlGn5G}{RGB}{120,198,121}

\definecolor{YlGn54}{RGB}{49,163,84}

\definecolor{YlGn5I}{RGB}{49,163,84}

\definecolor{YlGn55}{RGB}{0,104,55}

\definecolor{YlGn5K}{RGB}{0,104,55}

\definecolor{YlGn61}{RGB}{255,255,204}

\definecolor{YlGn6B}{RGB}{255,255,204}

\definecolor{YlGn62}{RGB}{217,240,163}

\definecolor{YlGn6D}{RGB}{217,240,163}

\definecolor{YlGn63}{RGB}{173,221,142}

\definecolor{YlGn6F}{RGB}{173,221,142}

\definecolor{YlGn64}{RGB}{120,198,121}

\definecolor{YlGn6G}{RGB}{120,198,121}

\definecolor{YlGn65}{RGB}{49,163,84}

\definecolor{YlGn6I}{RGB}{49,163,84}

\definecolor{YlGn66}{RGB}{0,104,55}

\definecolor{YlGn6K}{RGB}{0,104,55}

\definecolor{YlGn71}{RGB}{255,255,204}

\definecolor{YlGn7B}{RGB}{255,255,204}

\definecolor{YlGn72}{RGB}{217,240,163}

\definecolor{YlGn7D}{RGB}{217,240,163}

\definecolor{YlGn73}{RGB}{173,221,142}

\definecolor{YlGn7F}{RGB}{173,221,142}

\definecolor{YlGn74}{RGB}{120,198,121}

\definecolor{YlGn7G}{RGB}{120,198,121}

\definecolor{YlGn75}{RGB}{65,171,93}

\definecolor{YlGn7H}{RGB}{65,171,93}

\definecolor{YlGn76}{RGB}{35,132,67}

\definecolor{YlGn7J}{RGB}{35,132,67}

\definecolor{YlGn77}{RGB}{0,90,50}

\definecolor{YlGn7L}{RGB}{0,90,50}

\definecolor{YlGn81}{RGB}{255,255,229}

\definecolor{YlGn8A}{RGB}{255,255,229}

\definecolor{YlGn82}{RGB}{247,252,185}

\definecolor{YlGn8C}{RGB}{247,252,185}

\definecolor{YlGn83}{RGB}{217,240,163}

\definecolor{YlGn8D}{RGB}{217,240,163}

\definecolor{YlGn84}{RGB}{173,221,142}

\definecolor{YlGn8F}{RGB}{173,221,142}

\definecolor{YlGn85}{RGB}{120,198,121}

\definecolor{YlGn8G}{RGB}{120,198,121}

\definecolor{YlGn86}{RGB}{65,171,93}

\definecolor{YlGn8H}{RGB}{65,171,93}

\definecolor{YlGn87}{RGB}{35,132,67}

\definecolor{YlGn8J}{RGB}{35,132,67}

\definecolor{YlGn88}{RGB}{0,90,50}

\definecolor{YlGn8L}{RGB}{0,90,50}

\definecolor{YlGn91}{RGB}{255,255,229}

\definecolor{YlGn9A}{RGB}{255,255,229}

\definecolor{YlGn92}{RGB}{247,252,185}

\definecolor{YlGn9C}{RGB}{247,252,185}

\definecolor{YlGn93}{RGB}{217,240,163}

\definecolor{YlGn9D}{RGB}{217,240,163}

\definecolor{YlGn94}{RGB}{173,221,142}

\definecolor{YlGn9F}{RGB}{173,221,142}

\definecolor{YlGn95}{RGB}{120,198,121}

\definecolor{YlGn9G}{RGB}{120,198,121}

\definecolor{YlGn96}{RGB}{65,171,93}

\definecolor{YlGn9H}{RGB}{65,171,93}

\definecolor{YlGn97}{RGB}{35,132,67}

\definecolor{YlGn9J}{RGB}{35,132,67}

\definecolor{YlGn98}{RGB}{0,104,55}

\definecolor{YlGn9K}{RGB}{0,104,55}

\definecolor{YlGn99}{RGB}{0,69,41}

\definecolor{YlGn9M}{RGB}{0,69,41}

\definecolor{YlGnBu31}{RGB}{237,248,177}

\definecolor{YlGnBu3C}{RGB}{237,248,177}

\definecolor{YlGnBu32}{RGB}{127,205,187}

\definecolor{YlGnBu3F}{RGB}{127,205,187}

\definecolor{YlGnBu33}{RGB}{44,127,184}

\definecolor{YlGnBu3I}{RGB}{44,127,184}

\definecolor{YlGnBu41}{RGB}{255,255,204}

\definecolor{YlGnBu4B}{RGB}{255,255,204}

\definecolor{YlGnBu42}{RGB}{161,218,180}

\definecolor{YlGnBu4E}{RGB}{161,218,180}

\definecolor{YlGnBu43}{RGB}{65,182,196}

\definecolor{YlGnBu4G}{RGB}{65,182,196}

\definecolor{YlGnBu44}{RGB}{34,94,168}

\definecolor{YlGnBu4J}{RGB}{34,94,168}

\definecolor{YlGnBu51}{RGB}{255,255,204}

\definecolor{YlGnBu5B}{RGB}{255,255,204}

\definecolor{YlGnBu52}{RGB}{161,218,180}

\definecolor{YlGnBu5E}{RGB}{161,218,180}

\definecolor{YlGnBu53}{RGB}{65,182,196}

\definecolor{YlGnBu5G}{RGB}{65,182,196}

\definecolor{YlGnBu54}{RGB}{44,127,184}

\definecolor{YlGnBu5I}{RGB}{44,127,184}

\definecolor{YlGnBu55}{RGB}{37,52,148}

\definecolor{YlGnBu5K}{RGB}{37,52,148}

\definecolor{YlGnBu61}{RGB}{255,255,204}

\definecolor{YlGnBu6B}{RGB}{255,255,204}

\definecolor{YlGnBu62}{RGB}{199,233,180}

\definecolor{YlGnBu6D}{RGB}{199,233,180}

\definecolor{YlGnBu63}{RGB}{127,205,187}

\definecolor{YlGnBu6F}{RGB}{127,205,187}

\definecolor{YlGnBu64}{RGB}{65,182,196}

\definecolor{YlGnBu6G}{RGB}{65,182,196}

\definecolor{YlGnBu65}{RGB}{44,127,184}

\definecolor{YlGnBu6I}{RGB}{44,127,184}

\definecolor{YlGnBu66}{RGB}{37,52,148}

\definecolor{YlGnBu6K}{RGB}{37,52,148}

\definecolor{YlGnBu71}{RGB}{255,255,204}

\definecolor{YlGnBu7B}{RGB}{255,255,204}

\definecolor{YlGnBu72}{RGB}{199,233,180}

\definecolor{YlGnBu7D}{RGB}{199,233,180}

\definecolor{YlGnBu73}{RGB}{127,205,187}

\definecolor{YlGnBu7F}{RGB}{127,205,187}

\definecolor{YlGnBu74}{RGB}{65,182,196}

\definecolor{YlGnBu7G}{RGB}{65,182,196}

\definecolor{YlGnBu75}{RGB}{29,145,192}

\definecolor{YlGnBu7H}{RGB}{29,145,192}

\definecolor{YlGnBu76}{RGB}{34,94,168}

\definecolor{YlGnBu7J}{RGB}{34,94,168}

\definecolor{YlGnBu77}{RGB}{12,44,132}

\definecolor{YlGnBu7L}{RGB}{12,44,132}

\definecolor{YlGnBu81}{RGB}{255,255,217}

\definecolor{YlGnBu8A}{RGB}{255,255,217}

\definecolor{YlGnBu82}{RGB}{237,248,177}

\definecolor{YlGnBu8C}{RGB}{237,248,177}

\definecolor{YlGnBu83}{RGB}{199,233,180}

\definecolor{YlGnBu8D}{RGB}{199,233,180}

\definecolor{YlGnBu84}{RGB}{127,205,187}

\definecolor{YlGnBu8F}{RGB}{127,205,187}

\definecolor{YlGnBu85}{RGB}{65,182,196}

\definecolor{YlGnBu8G}{RGB}{65,182,196}

\definecolor{YlGnBu86}{RGB}{29,145,192}

\definecolor{YlGnBu8H}{RGB}{29,145,192}

\definecolor{YlGnBu87}{RGB}{34,94,168}

\definecolor{YlGnBu8J}{RGB}{34,94,168}

\definecolor{YlGnBu88}{RGB}{12,44,132}

\definecolor{YlGnBu8L}{RGB}{12,44,132}

\definecolor{YlGnBu91}{RGB}{255,255,217}

\definecolor{YlGnBu9A}{RGB}{255,255,217}

\definecolor{YlGnBu92}{RGB}{237,248,177}

\definecolor{YlGnBu9C}{RGB}{237,248,177}

\definecolor{YlGnBu93}{RGB}{199,233,180}

\definecolor{YlGnBu9D}{RGB}{199,233,180}

\definecolor{YlGnBu94}{RGB}{127,205,187}

\definecolor{YlGnBu9F}{RGB}{127,205,187}

\definecolor{YlGnBu95}{RGB}{65,182,196}

\definecolor{YlGnBu9G}{RGB}{65,182,196}

\definecolor{YlGnBu96}{RGB}{29,145,192}

\definecolor{YlGnBu9H}{RGB}{29,145,192}

\definecolor{YlGnBu97}{RGB}{34,94,168}

\definecolor{YlGnBu9J}{RGB}{34,94,168}

\definecolor{YlGnBu98}{RGB}{37,52,148}

\definecolor{YlGnBu9K}{RGB}{37,52,148}

\definecolor{YlGnBu99}{RGB}{8,29,88}

\definecolor{YlGnBu9M}{RGB}{8,29,88}

\definecolor{GnBu31}{RGB}{224,243,219}

\definecolor{GnBu3C}{RGB}{224,243,219}

\definecolor{GnBu32}{RGB}{168,221,181}

\definecolor{GnBu3F}{RGB}{168,221,181}

\definecolor{GnBu33}{RGB}{67,162,202}

\definecolor{GnBu3I}{RGB}{67,162,202}

\definecolor{GnBu41}{RGB}{240,249,232}

\definecolor{GnBu4B}{RGB}{240,249,232}

\definecolor{GnBu42}{RGB}{186,228,188}

\definecolor{GnBu4E}{RGB}{186,228,188}

\definecolor{GnBu43}{RGB}{123,204,196}

\definecolor{GnBu4G}{RGB}{123,204,196}

\definecolor{GnBu44}{RGB}{43,140,190}

\definecolor{GnBu4J}{RGB}{43,140,190}

\definecolor{GnBu51}{RGB}{240,249,232}

\definecolor{GnBu5B}{RGB}{240,249,232}

\definecolor{GnBu52}{RGB}{186,228,188}

\definecolor{GnBu5E}{RGB}{186,228,188}

\definecolor{GnBu53}{RGB}{123,204,196}

\definecolor{GnBu5G}{RGB}{123,204,196}

\definecolor{GnBu54}{RGB}{67,162,202}

\definecolor{GnBu5I}{RGB}{67,162,202}

\definecolor{GnBu55}{RGB}{8,104,172}

\definecolor{GnBu5K}{RGB}{8,104,172}

\definecolor{GnBu61}{RGB}{240,249,232}

\definecolor{GnBu6B}{RGB}{240,249,232}

\definecolor{GnBu62}{RGB}{204,235,197}

\definecolor{GnBu6D}{RGB}{204,235,197}

\definecolor{GnBu63}{RGB}{168,221,181}

\definecolor{GnBu6F}{RGB}{168,221,181}

\definecolor{GnBu64}{RGB}{123,204,196}

\definecolor{GnBu6G}{RGB}{123,204,196}

\definecolor{GnBu65}{RGB}{67,162,202}

\definecolor{GnBu6I}{RGB}{67,162,202}

\definecolor{GnBu66}{RGB}{8,104,172}

\definecolor{GnBu6K}{RGB}{8,104,172}

\definecolor{GnBu71}{RGB}{240,249,232}

\definecolor{GnBu7B}{RGB}{240,249,232}

\definecolor{GnBu72}{RGB}{204,235,197}

\definecolor{GnBu7D}{RGB}{204,235,197}

\definecolor{GnBu73}{RGB}{168,221,181}

\definecolor{GnBu7F}{RGB}{168,221,181}

\definecolor{GnBu74}{RGB}{123,204,196}

\definecolor{GnBu7G}{RGB}{123,204,196}

\definecolor{GnBu75}{RGB}{78,179,211}

\definecolor{GnBu7H}{RGB}{78,179,211}

\definecolor{GnBu76}{RGB}{43,140,190}

\definecolor{GnBu7J}{RGB}{43,140,190}

\definecolor{GnBu77}{RGB}{8,88,158}

\definecolor{GnBu7L}{RGB}{8,88,158}

\definecolor{GnBu81}{RGB}{247,252,240}

\definecolor{GnBu8A}{RGB}{247,252,240}

\definecolor{GnBu82}{RGB}{224,243,219}

\definecolor{GnBu8C}{RGB}{224,243,219}

\definecolor{GnBu83}{RGB}{204,235,197}

\definecolor{GnBu8D}{RGB}{204,235,197}

\definecolor{GnBu84}{RGB}{168,221,181}

\definecolor{GnBu8F}{RGB}{168,221,181}

\definecolor{GnBu85}{RGB}{123,204,196}

\definecolor{GnBu8G}{RGB}{123,204,196}

\definecolor{GnBu86}{RGB}{78,179,211}

\definecolor{GnBu8H}{RGB}{78,179,211}

\definecolor{GnBu87}{RGB}{43,140,190}

\definecolor{GnBu8J}{RGB}{43,140,190}

\definecolor{GnBu88}{RGB}{8,88,158}

\definecolor{GnBu8L}{RGB}{8,88,158}

\definecolor{GnBu91}{RGB}{247,252,240}

\definecolor{GnBu9A}{RGB}{247,252,240}

\definecolor{GnBu92}{RGB}{224,243,219}

\definecolor{GnBu9C}{RGB}{224,243,219}

\definecolor{GnBu93}{RGB}{204,235,197}

\definecolor{GnBu9D}{RGB}{204,235,197}

\definecolor{GnBu94}{RGB}{168,221,181}

\definecolor{GnBu9F}{RGB}{168,221,181}

\definecolor{GnBu95}{RGB}{123,204,196}

\definecolor{GnBu9G}{RGB}{123,204,196}

\definecolor{GnBu96}{RGB}{78,179,211}

\definecolor{GnBu9H}{RGB}{78,179,211}

\definecolor{GnBu97}{RGB}{43,140,190}

\definecolor{GnBu9J}{RGB}{43,140,190}

\definecolor{GnBu98}{RGB}{8,104,172}

\definecolor{GnBu9K}{RGB}{8,104,172}

\definecolor{GnBu99}{RGB}{8,64,129}

\definecolor{GnBu9M}{RGB}{8,64,129}

\definecolor{BuGn31}{RGB}{229,245,249}

\definecolor{BuGn3C}{RGB}{229,245,249}

\definecolor{BuGn32}{RGB}{153,216,201}

\definecolor{BuGn3F}{RGB}{153,216,201}

\definecolor{BuGn33}{RGB}{44,162,95}

\definecolor{BuGn3I}{RGB}{44,162,95}

\definecolor{BuGn41}{RGB}{237,248,251}

\definecolor{BuGn4B}{RGB}{237,248,251}

\definecolor{BuGn42}{RGB}{178,226,226}

\definecolor{BuGn4E}{RGB}{178,226,226}

\definecolor{BuGn43}{RGB}{102,194,164}

\definecolor{BuGn4G}{RGB}{102,194,164}

\definecolor{BuGn44}{RGB}{35,139,69}

\definecolor{BuGn4J}{RGB}{35,139,69}

\definecolor{BuGn51}{RGB}{237,248,251}

\definecolor{BuGn5B}{RGB}{237,248,251}

\definecolor{BuGn52}{RGB}{178,226,226}

\definecolor{BuGn5E}{RGB}{178,226,226}

\definecolor{BuGn53}{RGB}{102,194,164}

\definecolor{BuGn5G}{RGB}{102,194,164}

\definecolor{BuGn54}{RGB}{44,162,95}

\definecolor{BuGn5I}{RGB}{44,162,95}

\definecolor{BuGn55}{RGB}{0,109,44}

\definecolor{BuGn5K}{RGB}{0,109,44}

\definecolor{BuGn61}{RGB}{237,248,251}

\definecolor{BuGn6B}{RGB}{237,248,251}

\definecolor{BuGn62}{RGB}{204,236,230}

\definecolor{BuGn6D}{RGB}{204,236,230}

\definecolor{BuGn63}{RGB}{153,216,201}

\definecolor{BuGn6F}{RGB}{153,216,201}

\definecolor{BuGn64}{RGB}{102,194,164}

\definecolor{BuGn6G}{RGB}{102,194,164}

\definecolor{BuGn65}{RGB}{44,162,95}

\definecolor{BuGn6I}{RGB}{44,162,95}

\definecolor{BuGn66}{RGB}{0,109,44}

\definecolor{BuGn6K}{RGB}{0,109,44}

\definecolor{BuGn71}{RGB}{237,248,251}

\definecolor{BuGn7B}{RGB}{237,248,251}

\definecolor{BuGn72}{RGB}{204,236,230}

\definecolor{BuGn7D}{RGB}{204,236,230}

\definecolor{BuGn73}{RGB}{153,216,201}

\definecolor{BuGn7F}{RGB}{153,216,201}

\definecolor{BuGn74}{RGB}{102,194,164}

\definecolor{BuGn7G}{RGB}{102,194,164}

\definecolor{BuGn75}{RGB}{65,174,118}

\definecolor{BuGn7H}{RGB}{65,174,118}

\definecolor{BuGn76}{RGB}{35,139,69}

\definecolor{BuGn7J}{RGB}{35,139,69}

\definecolor{BuGn77}{RGB}{0,88,36}

\definecolor{BuGn7L}{RGB}{0,88,36}

\definecolor{BuGn81}{RGB}{247,252,253}

\definecolor{BuGn8A}{RGB}{247,252,253}

\definecolor{BuGn82}{RGB}{229,245,249}

\definecolor{BuGn8C}{RGB}{229,245,249}

\definecolor{BuGn83}{RGB}{204,236,230}

\definecolor{BuGn8D}{RGB}{204,236,230}

\definecolor{BuGn84}{RGB}{153,216,201}

\definecolor{BuGn8F}{RGB}{153,216,201}

\definecolor{BuGn85}{RGB}{102,194,164}

\definecolor{BuGn8G}{RGB}{102,194,164}

\definecolor{BuGn86}{RGB}{65,174,118}

\definecolor{BuGn8H}{RGB}{65,174,118}

\definecolor{BuGn87}{RGB}{35,139,69}

\definecolor{BuGn8J}{RGB}{35,139,69}

\definecolor{BuGn88}{RGB}{0,88,36}

\definecolor{BuGn8L}{RGB}{0,88,36}

\definecolor{BuGn91}{RGB}{247,252,253}

\definecolor{BuGn9A}{RGB}{247,252,253}

\definecolor{BuGn92}{RGB}{229,245,249}

\definecolor{BuGn9C}{RGB}{229,245,249}

\definecolor{BuGn93}{RGB}{204,236,230}

\definecolor{BuGn9D}{RGB}{204,236,230}

\definecolor{BuGn94}{RGB}{153,216,201}

\definecolor{BuGn9F}{RGB}{153,216,201}

\definecolor{BuGn95}{RGB}{102,194,164}

\definecolor{BuGn9G}{RGB}{102,194,164}

\definecolor{BuGn96}{RGB}{65,174,118}

\definecolor{BuGn9H}{RGB}{65,174,118}

\definecolor{BuGn97}{RGB}{35,139,69}

\definecolor{BuGn9J}{RGB}{35,139,69}

\definecolor{BuGn98}{RGB}{0,109,44}

\definecolor{BuGn9K}{RGB}{0,109,44}

\definecolor{BuGn99}{RGB}{0,68,27}

\definecolor{BuGn9M}{RGB}{0,68,27}

\definecolor{PuBuGn31}{RGB}{236,226,240}

\definecolor{PuBuGn3C}{RGB}{236,226,240}

\definecolor{PuBuGn32}{RGB}{166,189,219}

\definecolor{PuBuGn3F}{RGB}{166,189,219}

\definecolor{PuBuGn33}{RGB}{28,144,153}

\definecolor{PuBuGn3I}{RGB}{28,144,153}

\definecolor{PuBuGn41}{RGB}{246,239,247}

\definecolor{PuBuGn4B}{RGB}{246,239,247}

\definecolor{PuBuGn42}{RGB}{189,201,225}

\definecolor{PuBuGn4E}{RGB}{189,201,225}

\definecolor{PuBuGn43}{RGB}{103,169,207}

\definecolor{PuBuGn4G}{RGB}{103,169,207}

\definecolor{PuBuGn44}{RGB}{2,129,138}

\definecolor{PuBuGn4J}{RGB}{2,129,138}

\definecolor{PuBuGn51}{RGB}{246,239,247}

\definecolor{PuBuGn5B}{RGB}{246,239,247}

\definecolor{PuBuGn52}{RGB}{189,201,225}

\definecolor{PuBuGn5E}{RGB}{189,201,225}

\definecolor{PuBuGn53}{RGB}{103,169,207}

\definecolor{PuBuGn5G}{RGB}{103,169,207}

\definecolor{PuBuGn54}{RGB}{28,144,153}

\definecolor{PuBuGn5I}{RGB}{28,144,153}

\definecolor{PuBuGn55}{RGB}{1,108,89}

\definecolor{PuBuGn5K}{RGB}{1,108,89}

\definecolor{PuBuGn61}{RGB}{246,239,247}

\definecolor{PuBuGn6B}{RGB}{246,239,247}

\definecolor{PuBuGn62}{RGB}{208,209,230}

\definecolor{PuBuGn6D}{RGB}{208,209,230}

\definecolor{PuBuGn63}{RGB}{166,189,219}

\definecolor{PuBuGn6F}{RGB}{166,189,219}

\definecolor{PuBuGn64}{RGB}{103,169,207}

\definecolor{PuBuGn6G}{RGB}{103,169,207}

\definecolor{PuBuGn65}{RGB}{28,144,153}

\definecolor{PuBuGn6I}{RGB}{28,144,153}

\definecolor{PuBuGn66}{RGB}{1,108,89}

\definecolor{PuBuGn6K}{RGB}{1,108,89}

\definecolor{PuBuGn71}{RGB}{246,239,247}

\definecolor{PuBuGn7B}{RGB}{246,239,247}

\definecolor{PuBuGn72}{RGB}{208,209,230}

\definecolor{PuBuGn7D}{RGB}{208,209,230}

\definecolor{PuBuGn73}{RGB}{166,189,219}

\definecolor{PuBuGn7F}{RGB}{166,189,219}

\definecolor{PuBuGn74}{RGB}{103,169,207}

\definecolor{PuBuGn7G}{RGB}{103,169,207}

\definecolor{PuBuGn75}{RGB}{54,144,192}

\definecolor{PuBuGn7H}{RGB}{54,144,192}

\definecolor{PuBuGn76}{RGB}{2,129,138}

\definecolor{PuBuGn7J}{RGB}{2,129,138}

\definecolor{PuBuGn77}{RGB}{1,100,80}

\definecolor{PuBuGn7L}{RGB}{1,100,80}

\definecolor{PuBuGn81}{RGB}{255,247,251}

\definecolor{PuBuGn8A}{RGB}{255,247,251}

\definecolor{PuBuGn82}{RGB}{236,226,240}

\definecolor{PuBuGn8C}{RGB}{236,226,240}

\definecolor{PuBuGn83}{RGB}{208,209,230}

\definecolor{PuBuGn8D}{RGB}{208,209,230}

\definecolor{PuBuGn84}{RGB}{166,189,219}

\definecolor{PuBuGn8F}{RGB}{166,189,219}

\definecolor{PuBuGn85}{RGB}{103,169,207}

\definecolor{PuBuGn8G}{RGB}{103,169,207}

\definecolor{PuBuGn86}{RGB}{54,144,192}

\definecolor{PuBuGn8H}{RGB}{54,144,192}

\definecolor{PuBuGn87}{RGB}{2,129,138}

\definecolor{PuBuGn8J}{RGB}{2,129,138}

\definecolor{PuBuGn88}{RGB}{1,100,80}

\definecolor{PuBuGn8L}{RGB}{1,100,80}

\definecolor{PuBuGn91}{RGB}{255,247,251}

\definecolor{PuBuGn9A}{RGB}{255,247,251}

\definecolor{PuBuGn92}{RGB}{236,226,240}

\definecolor{PuBuGn9C}{RGB}{236,226,240}

\definecolor{PuBuGn93}{RGB}{208,209,230}

\definecolor{PuBuGn9D}{RGB}{208,209,230}

\definecolor{PuBuGn94}{RGB}{166,189,219}

\definecolor{PuBuGn9F}{RGB}{166,189,219}

\definecolor{PuBuGn95}{RGB}{103,169,207}

\definecolor{PuBuGn9G}{RGB}{103,169,207}

\definecolor{PuBuGn96}{RGB}{54,144,192}

\definecolor{PuBuGn9H}{RGB}{54,144,192}

\definecolor{PuBuGn97}{RGB}{2,129,138}

\definecolor{PuBuGn9J}{RGB}{2,129,138}

\definecolor{PuBuGn98}{RGB}{1,108,89}

\definecolor{PuBuGn9K}{RGB}{1,108,89}

\definecolor{PuBuGn99}{RGB}{1,70,54}

\definecolor{PuBuGn9M}{RGB}{1,70,54}

\definecolor{PuBu31}{RGB}{236,231,242}

\definecolor{PuBu3C}{RGB}{236,231,242}

\definecolor{PuBu32}{RGB}{166,189,219}

\definecolor{PuBu3F}{RGB}{166,189,219}

\definecolor{PuBu33}{RGB}{43,140,190}

\definecolor{PuBu3I}{RGB}{43,140,190}

\definecolor{PuBu41}{RGB}{241,238,246}

\definecolor{PuBu4B}{RGB}{241,238,246}

\definecolor{PuBu42}{RGB}{189,201,225}

\definecolor{PuBu4E}{RGB}{189,201,225}

\definecolor{PuBu43}{RGB}{116,169,207}

\definecolor{PuBu4G}{RGB}{116,169,207}

\definecolor{PuBu44}{RGB}{5,112,176}

\definecolor{PuBu4J}{RGB}{5,112,176}

\definecolor{PuBu51}{RGB}{241,238,246}

\definecolor{PuBu5B}{RGB}{241,238,246}

\definecolor{PuBu52}{RGB}{189,201,225}

\definecolor{PuBu5E}{RGB}{189,201,225}

\definecolor{PuBu53}{RGB}{116,169,207}

\definecolor{PuBu5G}{RGB}{116,169,207}

\definecolor{PuBu54}{RGB}{43,140,190}

\definecolor{PuBu5I}{RGB}{43,140,190}

\definecolor{PuBu55}{RGB}{4,90,141}

\definecolor{PuBu5K}{RGB}{4,90,141}

\definecolor{PuBu61}{RGB}{241,238,246}

\definecolor{PuBu6B}{RGB}{241,238,246}

\definecolor{PuBu62}{RGB}{208,209,230}

\definecolor{PuBu6D}{RGB}{208,209,230}

\definecolor{PuBu63}{RGB}{166,189,219}

\definecolor{PuBu6F}{RGB}{166,189,219}

\definecolor{PuBu64}{RGB}{116,169,207}

\definecolor{PuBu6G}{RGB}{116,169,207}

\definecolor{PuBu65}{RGB}{43,140,190}

\definecolor{PuBu6I}{RGB}{43,140,190}

\definecolor{PuBu66}{RGB}{4,90,141}

\definecolor{PuBu6K}{RGB}{4,90,141}

\definecolor{PuBu71}{RGB}{241,238,246}

\definecolor{PuBu7B}{RGB}{241,238,246}

\definecolor{PuBu72}{RGB}{208,209,230}

\definecolor{PuBu7D}{RGB}{208,209,230}

\definecolor{PuBu73}{RGB}{166,189,219}

\definecolor{PuBu7F}{RGB}{166,189,219}

\definecolor{PuBu74}{RGB}{116,169,207}

\definecolor{PuBu7G}{RGB}{116,169,207}

\definecolor{PuBu75}{RGB}{54,144,192}

\definecolor{PuBu7H}{RGB}{54,144,192}

\definecolor{PuBu76}{RGB}{5,112,176}

\definecolor{PuBu7J}{RGB}{5,112,176}

\definecolor{PuBu77}{RGB}{3,78,123}

\definecolor{PuBu7L}{RGB}{3,78,123}

\definecolor{PuBu81}{RGB}{255,247,251}

\definecolor{PuBu8A}{RGB}{255,247,251}

\definecolor{PuBu82}{RGB}{236,231,242}

\definecolor{PuBu8C}{RGB}{236,231,242}

\definecolor{PuBu83}{RGB}{208,209,230}

\definecolor{PuBu8D}{RGB}{208,209,230}

\definecolor{PuBu84}{RGB}{166,189,219}

\definecolor{PuBu8F}{RGB}{166,189,219}

\definecolor{PuBu85}{RGB}{116,169,207}

\definecolor{PuBu8G}{RGB}{116,169,207}

\definecolor{PuBu86}{RGB}{54,144,192}

\definecolor{PuBu8H}{RGB}{54,144,192}

\definecolor{PuBu87}{RGB}{5,112,176}

\definecolor{PuBu8J}{RGB}{5,112,176}

\definecolor{PuBu88}{RGB}{3,78,123}

\definecolor{PuBu8L}{RGB}{3,78,123}

\definecolor{PuBu91}{RGB}{255,247,251}

\definecolor{PuBu9A}{RGB}{255,247,251}

\definecolor{PuBu92}{RGB}{236,231,242}

\definecolor{PuBu9C}{RGB}{236,231,242}

\definecolor{PuBu93}{RGB}{208,209,230}

\definecolor{PuBu9D}{RGB}{208,209,230}

\definecolor{PuBu94}{RGB}{166,189,219}

\definecolor{PuBu9F}{RGB}{166,189,219}

\definecolor{PuBu95}{RGB}{116,169,207}

\definecolor{PuBu9G}{RGB}{116,169,207}

\definecolor{PuBu96}{RGB}{54,144,192}

\definecolor{PuBu9H}{RGB}{54,144,192}

\definecolor{PuBu97}{RGB}{5,112,176}

\definecolor{PuBu9J}{RGB}{5,112,176}

\definecolor{PuBu98}{RGB}{4,90,141}

\definecolor{PuBu9K}{RGB}{4,90,141}

\definecolor{PuBu99}{RGB}{2,56,88}

\definecolor{PuBu9M}{RGB}{2,56,88}

\definecolor{BuPu31}{RGB}{224,236,244}

\definecolor{BuPu3C}{RGB}{224,236,244}

\definecolor{BuPu32}{RGB}{158,188,218}

\definecolor{BuPu3F}{RGB}{158,188,218}

\definecolor{BuPu33}{RGB}{136,86,167}

\definecolor{BuPu3I}{RGB}{136,86,167}

\definecolor{BuPu41}{RGB}{237,248,251}

\definecolor{BuPu4B}{RGB}{237,248,251}

\definecolor{BuPu42}{RGB}{179,205,227}

\definecolor{BuPu4E}{RGB}{179,205,227}

\definecolor{BuPu43}{RGB}{140,150,198}

\definecolor{BuPu4G}{RGB}{140,150,198}

\definecolor{BuPu44}{RGB}{136,65,157}

\definecolor{BuPu4J}{RGB}{136,65,157}

\definecolor{BuPu51}{RGB}{237,248,251}

\definecolor{BuPu5B}{RGB}{237,248,251}

\definecolor{BuPu52}{RGB}{179,205,227}

\definecolor{BuPu5E}{RGB}{179,205,227}

\definecolor{BuPu53}{RGB}{140,150,198}

\definecolor{BuPu5G}{RGB}{140,150,198}

\definecolor{BuPu54}{RGB}{136,86,167}

\definecolor{BuPu5I}{RGB}{136,86,167}

\definecolor{BuPu55}{RGB}{129,15,124}

\definecolor{BuPu5K}{RGB}{129,15,124}

\definecolor{BuPu61}{RGB}{237,248,251}

\definecolor{BuPu6B}{RGB}{237,248,251}

\definecolor{BuPu62}{RGB}{191,211,230}

\definecolor{BuPu6D}{RGB}{191,211,230}

\definecolor{BuPu63}{RGB}{158,188,218}

\definecolor{BuPu6F}{RGB}{158,188,218}

\definecolor{BuPu64}{RGB}{140,150,198}

\definecolor{BuPu6G}{RGB}{140,150,198}

\definecolor{BuPu65}{RGB}{136,86,167}

\definecolor{BuPu6I}{RGB}{136,86,167}

\definecolor{BuPu66}{RGB}{129,15,124}

\definecolor{BuPu6K}{RGB}{129,15,124}

\definecolor{BuPu71}{RGB}{237,248,251}

\definecolor{BuPu7B}{RGB}{237,248,251}

\definecolor{BuPu72}{RGB}{191,211,230}

\definecolor{BuPu7D}{RGB}{191,211,230}

\definecolor{BuPu73}{RGB}{158,188,218}

\definecolor{BuPu7F}{RGB}{158,188,218}

\definecolor{BuPu74}{RGB}{140,150,198}

\definecolor{BuPu7G}{RGB}{140,150,198}

\definecolor{BuPu75}{RGB}{140,107,177}

\definecolor{BuPu7H}{RGB}{140,107,177}

\definecolor{BuPu76}{RGB}{136,65,157}

\definecolor{BuPu7J}{RGB}{136,65,157}

\definecolor{BuPu77}{RGB}{110,1,107}

\definecolor{BuPu7L}{RGB}{110,1,107}

\definecolor{BuPu81}{RGB}{247,252,253}

\definecolor{BuPu8A}{RGB}{247,252,253}

\definecolor{BuPu82}{RGB}{224,236,244}

\definecolor{BuPu8C}{RGB}{224,236,244}

\definecolor{BuPu83}{RGB}{191,211,230}

\definecolor{BuPu8D}{RGB}{191,211,230}

\definecolor{BuPu84}{RGB}{158,188,218}

\definecolor{BuPu8F}{RGB}{158,188,218}

\definecolor{BuPu85}{RGB}{140,150,198}

\definecolor{BuPu8G}{RGB}{140,150,198}

\definecolor{BuPu86}{RGB}{140,107,177}

\definecolor{BuPu8H}{RGB}{140,107,177}

\definecolor{BuPu87}{RGB}{136,65,157}

\definecolor{BuPu8J}{RGB}{136,65,157}

\definecolor{BuPu88}{RGB}{110,1,107}

\definecolor{BuPu8L}{RGB}{110,1,107}

\definecolor{BuPu91}{RGB}{247,252,253}

\definecolor{BuPu9A}{RGB}{247,252,253}

\definecolor{BuPu92}{RGB}{224,236,244}

\definecolor{BuPu9C}{RGB}{224,236,244}

\definecolor{BuPu93}{RGB}{191,211,230}

\definecolor{BuPu9D}{RGB}{191,211,230}

\definecolor{BuPu94}{RGB}{158,188,218}

\definecolor{BuPu9F}{RGB}{158,188,218}

\definecolor{BuPu95}{RGB}{140,150,198}

\definecolor{BuPu9G}{RGB}{140,150,198}

\definecolor{BuPu96}{RGB}{140,107,177}

\definecolor{BuPu9H}{RGB}{140,107,177}

\definecolor{BuPu97}{RGB}{136,65,157}

\definecolor{BuPu9J}{RGB}{136,65,157}

\definecolor{BuPu98}{RGB}{129,15,124}

\definecolor{BuPu9K}{RGB}{129,15,124}

\definecolor{BuPu99}{RGB}{77,0,75}

\definecolor{BuPu9M}{RGB}{77,0,75}

\definecolor{RdPu31}{RGB}{253,224,221}

\definecolor{RdPu3C}{RGB}{253,224,221}

\definecolor{RdPu32}{RGB}{250,159,181}

\definecolor{RdPu3F}{RGB}{250,159,181}

\definecolor{RdPu33}{RGB}{197,27,138}

\definecolor{RdPu3I}{RGB}{197,27,138}

\definecolor{RdPu41}{RGB}{254,235,226}

\definecolor{RdPu4B}{RGB}{254,235,226}

\definecolor{RdPu42}{RGB}{251,180,185}

\definecolor{RdPu4E}{RGB}{251,180,185}

\definecolor{RdPu43}{RGB}{247,104,161}

\definecolor{RdPu4G}{RGB}{247,104,161}

\definecolor{RdPu44}{RGB}{174,1,126}

\definecolor{RdPu4J}{RGB}{174,1,126}

\definecolor{RdPu51}{RGB}{254,235,226}

\definecolor{RdPu5B}{RGB}{254,235,226}

\definecolor{RdPu52}{RGB}{251,180,185}

\definecolor{RdPu5E}{RGB}{251,180,185}

\definecolor{RdPu53}{RGB}{247,104,161}

\definecolor{RdPu5G}{RGB}{247,104,161}

\definecolor{RdPu54}{RGB}{197,27,138}

\definecolor{RdPu5I}{RGB}{197,27,138}

\definecolor{RdPu55}{RGB}{122,1,119}

\definecolor{RdPu5K}{RGB}{122,1,119}

\definecolor{RdPu61}{RGB}{254,235,226}

\definecolor{RdPu6B}{RGB}{254,235,226}

\definecolor{RdPu62}{RGB}{252,197,192}

\definecolor{RdPu6D}{RGB}{252,197,192}

\definecolor{RdPu63}{RGB}{250,159,181}

\definecolor{RdPu6F}{RGB}{250,159,181}

\definecolor{RdPu64}{RGB}{247,104,161}

\definecolor{RdPu6G}{RGB}{247,104,161}

\definecolor{RdPu65}{RGB}{197,27,138}

\definecolor{RdPu6I}{RGB}{197,27,138}

\definecolor{RdPu66}{RGB}{122,1,119}

\definecolor{RdPu6K}{RGB}{122,1,119}

\definecolor{RdPu71}{RGB}{254,235,226}

\definecolor{RdPu7B}{RGB}{254,235,226}

\definecolor{RdPu72}{RGB}{252,197,192}

\definecolor{RdPu7D}{RGB}{252,197,192}

\definecolor{RdPu73}{RGB}{250,159,181}

\definecolor{RdPu7F}{RGB}{250,159,181}

\definecolor{RdPu74}{RGB}{247,104,161}

\definecolor{RdPu7G}{RGB}{247,104,161}

\definecolor{RdPu75}{RGB}{221,52,151}

\definecolor{RdPu7H}{RGB}{221,52,151}

\definecolor{RdPu76}{RGB}{174,1,126}

\definecolor{RdPu7J}{RGB}{174,1,126}

\definecolor{RdPu77}{RGB}{122,1,119}

\definecolor{RdPu7L}{RGB}{122,1,119}

\definecolor{RdPu81}{RGB}{255,247,243}

\definecolor{RdPu8A}{RGB}{255,247,243}

\definecolor{RdPu82}{RGB}{253,224,221}

\definecolor{RdPu8C}{RGB}{253,224,221}

\definecolor{RdPu83}{RGB}{252,197,192}

\definecolor{RdPu8D}{RGB}{252,197,192}

\definecolor{RdPu84}{RGB}{250,159,181}

\definecolor{RdPu8F}{RGB}{250,159,181}

\definecolor{RdPu85}{RGB}{247,104,161}

\definecolor{RdPu8G}{RGB}{247,104,161}

\definecolor{RdPu86}{RGB}{221,52,151}

\definecolor{RdPu8H}{RGB}{221,52,151}

\definecolor{RdPu87}{RGB}{174,1,126}

\definecolor{RdPu8J}{RGB}{174,1,126}

\definecolor{RdPu88}{RGB}{122,1,119}

\definecolor{RdPu8L}{RGB}{122,1,119}

\definecolor{RdPu91}{RGB}{255,247,243}

\definecolor{RdPu9A}{RGB}{255,247,243}

\definecolor{RdPu92}{RGB}{253,224,221}

\definecolor{RdPu9C}{RGB}{253,224,221}

\definecolor{RdPu93}{RGB}{252,197,192}

\definecolor{RdPu9D}{RGB}{252,197,192}

\definecolor{RdPu94}{RGB}{250,159,181}

\definecolor{RdPu9F}{RGB}{250,159,181}

\definecolor{RdPu95}{RGB}{247,104,161}

\definecolor{RdPu9G}{RGB}{247,104,161}

\definecolor{RdPu96}{RGB}{221,52,151}

\definecolor{RdPu9H}{RGB}{221,52,151}

\definecolor{RdPu97}{RGB}{174,1,126}

\definecolor{RdPu9J}{RGB}{174,1,126}

\definecolor{RdPu98}{RGB}{122,1,119}

\definecolor{RdPu9K}{RGB}{122,1,119}

\definecolor{RdPu99}{RGB}{73,0,106}

\definecolor{RdPu9M}{RGB}{73,0,106}

\definecolor{PuRd31}{RGB}{231,225,239}

\definecolor{PuRd3C}{RGB}{231,225,239}

\definecolor{PuRd32}{RGB}{201,148,199}

\definecolor{PuRd3F}{RGB}{201,148,199}

\definecolor{PuRd33}{RGB}{221,28,119}

\definecolor{PuRd3I}{RGB}{221,28,119}

\definecolor{PuRd41}{RGB}{241,238,246}

\definecolor{PuRd4B}{RGB}{241,238,246}

\definecolor{PuRd42}{RGB}{215,181,216}

\definecolor{PuRd4E}{RGB}{215,181,216}

\definecolor{PuRd43}{RGB}{223,101,176}

\definecolor{PuRd4G}{RGB}{223,101,176}

\definecolor{PuRd44}{RGB}{206,18,86}

\definecolor{PuRd4J}{RGB}{206,18,86}

\definecolor{PuRd51}{RGB}{241,238,246}

\definecolor{PuRd5B}{RGB}{241,238,246}

\definecolor{PuRd52}{RGB}{215,181,216}

\definecolor{PuRd5E}{RGB}{215,181,216}

\definecolor{PuRd53}{RGB}{223,101,176}

\definecolor{PuRd5G}{RGB}{223,101,176}

\definecolor{PuRd54}{RGB}{221,28,119}

\definecolor{PuRd5I}{RGB}{221,28,119}

\definecolor{PuRd55}{RGB}{152,0,67}

\definecolor{PuRd5K}{RGB}{152,0,67}

\definecolor{PuRd61}{RGB}{241,238,246}

\definecolor{PuRd6B}{RGB}{241,238,246}

\definecolor{PuRd62}{RGB}{212,185,218}

\definecolor{PuRd6D}{RGB}{212,185,218}

\definecolor{PuRd63}{RGB}{201,148,199}

\definecolor{PuRd6F}{RGB}{201,148,199}

\definecolor{PuRd64}{RGB}{223,101,176}

\definecolor{PuRd6G}{RGB}{223,101,176}

\definecolor{PuRd65}{RGB}{221,28,119}

\definecolor{PuRd6I}{RGB}{221,28,119}

\definecolor{PuRd66}{RGB}{152,0,67}

\definecolor{PuRd6K}{RGB}{152,0,67}

\definecolor{PuRd71}{RGB}{241,238,246}

\definecolor{PuRd7B}{RGB}{241,238,246}

\definecolor{PuRd72}{RGB}{212,185,218}

\definecolor{PuRd7D}{RGB}{212,185,218}

\definecolor{PuRd73}{RGB}{201,148,199}

\definecolor{PuRd7F}{RGB}{201,148,199}

\definecolor{PuRd74}{RGB}{223,101,176}

\definecolor{PuRd7G}{RGB}{223,101,176}

\definecolor{PuRd75}{RGB}{231,41,138}

\definecolor{PuRd7H}{RGB}{231,41,138}

\definecolor{PuRd76}{RGB}{206,18,86}

\definecolor{PuRd7J}{RGB}{206,18,86}

\definecolor{PuRd77}{RGB}{145,0,63}

\definecolor{PuRd7L}{RGB}{145,0,63}

\definecolor{PuRd81}{RGB}{247,244,249}

\definecolor{PuRd8A}{RGB}{247,244,249}

\definecolor{PuRd82}{RGB}{231,225,239}

\definecolor{PuRd8C}{RGB}{231,225,239}

\definecolor{PuRd83}{RGB}{212,185,218}

\definecolor{PuRd8D}{RGB}{212,185,218}

\definecolor{PuRd84}{RGB}{201,148,199}

\definecolor{PuRd8F}{RGB}{201,148,199}

\definecolor{PuRd85}{RGB}{223,101,176}

\definecolor{PuRd8G}{RGB}{223,101,176}

\definecolor{PuRd86}{RGB}{231,41,138}

\definecolor{PuRd8H}{RGB}{231,41,138}

\definecolor{PuRd87}{RGB}{206,18,86}

\definecolor{PuRd8J}{RGB}{206,18,86}

\definecolor{PuRd88}{RGB}{145,0,63}

\definecolor{PuRd8L}{RGB}{145,0,63}

\definecolor{PuRd91}{RGB}{247,244,249}

\definecolor{PuRd9A}{RGB}{247,244,249}

\definecolor{PuRd92}{RGB}{231,225,239}

\definecolor{PuRd9C}{RGB}{231,225,239}

\definecolor{PuRd93}{RGB}{212,185,218}

\definecolor{PuRd9D}{RGB}{212,185,218}

\definecolor{PuRd94}{RGB}{201,148,199}

\definecolor{PuRd9F}{RGB}{201,148,199}

\definecolor{PuRd95}{RGB}{223,101,176}

\definecolor{PuRd9G}{RGB}{223,101,176}

\definecolor{PuRd96}{RGB}{231,41,138}

\definecolor{PuRd9H}{RGB}{231,41,138}

\definecolor{PuRd97}{RGB}{206,18,86}

\definecolor{PuRd9J}{RGB}{206,18,86}

\definecolor{PuRd98}{RGB}{152,0,67}

\definecolor{PuRd9K}{RGB}{152,0,67}

\definecolor{PuRd99}{RGB}{103,0,31}

\definecolor{PuRd9M}{RGB}{103,0,31}

\definecolor{OrRd31}{RGB}{254,232,200}

\definecolor{OrRd3C}{RGB}{254,232,200}

\definecolor{OrRd32}{RGB}{253,187,132}

\definecolor{OrRd3F}{RGB}{253,187,132}

\definecolor{OrRd33}{RGB}{227,74,51}

\definecolor{OrRd3I}{RGB}{227,74,51}

\definecolor{OrRd41}{RGB}{254,240,217}

\definecolor{OrRd4B}{RGB}{254,240,217}

\definecolor{OrRd42}{RGB}{253,204,138}

\definecolor{OrRd4E}{RGB}{253,204,138}

\definecolor{OrRd43}{RGB}{252,141,89}

\definecolor{OrRd4G}{RGB}{252,141,89}

\definecolor{OrRd44}{RGB}{215,48,31}

\definecolor{OrRd4J}{RGB}{215,48,31}

\definecolor{OrRd51}{RGB}{254,240,217}

\definecolor{OrRd5B}{RGB}{254,240,217}

\definecolor{OrRd52}{RGB}{253,204,138}

\definecolor{OrRd5E}{RGB}{253,204,138}

\definecolor{OrRd53}{RGB}{252,141,89}

\definecolor{OrRd5G}{RGB}{252,141,89}

\definecolor{OrRd54}{RGB}{227,74,51}

\definecolor{OrRd5I}{RGB}{227,74,51}

\definecolor{OrRd55}{RGB}{179,0,0}

\definecolor{OrRd5K}{RGB}{179,0,0}

\definecolor{OrRd61}{RGB}{254,240,217}

\definecolor{OrRd6B}{RGB}{254,240,217}

\definecolor{OrRd62}{RGB}{253,212,158}

\definecolor{OrRd6D}{RGB}{253,212,158}

\definecolor{OrRd63}{RGB}{253,187,132}

\definecolor{OrRd6F}{RGB}{253,187,132}

\definecolor{OrRd64}{RGB}{252,141,89}

\definecolor{OrRd6G}{RGB}{252,141,89}

\definecolor{OrRd65}{RGB}{227,74,51}

\definecolor{OrRd6I}{RGB}{227,74,51}

\definecolor{OrRd66}{RGB}{179,0,0}

\definecolor{OrRd6K}{RGB}{179,0,0}

\definecolor{OrRd71}{RGB}{254,240,217}

\definecolor{OrRd7B}{RGB}{254,240,217}

\definecolor{OrRd72}{RGB}{253,212,158}

\definecolor{OrRd7D}{RGB}{253,212,158}

\definecolor{OrRd73}{RGB}{253,187,132}

\definecolor{OrRd7F}{RGB}{253,187,132}

\definecolor{OrRd74}{RGB}{252,141,89}

\definecolor{OrRd7G}{RGB}{252,141,89}

\definecolor{OrRd75}{RGB}{239,101,72}

\definecolor{OrRd7H}{RGB}{239,101,72}

\definecolor{OrRd76}{RGB}{215,48,31}

\definecolor{OrRd7J}{RGB}{215,48,31}

\definecolor{OrRd77}{RGB}{153,0,0}

\definecolor{OrRd7L}{RGB}{153,0,0}

\definecolor{OrRd81}{RGB}{255,247,236}

\definecolor{OrRd8A}{RGB}{255,247,236}

\definecolor{OrRd82}{RGB}{254,232,200}

\definecolor{OrRd8C}{RGB}{254,232,200}

\definecolor{OrRd83}{RGB}{253,212,158}

\definecolor{OrRd8D}{RGB}{253,212,158}

\definecolor{OrRd84}{RGB}{253,187,132}

\definecolor{OrRd8F}{RGB}{253,187,132}

\definecolor{OrRd85}{RGB}{252,141,89}

\definecolor{OrRd8G}{RGB}{252,141,89}

\definecolor{OrRd86}{RGB}{239,101,72}

\definecolor{OrRd8H}{RGB}{239,101,72}

\definecolor{OrRd87}{RGB}{215,48,31}

\definecolor{OrRd8J}{RGB}{215,48,31}

\definecolor{OrRd88}{RGB}{153,0,0}

\definecolor{OrRd8L}{RGB}{153,0,0}

\definecolor{OrRd91}{RGB}{255,247,236}

\definecolor{OrRd9A}{RGB}{255,247,236}

\definecolor{OrRd92}{RGB}{254,232,200}

\definecolor{OrRd9C}{RGB}{254,232,200}

\definecolor{OrRd93}{RGB}{253,212,158}

\definecolor{OrRd9D}{RGB}{253,212,158}

\definecolor{OrRd94}{RGB}{253,187,132}

\definecolor{OrRd9F}{RGB}{253,187,132}

\definecolor{OrRd95}{RGB}{252,141,89}

\definecolor{OrRd9G}{RGB}{252,141,89}

\definecolor{OrRd96}{RGB}{239,101,72}

\definecolor{OrRd9H}{RGB}{239,101,72}

\definecolor{OrRd97}{RGB}{215,48,31}

\definecolor{OrRd9J}{RGB}{215,48,31}

\definecolor{OrRd98}{RGB}{179,0,0}

\definecolor{OrRd9K}{RGB}{179,0,0}

\definecolor{OrRd99}{RGB}{127,0,0}

\definecolor{OrRd9M}{RGB}{127,0,0}

\definecolor{YlOrRd31}{RGB}{255,237,160}

\definecolor{YlOrRd3C}{RGB}{255,237,160}

\definecolor{YlOrRd32}{RGB}{254,178,76}

\definecolor{YlOrRd3F}{RGB}{254,178,76}

\definecolor{YlOrRd33}{RGB}{240,59,32}

\definecolor{YlOrRd3I}{RGB}{240,59,32}

\definecolor{YlOrRd41}{RGB}{255,255,178}

\definecolor{YlOrRd4B}{RGB}{255,255,178}

\definecolor{YlOrRd42}{RGB}{254,204,92}

\definecolor{YlOrRd4E}{RGB}{254,204,92}

\definecolor{YlOrRd43}{RGB}{253,141,60}

\definecolor{YlOrRd4G}{RGB}{253,141,60}

\definecolor{YlOrRd44}{RGB}{227,26,28}

\definecolor{YlOrRd4J}{RGB}{227,26,28}

\definecolor{YlOrRd51}{RGB}{255,255,178}

\definecolor{YlOrRd5B}{RGB}{255,255,178}

\definecolor{YlOrRd52}{RGB}{254,204,92}

\definecolor{YlOrRd5E}{RGB}{254,204,92}

\definecolor{YlOrRd53}{RGB}{253,141,60}

\definecolor{YlOrRd5G}{RGB}{253,141,60}

\definecolor{YlOrRd54}{RGB}{240,59,32}

\definecolor{YlOrRd5I}{RGB}{240,59,32}

\definecolor{YlOrRd55}{RGB}{189,0,38}

\definecolor{YlOrRd5K}{RGB}{189,0,38}

\definecolor{YlOrRd61}{RGB}{255,255,178}

\definecolor{YlOrRd6B}{RGB}{255,255,178}

\definecolor{YlOrRd62}{RGB}{254,217,118}

\definecolor{YlOrRd6D}{RGB}{254,217,118}

\definecolor{YlOrRd63}{RGB}{254,178,76}

\definecolor{YlOrRd6F}{RGB}{254,178,76}

\definecolor{YlOrRd64}{RGB}{253,141,60}

\definecolor{YlOrRd6G}{RGB}{253,141,60}

\definecolor{YlOrRd65}{RGB}{240,59,32}

\definecolor{YlOrRd6I}{RGB}{240,59,32}

\definecolor{YlOrRd66}{RGB}{189,0,38}

\definecolor{YlOrRd6K}{RGB}{189,0,38}

\definecolor{YlOrRd71}{RGB}{255,255,178}

\definecolor{YlOrRd7B}{RGB}{255,255,178}

\definecolor{YlOrRd72}{RGB}{254,217,118}

\definecolor{YlOrRd7D}{RGB}{254,217,118}

\definecolor{YlOrRd73}{RGB}{254,178,76}

\definecolor{YlOrRd7F}{RGB}{254,178,76}

\definecolor{YlOrRd74}{RGB}{253,141,60}

\definecolor{YlOrRd7G}{RGB}{253,141,60}

\definecolor{YlOrRd75}{RGB}{252,78,42}

\definecolor{YlOrRd7H}{RGB}{252,78,42}

\definecolor{YlOrRd76}{RGB}{227,26,28}

\definecolor{YlOrRd7J}{RGB}{227,26,28}

\definecolor{YlOrRd77}{RGB}{177,0,38}

\definecolor{YlOrRd7L}{RGB}{177,0,38}

\definecolor{YlOrRd81}{RGB}{255,255,204}

\definecolor{YlOrRd8A}{RGB}{255,255,204}

\definecolor{YlOrRd82}{RGB}{255,237,160}

\definecolor{YlOrRd8C}{RGB}{255,237,160}

\definecolor{YlOrRd83}{RGB}{254,217,118}

\definecolor{YlOrRd8D}{RGB}{254,217,118}

\definecolor{YlOrRd84}{RGB}{254,178,76}

\definecolor{YlOrRd8F}{RGB}{254,178,76}

\definecolor{YlOrRd85}{RGB}{253,141,60}

\definecolor{YlOrRd8G}{RGB}{253,141,60}

\definecolor{YlOrRd86}{RGB}{252,78,42}

\definecolor{YlOrRd8H}{RGB}{252,78,42}

\definecolor{YlOrRd87}{RGB}{227,26,28}

\definecolor{YlOrRd8J}{RGB}{227,26,28}

\definecolor{YlOrRd88}{RGB}{177,0,38}

\definecolor{YlOrRd8L}{RGB}{177,0,38}

\definecolor{YlOrRd91}{RGB}{255,255,204}

\definecolor{YlOrRd9A}{RGB}{255,255,204}

\definecolor{YlOrRd92}{RGB}{255,237,160}

\definecolor{YlOrRd9C}{RGB}{255,237,160}

\definecolor{YlOrRd93}{RGB}{254,217,118}

\definecolor{YlOrRd9D}{RGB}{254,217,118}

\definecolor{YlOrRd94}{RGB}{254,178,76}

\definecolor{YlOrRd9F}{RGB}{254,178,76}

\definecolor{YlOrRd95}{RGB}{253,141,60}

\definecolor{YlOrRd9G}{RGB}{253,141,60}

\definecolor{YlOrRd96}{RGB}{252,78,42}

\definecolor{YlOrRd9H}{RGB}{252,78,42}

\definecolor{YlOrRd97}{RGB}{227,26,28}

\definecolor{YlOrRd9J}{RGB}{227,26,28}

\definecolor{YlOrRd98}{RGB}{189,0,38}

\definecolor{YlOrRd9K}{RGB}{189,0,38}

\definecolor{YlOrRd99}{RGB}{128,0,38}

\definecolor{YlOrRd9M}{RGB}{128,0,38}

\definecolor{YlOrBr31}{RGB}{255,247,188}

\definecolor{YlOrBr3C}{RGB}{255,247,188}

\definecolor{YlOrBr32}{RGB}{254,196,79}

\definecolor{YlOrBr3F}{RGB}{254,196,79}

\definecolor{YlOrBr33}{RGB}{217,95,14}

\definecolor{YlOrBr3I}{RGB}{217,95,14}

\definecolor{YlOrBr41}{RGB}{255,255,212}

\definecolor{YlOrBr4B}{RGB}{255,255,212}

\definecolor{YlOrBr42}{RGB}{254,217,142}

\definecolor{YlOrBr4E}{RGB}{254,217,142}

\definecolor{YlOrBr43}{RGB}{254,153,41}

\definecolor{YlOrBr4G}{RGB}{254,153,41}

\definecolor{YlOrBr44}{RGB}{204,76,2}

\definecolor{YlOrBr4J}{RGB}{204,76,2}

\definecolor{YlOrBr51}{RGB}{255,255,212}

\definecolor{YlOrBr5B}{RGB}{255,255,212}

\definecolor{YlOrBr52}{RGB}{254,217,142}

\definecolor{YlOrBr5E}{RGB}{254,217,142}

\definecolor{YlOrBr53}{RGB}{254,153,41}

\definecolor{YlOrBr5G}{RGB}{254,153,41}

\definecolor{YlOrBr54}{RGB}{217,95,14}

\definecolor{YlOrBr5I}{RGB}{217,95,14}

\definecolor{YlOrBr55}{RGB}{153,52,4}

\definecolor{YlOrBr5K}{RGB}{153,52,4}

\definecolor{YlOrBr61}{RGB}{255,255,212}

\definecolor{YlOrBr6B}{RGB}{255,255,212}

\definecolor{YlOrBr62}{RGB}{254,227,145}

\definecolor{YlOrBr6D}{RGB}{254,227,145}

\definecolor{YlOrBr63}{RGB}{254,196,79}

\definecolor{YlOrBr6F}{RGB}{254,196,79}

\definecolor{YlOrBr64}{RGB}{254,153,41}

\definecolor{YlOrBr6G}{RGB}{254,153,41}

\definecolor{YlOrBr65}{RGB}{217,95,14}

\definecolor{YlOrBr6I}{RGB}{217,95,14}

\definecolor{YlOrBr66}{RGB}{153,52,4}

\definecolor{YlOrBr6K}{RGB}{153,52,4}

\definecolor{YlOrBr71}{RGB}{255,255,212}

\definecolor{YlOrBr7B}{RGB}{255,255,212}

\definecolor{YlOrBr72}{RGB}{254,227,145}

\definecolor{YlOrBr7D}{RGB}{254,227,145}

\definecolor{YlOrBr73}{RGB}{254,196,79}

\definecolor{YlOrBr7F}{RGB}{254,196,79}

\definecolor{YlOrBr74}{RGB}{254,153,41}

\definecolor{YlOrBr7G}{RGB}{254,153,41}

\definecolor{YlOrBr75}{RGB}{236,112,20}

\definecolor{YlOrBr7H}{RGB}{236,112,20}

\definecolor{YlOrBr76}{RGB}{204,76,2}

\definecolor{YlOrBr7J}{RGB}{204,76,2}

\definecolor{YlOrBr77}{RGB}{140,45,4}

\definecolor{YlOrBr7L}{RGB}{140,45,4}

\definecolor{YlOrBr81}{RGB}{255,255,229}

\definecolor{YlOrBr8A}{RGB}{255,255,229}

\definecolor{YlOrBr82}{RGB}{255,247,188}

\definecolor{YlOrBr8C}{RGB}{255,247,188}

\definecolor{YlOrBr83}{RGB}{254,227,145}

\definecolor{YlOrBr8D}{RGB}{254,227,145}

\definecolor{YlOrBr84}{RGB}{254,196,79}

\definecolor{YlOrBr8F}{RGB}{254,196,79}

\definecolor{YlOrBr85}{RGB}{254,153,41}

\definecolor{YlOrBr8G}{RGB}{254,153,41}

\definecolor{YlOrBr86}{RGB}{236,112,20}

\definecolor{YlOrBr8H}{RGB}{236,112,20}

\definecolor{YlOrBr87}{RGB}{204,76,2}

\definecolor{YlOrBr8J}{RGB}{204,76,2}

\definecolor{YlOrBr88}{RGB}{140,45,4}

\definecolor{YlOrBr8L}{RGB}{140,45,4}

\definecolor{YlOrBr91}{RGB}{255,255,229}

\definecolor{YlOrBr9A}{RGB}{255,255,229}

\definecolor{YlOrBr92}{RGB}{255,247,188}

\definecolor{YlOrBr9C}{RGB}{255,247,188}

\definecolor{YlOrBr93}{RGB}{254,227,145}

\definecolor{YlOrBr9D}{RGB}{254,227,145}

\definecolor{YlOrBr94}{RGB}{254,196,79}

\definecolor{YlOrBr9F}{RGB}{254,196,79}

\definecolor{YlOrBr95}{RGB}{254,153,41}

\definecolor{YlOrBr9G}{RGB}{254,153,41}

\definecolor{YlOrBr96}{RGB}{236,112,20}

\definecolor{YlOrBr9H}{RGB}{236,112,20}

\definecolor{YlOrBr97}{RGB}{204,76,2}

\definecolor{YlOrBr9J}{RGB}{204,76,2}

\definecolor{YlOrBr98}{RGB}{153,52,4}

\definecolor{YlOrBr9K}{RGB}{153,52,4}

\definecolor{YlOrBr99}{RGB}{102,37,6}

\definecolor{YlOrBr9M}{RGB}{102,37,6}

\definecolor{Purples31}{RGB}{239,237,245}

\definecolor{Purples3C}{RGB}{239,237,245}

\definecolor{Purples32}{RGB}{188,189,220}

\definecolor{Purples3F}{RGB}{188,189,220}

\definecolor{Purples33}{RGB}{117,107,177}

\definecolor{Purples3I}{RGB}{117,107,177}

\definecolor{Purples41}{RGB}{242,240,247}

\definecolor{Purples4B}{RGB}{242,240,247}

\definecolor{Purples42}{RGB}{203,201,226}

\definecolor{Purples4E}{RGB}{203,201,226}

\definecolor{Purples43}{RGB}{158,154,200}

\definecolor{Purples4G}{RGB}{158,154,200}

\definecolor{Purples44}{RGB}{106,81,163}

\definecolor{Purples4J}{RGB}{106,81,163}

\definecolor{Purples51}{RGB}{242,240,247}

\definecolor{Purples5B}{RGB}{242,240,247}

\definecolor{Purples52}{RGB}{203,201,226}

\definecolor{Purples5E}{RGB}{203,201,226}

\definecolor{Purples53}{RGB}{158,154,200}

\definecolor{Purples5G}{RGB}{158,154,200}

\definecolor{Purples54}{RGB}{117,107,177}

\definecolor{Purples5I}{RGB}{117,107,177}

\definecolor{Purples55}{RGB}{84,39,143}

\definecolor{Purples5K}{RGB}{84,39,143}

\definecolor{Purples61}{RGB}{242,240,247}

\definecolor{Purples6B}{RGB}{242,240,247}

\definecolor{Purples62}{RGB}{218,218,235}

\definecolor{Purples6D}{RGB}{218,218,235}

\definecolor{Purples63}{RGB}{188,189,220}

\definecolor{Purples6F}{RGB}{188,189,220}

\definecolor{Purples64}{RGB}{158,154,200}

\definecolor{Purples6G}{RGB}{158,154,200}

\definecolor{Purples65}{RGB}{117,107,177}

\definecolor{Purples6I}{RGB}{117,107,177}

\definecolor{Purples66}{RGB}{84,39,143}

\definecolor{Purples6K}{RGB}{84,39,143}

\definecolor{Purples71}{RGB}{242,240,247}

\definecolor{Purples7B}{RGB}{242,240,247}

\definecolor{Purples72}{RGB}{218,218,235}

\definecolor{Purples7D}{RGB}{218,218,235}

\definecolor{Purples73}{RGB}{188,189,220}

\definecolor{Purples7F}{RGB}{188,189,220}

\definecolor{Purples74}{RGB}{158,154,200}

\definecolor{Purples7G}{RGB}{158,154,200}

\definecolor{Purples75}{RGB}{128,125,186}

\definecolor{Purples7H}{RGB}{128,125,186}

\definecolor{Purples76}{RGB}{106,81,163}

\definecolor{Purples7J}{RGB}{106,81,163}

\definecolor{Purples77}{RGB}{74,20,134}

\definecolor{Purples7L}{RGB}{74,20,134}

\definecolor{Purples81}{RGB}{252,251,253}

\definecolor{Purples8A}{RGB}{252,251,253}

\definecolor{Purples82}{RGB}{239,237,245}

\definecolor{Purples8C}{RGB}{239,237,245}

\definecolor{Purples83}{RGB}{218,218,235}

\definecolor{Purples8D}{RGB}{218,218,235}

\definecolor{Purples84}{RGB}{188,189,220}

\definecolor{Purples8F}{RGB}{188,189,220}

\definecolor{Purples85}{RGB}{158,154,200}

\definecolor{Purples8G}{RGB}{158,154,200}

\definecolor{Purples86}{RGB}{128,125,186}

\definecolor{Purples8H}{RGB}{128,125,186}

\definecolor{Purples87}{RGB}{106,81,163}

\definecolor{Purples8J}{RGB}{106,81,163}

\definecolor{Purples88}{RGB}{74,20,134}

\definecolor{Purples8L}{RGB}{74,20,134}

\definecolor{Purples91}{RGB}{252,251,253}

\definecolor{Purples9A}{RGB}{252,251,253}

\definecolor{Purples92}{RGB}{239,237,245}

\definecolor{Purples9C}{RGB}{239,237,245}

\definecolor{Purples93}{RGB}{218,218,235}

\definecolor{Purples9D}{RGB}{218,218,235}

\definecolor{Purples94}{RGB}{188,189,220}

\definecolor{Purples9F}{RGB}{188,189,220}

\definecolor{Purples95}{RGB}{158,154,200}

\definecolor{Purples9G}{RGB}{158,154,200}

\definecolor{Purples96}{RGB}{128,125,186}

\definecolor{Purples9H}{RGB}{128,125,186}

\definecolor{Purples97}{RGB}{106,81,163}

\definecolor{Purples9J}{RGB}{106,81,163}

\definecolor{Purples98}{RGB}{84,39,143}

\definecolor{Purples9K}{RGB}{84,39,143}

\definecolor{Purples99}{RGB}{63,0,125}

\definecolor{Purples9M}{RGB}{63,0,125}

\definecolor{Blues31}{RGB}{222,235,247}

\definecolor{Blues3C}{RGB}{222,235,247}

\definecolor{Blues32}{RGB}{158,202,225}

\definecolor{Blues3F}{RGB}{158,202,225}

\definecolor{Blues33}{RGB}{49,130,189}

\definecolor{Blues3I}{RGB}{49,130,189}

\definecolor{Blues41}{RGB}{239,243,255}

\definecolor{Blues4B}{RGB}{239,243,255}

\definecolor{Blues42}{RGB}{189,215,231}

\definecolor{Blues4E}{RGB}{189,215,231}

\definecolor{Blues43}{RGB}{107,174,214}

\definecolor{Blues4G}{RGB}{107,174,214}

\definecolor{Blues44}{RGB}{33,113,181}

\definecolor{Blues4J}{RGB}{33,113,181}

\definecolor{Blues51}{RGB}{239,243,255}

\definecolor{Blues5B}{RGB}{239,243,255}

\definecolor{Blues52}{RGB}{189,215,231}

\definecolor{Blues5E}{RGB}{189,215,231}

\definecolor{Blues53}{RGB}{107,174,214}

\definecolor{Blues5G}{RGB}{107,174,214}

\definecolor{Blues54}{RGB}{49,130,189}

\definecolor{Blues5I}{RGB}{49,130,189}

\definecolor{Blues55}{RGB}{8,81,156}

\definecolor{Blues5K}{RGB}{8,81,156}

\definecolor{Blues61}{RGB}{239,243,255}

\definecolor{Blues6B}{RGB}{239,243,255}

\definecolor{Blues62}{RGB}{198,219,239}

\definecolor{Blues6D}{RGB}{198,219,239}

\definecolor{Blues63}{RGB}{158,202,225}

\definecolor{Blues6F}{RGB}{158,202,225}

\definecolor{Blues64}{RGB}{107,174,214}

\definecolor{Blues6G}{RGB}{107,174,214}

\definecolor{Blues65}{RGB}{49,130,189}

\definecolor{Blues6I}{RGB}{49,130,189}

\definecolor{Blues66}{RGB}{8,81,156}

\definecolor{Blues6K}{RGB}{8,81,156}

\definecolor{Blues71}{RGB}{239,243,255}

\definecolor{Blues7B}{RGB}{239,243,255}

\definecolor{Blues72}{RGB}{198,219,239}

\definecolor{Blues7D}{RGB}{198,219,239}

\definecolor{Blues73}{RGB}{158,202,225}

\definecolor{Blues7F}{RGB}{158,202,225}

\definecolor{Blues74}{RGB}{107,174,214}

\definecolor{Blues7G}{RGB}{107,174,214}

\definecolor{Blues75}{RGB}{66,146,198}

\definecolor{Blues7H}{RGB}{66,146,198}

\definecolor{Blues76}{RGB}{33,113,181}

\definecolor{Blues7J}{RGB}{33,113,181}

\definecolor{Blues77}{RGB}{8,69,148}

\definecolor{Blues7L}{RGB}{8,69,148}

\definecolor{Blues81}{RGB}{247,251,255}

\definecolor{Blues8A}{RGB}{247,251,255}

\definecolor{Blues82}{RGB}{222,235,247}

\definecolor{Blues8C}{RGB}{222,235,247}

\definecolor{Blues83}{RGB}{198,219,239}

\definecolor{Blues8D}{RGB}{198,219,239}

\definecolor{Blues84}{RGB}{158,202,225}

\definecolor{Blues8F}{RGB}{158,202,225}

\definecolor{Blues85}{RGB}{107,174,214}

\definecolor{Blues8G}{RGB}{107,174,214}

\definecolor{Blues86}{RGB}{66,146,198}

\definecolor{Blues8H}{RGB}{66,146,198}

\definecolor{Blues87}{RGB}{33,113,181}

\definecolor{Blues8J}{RGB}{33,113,181}

\definecolor{Blues88}{RGB}{8,69,148}

\definecolor{Blues8L}{RGB}{8,69,148}

\definecolor{Blues91}{RGB}{247,251,255}

\definecolor{Blues9A}{RGB}{247,251,255}

\definecolor{Blues92}{RGB}{222,235,247}

\definecolor{Blues9C}{RGB}{222,235,247}

\definecolor{Blues93}{RGB}{198,219,239}

\definecolor{Blues9D}{RGB}{198,219,239}

\definecolor{Blues94}{RGB}{158,202,225}

\definecolor{Blues9F}{RGB}{158,202,225}

\definecolor{Blues95}{RGB}{107,174,214}

\definecolor{Blues9G}{RGB}{107,174,214}

\definecolor{Blues96}{RGB}{66,146,198}

\definecolor{Blues9H}{RGB}{66,146,198}

\definecolor{Blues97}{RGB}{33,113,181}

\definecolor{Blues9J}{RGB}{33,113,181}

\definecolor{Blues98}{RGB}{8,81,156}

\definecolor{Blues9K}{RGB}{8,81,156}

\definecolor{Blues99}{RGB}{8,48,107}

\definecolor{Blues9M}{RGB}{8,48,107}

\definecolor{Greens31}{RGB}{229,245,224}

\definecolor{Greens3C}{RGB}{229,245,224}

\definecolor{Greens32}{RGB}{161,217,155}

\definecolor{Greens3F}{RGB}{161,217,155}

\definecolor{Greens33}{RGB}{49,163,84}

\definecolor{Greens3I}{RGB}{49,163,84}

\definecolor{Greens41}{RGB}{237,248,233}

\definecolor{Greens4B}{RGB}{237,248,233}

\definecolor{Greens42}{RGB}{186,228,179}

\definecolor{Greens4E}{RGB}{186,228,179}

\definecolor{Greens43}{RGB}{116,196,118}

\definecolor{Greens4G}{RGB}{116,196,118}

\definecolor{Greens44}{RGB}{35,139,69}

\definecolor{Greens4J}{RGB}{35,139,69}

\definecolor{Greens51}{RGB}{237,248,233}

\definecolor{Greens5B}{RGB}{237,248,233}

\definecolor{Greens52}{RGB}{186,228,179}

\definecolor{Greens5E}{RGB}{186,228,179}

\definecolor{Greens53}{RGB}{116,196,118}

\definecolor{Greens5G}{RGB}{116,196,118}

\definecolor{Greens54}{RGB}{49,163,84}

\definecolor{Greens5I}{RGB}{49,163,84}

\definecolor{Greens55}{RGB}{0,109,44}

\definecolor{Greens5K}{RGB}{0,109,44}

\definecolor{Greens61}{RGB}{237,248,233}

\definecolor{Greens6B}{RGB}{237,248,233}

\definecolor{Greens62}{RGB}{199,233,192}

\definecolor{Greens6D}{RGB}{199,233,192}

\definecolor{Greens63}{RGB}{161,217,155}

\definecolor{Greens6F}{RGB}{161,217,155}

\definecolor{Greens64}{RGB}{116,196,118}

\definecolor{Greens6G}{RGB}{116,196,118}

\definecolor{Greens65}{RGB}{49,163,84}

\definecolor{Greens6I}{RGB}{49,163,84}

\definecolor{Greens66}{RGB}{0,109,44}

\definecolor{Greens6K}{RGB}{0,109,44}

\definecolor{Greens71}{RGB}{237,248,233}

\definecolor{Greens7B}{RGB}{237,248,233}

\definecolor{Greens72}{RGB}{199,233,192}

\definecolor{Greens7D}{RGB}{199,233,192}

\definecolor{Greens73}{RGB}{161,217,155}

\definecolor{Greens7F}{RGB}{161,217,155}

\definecolor{Greens74}{RGB}{116,196,118}

\definecolor{Greens7G}{RGB}{116,196,118}

\definecolor{Greens75}{RGB}{65,171,93}

\definecolor{Greens7H}{RGB}{65,171,93}

\definecolor{Greens76}{RGB}{35,139,69}

\definecolor{Greens7J}{RGB}{35,139,69}

\definecolor{Greens77}{RGB}{0,90,50}

\definecolor{Greens7L}{RGB}{0,90,50}

\definecolor{Greens81}{RGB}{247,252,245}

\definecolor{Greens8A}{RGB}{247,252,245}

\definecolor{Greens82}{RGB}{229,245,224}

\definecolor{Greens8C}{RGB}{229,245,224}

\definecolor{Greens83}{RGB}{199,233,192}

\definecolor{Greens8D}{RGB}{199,233,192}

\definecolor{Greens84}{RGB}{161,217,155}

\definecolor{Greens8F}{RGB}{161,217,155}

\definecolor{Greens85}{RGB}{116,196,118}

\definecolor{Greens8G}{RGB}{116,196,118}

\definecolor{Greens86}{RGB}{65,171,93}

\definecolor{Greens8H}{RGB}{65,171,93}

\definecolor{Greens87}{RGB}{35,139,69}

\definecolor{Greens8J}{RGB}{35,139,69}

\definecolor{Greens88}{RGB}{0,90,50}

\definecolor{Greens8L}{RGB}{0,90,50}

\definecolor{Greens91}{RGB}{247,252,245}

\definecolor{Greens9A}{RGB}{247,252,245}

\definecolor{Greens92}{RGB}{229,245,224}

\definecolor{Greens9C}{RGB}{229,245,224}

\definecolor{Greens93}{RGB}{199,233,192}

\definecolor{Greens9D}{RGB}{199,233,192}

\definecolor{Greens94}{RGB}{161,217,155}

\definecolor{Greens9F}{RGB}{161,217,155}

\definecolor{Greens95}{RGB}{116,196,118}

\definecolor{Greens9G}{RGB}{116,196,118}

\definecolor{Greens96}{RGB}{65,171,93}

\definecolor{Greens9H}{RGB}{65,171,93}

\definecolor{Greens97}{RGB}{35,139,69}

\definecolor{Greens9J}{RGB}{35,139,69}

\definecolor{Greens98}{RGB}{0,109,44}

\definecolor{Greens9K}{RGB}{0,109,44}

\definecolor{Greens99}{RGB}{0,68,27}

\definecolor{Greens9M}{RGB}{0,68,27}

\definecolor{Oranges31}{RGB}{254,230,206}

\definecolor{Oranges3C}{RGB}{254,230,206}

\definecolor{Oranges32}{RGB}{253,174,107}

\definecolor{Oranges3F}{RGB}{253,174,107}

\definecolor{Oranges33}{RGB}{230,85,13}

\definecolor{Oranges3I}{RGB}{230,85,13}

\definecolor{Oranges41}{RGB}{254,237,222}

\definecolor{Oranges4B}{RGB}{254,237,222}

\definecolor{Oranges42}{RGB}{253,190,133}

\definecolor{Oranges4E}{RGB}{253,190,133}

\definecolor{Oranges43}{RGB}{253,141,60}

\definecolor{Oranges4G}{RGB}{253,141,60}

\definecolor{Oranges44}{RGB}{217,71,1}

\definecolor{Oranges4J}{RGB}{217,71,1}

\definecolor{Oranges51}{RGB}{254,237,222}

\definecolor{Oranges5B}{RGB}{254,237,222}

\definecolor{Oranges52}{RGB}{253,190,133}

\definecolor{Oranges5E}{RGB}{253,190,133}

\definecolor{Oranges53}{RGB}{253,141,60}

\definecolor{Oranges5G}{RGB}{253,141,60}

\definecolor{Oranges54}{RGB}{230,85,13}

\definecolor{Oranges5I}{RGB}{230,85,13}

\definecolor{Oranges55}{RGB}{166,54,3}

\definecolor{Oranges5K}{RGB}{166,54,3}

\definecolor{Oranges61}{RGB}{254,237,222}

\definecolor{Oranges6B}{RGB}{254,237,222}

\definecolor{Oranges62}{RGB}{253,208,162}

\definecolor{Oranges6D}{RGB}{253,208,162}

\definecolor{Oranges63}{RGB}{253,174,107}

\definecolor{Oranges6F}{RGB}{253,174,107}

\definecolor{Oranges64}{RGB}{253,141,60}

\definecolor{Oranges6G}{RGB}{253,141,60}

\definecolor{Oranges65}{RGB}{230,85,13}

\definecolor{Oranges6I}{RGB}{230,85,13}

\definecolor{Oranges66}{RGB}{166,54,3}

\definecolor{Oranges6K}{RGB}{166,54,3}

\definecolor{Oranges71}{RGB}{254,237,222}

\definecolor{Oranges7B}{RGB}{254,237,222}

\definecolor{Oranges72}{RGB}{253,208,162}

\definecolor{Oranges7D}{RGB}{253,208,162}

\definecolor{Oranges73}{RGB}{253,174,107}

\definecolor{Oranges7F}{RGB}{253,174,107}

\definecolor{Oranges74}{RGB}{253,141,60}

\definecolor{Oranges7G}{RGB}{253,141,60}

\definecolor{Oranges75}{RGB}{241,105,19}

\definecolor{Oranges7H}{RGB}{241,105,19}

\definecolor{Oranges76}{RGB}{217,72,1}

\definecolor{Oranges7J}{RGB}{217,72,1}

\definecolor{Oranges77}{RGB}{140,45,4}

\definecolor{Oranges7L}{RGB}{140,45,4}

\definecolor{Oranges81}{RGB}{255,245,235}

\definecolor{Oranges8A}{RGB}{255,245,235}

\definecolor{Oranges82}{RGB}{254,230,206}

\definecolor{Oranges8C}{RGB}{254,230,206}

\definecolor{Oranges83}{RGB}{253,208,162}

\definecolor{Oranges8D}{RGB}{253,208,162}

\definecolor{Oranges84}{RGB}{253,174,107}

\definecolor{Oranges8F}{RGB}{253,174,107}

\definecolor{Oranges85}{RGB}{253,141,60}

\definecolor{Oranges8G}{RGB}{253,141,60}

\definecolor{Oranges86}{RGB}{241,105,19}

\definecolor{Oranges8H}{RGB}{241,105,19}

\definecolor{Oranges87}{RGB}{217,72,1}

\definecolor{Oranges8J}{RGB}{217,72,1}

\definecolor{Oranges88}{RGB}{140,45,4}

\definecolor{Oranges8L}{RGB}{140,45,4}

\definecolor{Oranges91}{RGB}{255,245,235}

\definecolor{Oranges9A}{RGB}{255,245,235}

\definecolor{Oranges92}{RGB}{254,230,206}

\definecolor{Oranges9C}{RGB}{254,230,206}

\definecolor{Oranges93}{RGB}{253,208,162}

\definecolor{Oranges9D}{RGB}{253,208,162}

\definecolor{Oranges94}{RGB}{253,174,107}

\definecolor{Oranges9F}{RGB}{253,174,107}

\definecolor{Oranges95}{RGB}{253,141,60}

\definecolor{Oranges9G}{RGB}{253,141,60}

\definecolor{Oranges96}{RGB}{241,105,19}

\definecolor{Oranges9H}{RGB}{241,105,19}

\definecolor{Oranges97}{RGB}{217,72,1}

\definecolor{Oranges9J}{RGB}{217,72,1}

\definecolor{Oranges98}{RGB}{166,54,3}

\definecolor{Oranges9K}{RGB}{166,54,3}

\definecolor{Oranges99}{RGB}{127,39,4}

\definecolor{Oranges9M}{RGB}{127,39,4}

\definecolor{Reds31}{RGB}{254,224,210}

\definecolor{Reds3C}{RGB}{254,224,210}

\definecolor{Reds32}{RGB}{252,146,114}

\definecolor{Reds3F}{RGB}{252,146,114}

\definecolor{Reds33}{RGB}{222,45,38}

\definecolor{Reds3I}{RGB}{222,45,38}

\definecolor{Reds41}{RGB}{254,229,217}

\definecolor{Reds4B}{RGB}{254,229,217}

\definecolor{Reds42}{RGB}{252,174,145}

\definecolor{Reds4E}{RGB}{252,174,145}

\definecolor{Reds43}{RGB}{251,106,74}

\definecolor{Reds4G}{RGB}{251,106,74}

\definecolor{Reds44}{RGB}{203,24,29}

\definecolor{Reds4J}{RGB}{203,24,29}

\definecolor{Reds51}{RGB}{254,229,217}

\definecolor{Reds5B}{RGB}{254,229,217}

\definecolor{Reds52}{RGB}{252,174,145}

\definecolor{Reds5E}{RGB}{252,174,145}

\definecolor{Reds53}{RGB}{251,106,74}

\definecolor{Reds5G}{RGB}{251,106,74}

\definecolor{Reds54}{RGB}{222,45,38}

\definecolor{Reds5I}{RGB}{222,45,38}

\definecolor{Reds55}{RGB}{165,15,21}

\definecolor{Reds5K}{RGB}{165,15,21}

\definecolor{Reds61}{RGB}{254,229,217}

\definecolor{Reds6B}{RGB}{254,229,217}

\definecolor{Reds62}{RGB}{252,187,161}

\definecolor{Reds6D}{RGB}{252,187,161}

\definecolor{Reds63}{RGB}{252,146,114}

\definecolor{Reds6F}{RGB}{252,146,114}

\definecolor{Reds64}{RGB}{251,106,74}

\definecolor{Reds6G}{RGB}{251,106,74}

\definecolor{Reds65}{RGB}{222,45,38}

\definecolor{Reds6I}{RGB}{222,45,38}

\definecolor{Reds66}{RGB}{165,15,21}

\definecolor{Reds6K}{RGB}{165,15,21}

\definecolor{Reds71}{RGB}{254,229,217}

\definecolor{Reds7B}{RGB}{254,229,217}

\definecolor{Reds72}{RGB}{252,187,161}

\definecolor{Reds7D}{RGB}{252,187,161}

\definecolor{Reds73}{RGB}{252,146,114}

\definecolor{Reds7F}{RGB}{252,146,114}

\definecolor{Reds74}{RGB}{251,106,74}

\definecolor{Reds7G}{RGB}{251,106,74}

\definecolor{Reds75}{RGB}{239,59,44}

\definecolor{Reds7H}{RGB}{239,59,44}

\definecolor{Reds76}{RGB}{203,24,29}

\definecolor{Reds7J}{RGB}{203,24,29}

\definecolor{Reds77}{RGB}{153,0,13}

\definecolor{Reds7L}{RGB}{153,0,13}

\definecolor{Reds81}{RGB}{255,245,240}

\definecolor{Reds8A}{RGB}{255,245,240}

\definecolor{Reds82}{RGB}{254,224,210}

\definecolor{Reds8C}{RGB}{254,224,210}

\definecolor{Reds83}{RGB}{252,187,161}

\definecolor{Reds8D}{RGB}{252,187,161}

\definecolor{Reds84}{RGB}{252,146,114}

\definecolor{Reds8F}{RGB}{252,146,114}

\definecolor{Reds85}{RGB}{251,106,74}

\definecolor{Reds8G}{RGB}{251,106,74}

\definecolor{Reds86}{RGB}{239,59,44}

\definecolor{Reds8H}{RGB}{239,59,44}

\definecolor{Reds87}{RGB}{203,24,29}

\definecolor{Reds8J}{RGB}{203,24,29}

\definecolor{Reds88}{RGB}{153,0,13}

\definecolor{Reds8L}{RGB}{153,0,13}

\definecolor{Reds91}{RGB}{255,245,240}

\definecolor{Reds9A}{RGB}{255,245,240}

\definecolor{Reds92}{RGB}{254,224,210}

\definecolor{Reds9C}{RGB}{254,224,210}

\definecolor{Reds93}{RGB}{252,187,161}

\definecolor{Reds9D}{RGB}{252,187,161}

\definecolor{Reds94}{RGB}{252,146,114}

\definecolor{Reds9F}{RGB}{252,146,114}

\definecolor{Reds95}{RGB}{251,106,74}

\definecolor{Reds9G}{RGB}{251,106,74}

\definecolor{Reds96}{RGB}{239,59,44}

\definecolor{Reds9H}{RGB}{239,59,44}

\definecolor{Reds97}{RGB}{203,24,29}

\definecolor{Reds9J}{RGB}{203,24,29}

\definecolor{Reds98}{RGB}{165,15,21}

\definecolor{Reds9K}{RGB}{165,15,21}

\definecolor{Reds99}{RGB}{103,0,13}

\definecolor{Reds9M}{RGB}{103,0,13}

\definecolor{Greys31}{RGB}{240,240,240}

\definecolor{Greys3C}{RGB}{240,240,240}

\definecolor{Greys32}{RGB}{189,189,189}

\definecolor{Greys3F}{RGB}{189,189,189}

\definecolor{Greys33}{RGB}{99,99,99}

\definecolor{Greys3I}{RGB}{99,99,99}

\definecolor{Greys41}{RGB}{247,247,247}

\definecolor{Greys4B}{RGB}{247,247,247}

\definecolor{Greys42}{RGB}{204,204,204}

\definecolor{Greys4E}{RGB}{204,204,204}

\definecolor{Greys43}{RGB}{150,150,150}

\definecolor{Greys4G}{RGB}{150,150,150}

\definecolor{Greys44}{RGB}{82,82,82}

\definecolor{Greys4J}{RGB}{82,82,82}

\definecolor{Greys51}{RGB}{247,247,247}

\definecolor{Greys5B}{RGB}{247,247,247}

\definecolor{Greys52}{RGB}{204,204,204}

\definecolor{Greys5E}{RGB}{204,204,204}

\definecolor{Greys53}{RGB}{150,150,150}

\definecolor{Greys5G}{RGB}{150,150,150}

\definecolor{Greys54}{RGB}{99,99,99}

\definecolor{Greys5I}{RGB}{99,99,99}

\definecolor{Greys55}{RGB}{37,37,37}

\definecolor{Greys5K}{RGB}{37,37,37}

\definecolor{Greys61}{RGB}{247,247,247}

\definecolor{Greys6B}{RGB}{247,247,247}

\definecolor{Greys62}{RGB}{217,217,217}

\definecolor{Greys6D}{RGB}{217,217,217}

\definecolor{Greys63}{RGB}{189,189,189}

\definecolor{Greys6F}{RGB}{189,189,189}

\definecolor{Greys64}{RGB}{150,150,150}

\definecolor{Greys6G}{RGB}{150,150,150}

\definecolor{Greys65}{RGB}{99,99,99}

\definecolor{Greys6I}{RGB}{99,99,99}

\definecolor{Greys66}{RGB}{37,37,37}

\definecolor{Greys6K}{RGB}{37,37,37}

\definecolor{Greys71}{RGB}{247,247,247}

\definecolor{Greys7B}{RGB}{247,247,247}

\definecolor{Greys72}{RGB}{217,217,217}

\definecolor{Greys7D}{RGB}{217,217,217}

\definecolor{Greys73}{RGB}{189,189,189}

\definecolor{Greys7F}{RGB}{189,189,189}

\definecolor{Greys74}{RGB}{150,150,150}

\definecolor{Greys7G}{RGB}{150,150,150}

\definecolor{Greys75}{RGB}{115,115,115}

\definecolor{Greys7H}{RGB}{115,115,115}

\definecolor{Greys76}{RGB}{82,82,82}

\definecolor{Greys7J}{RGB}{82,82,82}

\definecolor{Greys77}{RGB}{37,37,37}

\definecolor{Greys7L}{RGB}{37,37,37}

\definecolor{Greys81}{RGB}{255,255,255}

\definecolor{Greys8A}{RGB}{255,255,255}

\definecolor{Greys82}{RGB}{240,240,240}

\definecolor{Greys8C}{RGB}{240,240,240}

\definecolor{Greys83}{RGB}{217,217,217}

\definecolor{Greys8D}{RGB}{217,217,217}

\definecolor{Greys84}{RGB}{189,189,189}

\definecolor{Greys8F}{RGB}{189,189,189}

\definecolor{Greys85}{RGB}{150,150,150}

\definecolor{Greys8G}{RGB}{150,150,150}

\definecolor{Greys86}{RGB}{115,115,115}

\definecolor{Greys8H}{RGB}{115,115,115}

\definecolor{Greys87}{RGB}{82,82,82}

\definecolor{Greys8J}{RGB}{82,82,82}

\definecolor{Greys88}{RGB}{37,37,37}

\definecolor{Greys8L}{RGB}{37,37,37}

\definecolor{Greys91}{RGB}{255,255,255}

\definecolor{Greys9A}{RGB}{255,255,255}

\definecolor{Greys92}{RGB}{240,240,240}

\definecolor{Greys9C}{RGB}{240,240,240}

\definecolor{Greys93}{RGB}{217,217,217}

\definecolor{Greys9D}{RGB}{217,217,217}

\definecolor{Greys94}{RGB}{189,189,189}

\definecolor{Greys9F}{RGB}{189,189,189}

\definecolor{Greys95}{RGB}{150,150,150}

\definecolor{Greys9G}{RGB}{150,150,150}

\definecolor{Greys96}{RGB}{115,115,115}

\definecolor{Greys9H}{RGB}{115,115,115}

\definecolor{Greys97}{RGB}{82,82,82}

\definecolor{Greys9J}{RGB}{82,82,82}

\definecolor{Greys98}{RGB}{37,37,37}

\definecolor{Greys9K}{RGB}{37,37,37}

\definecolor{Greys99}{RGB}{0,0,0}

\definecolor{Greys9M}{RGB}{0,0,0}

\definecolor{PuOr31}{RGB}{241,163,64}

\definecolor{PuOr3E}{RGB}{241,163,64}

\definecolor{PuOr32}{RGB}{247,247,247}

\definecolor{PuOr3H}{RGB}{247,247,247}

\definecolor{PuOr33}{RGB}{153,142,195}

\definecolor{PuOr3K}{RGB}{153,142,195}

\definecolor{PuOr41}{RGB}{230,97,1}

\definecolor{PuOr4C}{RGB}{230,97,1}

\definecolor{PuOr42}{RGB}{253,184,99}

\definecolor{PuOr4F}{RGB}{253,184,99}

\definecolor{PuOr43}{RGB}{178,171,210}

\definecolor{PuOr4J}{RGB}{178,171,210}

\definecolor{PuOr44}{RGB}{94,60,153}

\definecolor{PuOr4M}{RGB}{94,60,153}

\definecolor{PuOr51}{RGB}{230,97,1}

\definecolor{PuOr5C}{RGB}{230,97,1}

\definecolor{PuOr52}{RGB}{253,184,99}

\definecolor{PuOr5F}{RGB}{253,184,99}

\definecolor{PuOr53}{RGB}{247,247,247}

\definecolor{PuOr5H}{RGB}{247,247,247}

\definecolor{PuOr54}{RGB}{178,171,210}

\definecolor{PuOr5J}{RGB}{178,171,210}

\definecolor{PuOr55}{RGB}{94,60,153}

\definecolor{PuOr5M}{RGB}{94,60,153}

\definecolor{PuOr61}{RGB}{179,88,6}

\definecolor{PuOr6B}{RGB}{179,88,6}

\definecolor{PuOr62}{RGB}{241,163,64}

\definecolor{PuOr6E}{RGB}{241,163,64}

\definecolor{PuOr63}{RGB}{254,224,182}

\definecolor{PuOr6G}{RGB}{254,224,182}

\definecolor{PuOr64}{RGB}{216,218,235}

\definecolor{PuOr6I}{RGB}{216,218,235}

\definecolor{PuOr65}{RGB}{153,142,195}

\definecolor{PuOr6K}{RGB}{153,142,195}

\definecolor{PuOr66}{RGB}{84,39,136}

\definecolor{PuOr6N}{RGB}{84,39,136}

\definecolor{PuOr71}{RGB}{179,88,6}

\definecolor{PuOr7B}{RGB}{179,88,6}

\definecolor{PuOr72}{RGB}{241,163,64}

\definecolor{PuOr7E}{RGB}{241,163,64}

\definecolor{PuOr73}{RGB}{254,224,182}

\definecolor{PuOr7G}{RGB}{254,224,182}

\definecolor{PuOr74}{RGB}{247,247,247}

\definecolor{PuOr7H}{RGB}{247,247,247}

\definecolor{PuOr75}{RGB}{216,218,235}

\definecolor{PuOr7I}{RGB}{216,218,235}

\definecolor{PuOr76}{RGB}{153,142,195}

\definecolor{PuOr7K}{RGB}{153,142,195}

\definecolor{PuOr77}{RGB}{84,39,136}

\definecolor{PuOr7N}{RGB}{84,39,136}

\definecolor{PuOr81}{RGB}{179,88,6}

\definecolor{PuOr8B}{RGB}{179,88,6}

\definecolor{PuOr82}{RGB}{224,130,20}

\definecolor{PuOr8D}{RGB}{224,130,20}

\definecolor{PuOr83}{RGB}{253,184,99}

\definecolor{PuOr8F}{RGB}{253,184,99}

\definecolor{PuOr84}{RGB}{254,224,182}

\definecolor{PuOr8G}{RGB}{254,224,182}

\definecolor{PuOr85}{RGB}{216,218,235}

\definecolor{PuOr8I}{RGB}{216,218,235}

\definecolor{PuOr86}{RGB}{178,171,210}

\definecolor{PuOr8J}{RGB}{178,171,210}

\definecolor{PuOr87}{RGB}{128,115,172}

\definecolor{PuOr8L}{RGB}{128,115,172}

\definecolor{PuOr88}{RGB}{84,39,136}

\definecolor{PuOr8N}{RGB}{84,39,136}

\definecolor{PuOr91}{RGB}{179,88,6}

\definecolor{PuOr9B}{RGB}{179,88,6}

\definecolor{PuOr92}{RGB}{224,130,20}

\definecolor{PuOr9D}{RGB}{224,130,20}

\definecolor{PuOr93}{RGB}{253,184,99}

\definecolor{PuOr9F}{RGB}{253,184,99}

\definecolor{PuOr94}{RGB}{254,224,182}

\definecolor{PuOr9G}{RGB}{254,224,182}

\definecolor{PuOr95}{RGB}{247,247,247}

\definecolor{PuOr9H}{RGB}{247,247,247}

\definecolor{PuOr96}{RGB}{216,218,235}

\definecolor{PuOr9I}{RGB}{216,218,235}

\definecolor{PuOr97}{RGB}{178,171,210}

\definecolor{PuOr9J}{RGB}{178,171,210}

\definecolor{PuOr98}{RGB}{128,115,172}

\definecolor{PuOr9L}{RGB}{128,115,172}

\definecolor{PuOr99}{RGB}{84,39,136}

\definecolor{PuOr9N}{RGB}{84,39,136}

\definecolor{PuOr101}{RGB}{127,59,8}

\definecolor{PuOr10A}{RGB}{127,59,8}

\definecolor{PuOr102}{RGB}{179,88,6}

\definecolor{PuOr10B}{RGB}{179,88,6}

\definecolor{PuOr103}{RGB}{224,130,20}

\definecolor{PuOr10D}{RGB}{224,130,20}

\definecolor{PuOr104}{RGB}{253,184,99}

\definecolor{PuOr10F}{RGB}{253,184,99}

\definecolor{PuOr105}{RGB}{254,224,182}

\definecolor{PuOr10G}{RGB}{254,224,182}

\definecolor{PuOr106}{RGB}{216,218,235}

\definecolor{PuOr10I}{RGB}{216,218,235}

\definecolor{PuOr107}{RGB}{178,171,210}

\definecolor{PuOr10J}{RGB}{178,171,210}

\definecolor{PuOr108}{RGB}{128,115,172}

\definecolor{PuOr10L}{RGB}{128,115,172}

\definecolor{PuOr109}{RGB}{84,39,136}

\definecolor{PuOr10N}{RGB}{84,39,136}

\definecolor{PuOr1010}{RGB}{45,0,75}

\definecolor{PuOr10O}{RGB}{45,0,75}

\definecolor{PuOr111}{RGB}{127,59,8}

\definecolor{PuOr11A}{RGB}{127,59,8}

\definecolor{PuOr112}{RGB}{179,88,6}

\definecolor{PuOr11B}{RGB}{179,88,6}

\definecolor{PuOr113}{RGB}{224,130,20}

\definecolor{PuOr11D}{RGB}{224,130,20}

\definecolor{PuOr114}{RGB}{253,184,99}

\definecolor{PuOr11F}{RGB}{253,184,99}

\definecolor{PuOr115}{RGB}{254,224,182}

\definecolor{PuOr11G}{RGB}{254,224,182}

\definecolor{PuOr116}{RGB}{247,247,247}

\definecolor{PuOr11H}{RGB}{247,247,247}

\definecolor{PuOr117}{RGB}{216,218,235}

\definecolor{PuOr11I}{RGB}{216,218,235}

\definecolor{PuOr118}{RGB}{178,171,210}

\definecolor{PuOr11J}{RGB}{178,171,210}

\definecolor{PuOr119}{RGB}{128,115,172}

\definecolor{PuOr11L}{RGB}{128,115,172}

\definecolor{PuOr1110}{RGB}{84,39,136}

\definecolor{PuOr11N}{RGB}{84,39,136}

\definecolor{PuOr1111}{RGB}{45,0,75}

\definecolor{PuOr11O}{RGB}{45,0,75}

\definecolor{BrBG31}{RGB}{216,179,101}

\definecolor{BrBG3E}{RGB}{216,179,101}

\definecolor{BrBG32}{RGB}{245,245,245}

\definecolor{BrBG3H}{RGB}{245,245,245}

\definecolor{BrBG33}{RGB}{90,180,172}

\definecolor{BrBG3K}{RGB}{90,180,172}

\definecolor{BrBG41}{RGB}{166,97,26}

\definecolor{BrBG4C}{RGB}{166,97,26}

\definecolor{BrBG42}{RGB}{223,194,125}

\definecolor{BrBG4F}{RGB}{223,194,125}

\definecolor{BrBG43}{RGB}{128,205,193}

\definecolor{BrBG4J}{RGB}{128,205,193}

\definecolor{BrBG44}{RGB}{1,133,113}

\definecolor{BrBG4M}{RGB}{1,133,113}

\definecolor{BrBG51}{RGB}{166,97,26}

\definecolor{BrBG5C}{RGB}{166,97,26}

\definecolor{BrBG52}{RGB}{223,194,125}

\definecolor{BrBG5F}{RGB}{223,194,125}

\definecolor{BrBG53}{RGB}{245,245,245}

\definecolor{BrBG5H}{RGB}{245,245,245}

\definecolor{BrBG54}{RGB}{128,205,193}

\definecolor{BrBG5J}{RGB}{128,205,193}

\definecolor{BrBG55}{RGB}{1,133,113}

\definecolor{BrBG5M}{RGB}{1,133,113}

\definecolor{BrBG61}{RGB}{140,81,10}

\definecolor{BrBG6B}{RGB}{140,81,10}

\definecolor{BrBG62}{RGB}{216,179,101}

\definecolor{BrBG6E}{RGB}{216,179,101}

\definecolor{BrBG63}{RGB}{246,232,195}

\definecolor{BrBG6G}{RGB}{246,232,195}

\definecolor{BrBG64}{RGB}{199,234,229}

\definecolor{BrBG6I}{RGB}{199,234,229}

\definecolor{BrBG65}{RGB}{90,180,172}

\definecolor{BrBG6K}{RGB}{90,180,172}

\definecolor{BrBG66}{RGB}{1,102,94}

\definecolor{BrBG6N}{RGB}{1,102,94}

\definecolor{BrBG71}{RGB}{140,81,10}

\definecolor{BrBG7B}{RGB}{140,81,10}

\definecolor{BrBG72}{RGB}{216,179,101}

\definecolor{BrBG7E}{RGB}{216,179,101}

\definecolor{BrBG73}{RGB}{246,232,195}

\definecolor{BrBG7G}{RGB}{246,232,195}

\definecolor{BrBG74}{RGB}{245,245,245}

\definecolor{BrBG7H}{RGB}{245,245,245}

\definecolor{BrBG75}{RGB}{199,234,229}

\definecolor{BrBG7I}{RGB}{199,234,229}

\definecolor{BrBG76}{RGB}{90,180,172}

\definecolor{BrBG7K}{RGB}{90,180,172}

\definecolor{BrBG77}{RGB}{1,102,94}

\definecolor{BrBG7N}{RGB}{1,102,94}

\definecolor{BrBG81}{RGB}{140,81,10}

\definecolor{BrBG8B}{RGB}{140,81,10}

\definecolor{BrBG82}{RGB}{191,129,45}

\definecolor{BrBG8D}{RGB}{191,129,45}

\definecolor{BrBG83}{RGB}{223,194,125}

\definecolor{BrBG8F}{RGB}{223,194,125}

\definecolor{BrBG84}{RGB}{246,232,195}

\definecolor{BrBG8G}{RGB}{246,232,195}

\definecolor{BrBG85}{RGB}{199,234,229}

\definecolor{BrBG8I}{RGB}{199,234,229}

\definecolor{BrBG86}{RGB}{128,205,193}

\definecolor{BrBG8J}{RGB}{128,205,193}

\definecolor{BrBG87}{RGB}{53,151,143}

\definecolor{BrBG8L}{RGB}{53,151,143}

\definecolor{BrBG88}{RGB}{1,102,94}

\definecolor{BrBG8N}{RGB}{1,102,94}

\definecolor{BrBG91}{RGB}{140,81,10}

\definecolor{BrBG9B}{RGB}{140,81,10}

\definecolor{BrBG92}{RGB}{191,129,45}

\definecolor{BrBG9D}{RGB}{191,129,45}

\definecolor{BrBG93}{RGB}{223,194,125}

\definecolor{BrBG9F}{RGB}{223,194,125}

\definecolor{BrBG94}{RGB}{246,232,195}

\definecolor{BrBG9G}{RGB}{246,232,195}

\definecolor{BrBG95}{RGB}{245,245,245}

\definecolor{BrBG9H}{RGB}{245,245,245}

\definecolor{BrBG96}{RGB}{199,234,229}

\definecolor{BrBG9I}{RGB}{199,234,229}

\definecolor{BrBG97}{RGB}{128,205,193}

\definecolor{BrBG9J}{RGB}{128,205,193}

\definecolor{BrBG98}{RGB}{53,151,143}

\definecolor{BrBG9L}{RGB}{53,151,143}

\definecolor{BrBG99}{RGB}{1,102,94}

\definecolor{BrBG9N}{RGB}{1,102,94}

\definecolor{BrBG101}{RGB}{84,48,5}

\definecolor{BrBG10A}{RGB}{84,48,5}

\definecolor{BrBG102}{RGB}{140,81,10}

\definecolor{BrBG10B}{RGB}{140,81,10}

\definecolor{BrBG103}{RGB}{191,129,45}

\definecolor{BrBG10D}{RGB}{191,129,45}

\definecolor{BrBG104}{RGB}{223,194,125}

\definecolor{BrBG10F}{RGB}{223,194,125}

\definecolor{BrBG105}{RGB}{246,232,195}

\definecolor{BrBG10G}{RGB}{246,232,195}

\definecolor{BrBG106}{RGB}{199,234,229}

\definecolor{BrBG10I}{RGB}{199,234,229}

\definecolor{BrBG107}{RGB}{128,205,193}

\definecolor{BrBG10J}{RGB}{128,205,193}

\definecolor{BrBG108}{RGB}{53,151,143}

\definecolor{BrBG10L}{RGB}{53,151,143}

\definecolor{BrBG109}{RGB}{1,102,94}

\definecolor{BrBG10N}{RGB}{1,102,94}

\definecolor{BrBG1010}{RGB}{0,60,48}

\definecolor{BrBG10O}{RGB}{0,60,48}

\definecolor{BrBG111}{RGB}{84,48,5}

\definecolor{BrBG11A}{RGB}{84,48,5}

\definecolor{BrBG112}{RGB}{140,81,10}

\definecolor{BrBG11B}{RGB}{140,81,10}

\definecolor{BrBG113}{RGB}{191,129,45}

\definecolor{BrBG11D}{RGB}{191,129,45}

\definecolor{BrBG114}{RGB}{223,194,125}

\definecolor{BrBG11F}{RGB}{223,194,125}

\definecolor{BrBG115}{RGB}{246,232,195}

\definecolor{BrBG11G}{RGB}{246,232,195}

\definecolor{BrBG116}{RGB}{245,245,245}

\definecolor{BrBG11H}{RGB}{245,245,245}

\definecolor{BrBG117}{RGB}{199,234,229}

\definecolor{BrBG11I}{RGB}{199,234,229}

\definecolor{BrBG118}{RGB}{128,205,193}

\definecolor{BrBG11J}{RGB}{128,205,193}

\definecolor{BrBG119}{RGB}{53,151,143}

\definecolor{BrBG11L}{RGB}{53,151,143}

\definecolor{BrBG1110}{RGB}{1,102,94}

\definecolor{BrBG11N}{RGB}{1,102,94}

\definecolor{BrBG1111}{RGB}{0,60,48}

\definecolor{BrBG11O}{RGB}{0,60,48}

\definecolor{PRGn31}{RGB}{175,141,195}

\definecolor{PRGn3E}{RGB}{175,141,195}

\definecolor{PRGn32}{RGB}{247,247,247}

\definecolor{PRGn3H}{RGB}{247,247,247}

\definecolor{PRGn33}{RGB}{127,191,123}

\definecolor{PRGn3K}{RGB}{127,191,123}

\definecolor{PRGn41}{RGB}{123,50,148}

\definecolor{PRGn4C}{RGB}{123,50,148}

\definecolor{PRGn42}{RGB}{194,165,207}

\definecolor{PRGn4F}{RGB}{194,165,207}

\definecolor{PRGn43}{RGB}{166,219,160}

\definecolor{PRGn4J}{RGB}{166,219,160}

\definecolor{PRGn44}{RGB}{0,136,55}

\definecolor{PRGn4M}{RGB}{0,136,55}

\definecolor{PRGn51}{RGB}{123,50,148}

\definecolor{PRGn5C}{RGB}{123,50,148}

\definecolor{PRGn52}{RGB}{194,165,207}

\definecolor{PRGn5F}{RGB}{194,165,207}

\definecolor{PRGn53}{RGB}{247,247,247}

\definecolor{PRGn5H}{RGB}{247,247,247}

\definecolor{PRGn54}{RGB}{166,219,160}

\definecolor{PRGn5J}{RGB}{166,219,160}

\definecolor{PRGn55}{RGB}{0,136,55}

\definecolor{PRGn5M}{RGB}{0,136,55}

\definecolor{PRGn61}{RGB}{118,42,131}

\definecolor{PRGn6B}{RGB}{118,42,131}

\definecolor{PRGn62}{RGB}{175,141,195}

\definecolor{PRGn6E}{RGB}{175,141,195}

\definecolor{PRGn63}{RGB}{231,212,232}

\definecolor{PRGn6G}{RGB}{231,212,232}

\definecolor{PRGn64}{RGB}{217,240,211}

\definecolor{PRGn6I}{RGB}{217,240,211}

\definecolor{PRGn65}{RGB}{127,191,123}

\definecolor{PRGn6K}{RGB}{127,191,123}

\definecolor{PRGn66}{RGB}{27,120,55}

\definecolor{PRGn6N}{RGB}{27,120,55}

\definecolor{PRGn71}{RGB}{118,42,131}

\definecolor{PRGn7B}{RGB}{118,42,131}

\definecolor{PRGn72}{RGB}{175,141,195}

\definecolor{PRGn7E}{RGB}{175,141,195}

\definecolor{PRGn73}{RGB}{231,212,232}

\definecolor{PRGn7G}{RGB}{231,212,232}

\definecolor{PRGn74}{RGB}{247,247,247}

\definecolor{PRGn7H}{RGB}{247,247,247}

\definecolor{PRGn75}{RGB}{217,240,211}

\definecolor{PRGn7I}{RGB}{217,240,211}

\definecolor{PRGn76}{RGB}{127,191,123}

\definecolor{PRGn7K}{RGB}{127,191,123}

\definecolor{PRGn77}{RGB}{27,120,55}

\definecolor{PRGn7N}{RGB}{27,120,55}

\definecolor{PRGn81}{RGB}{118,42,131}

\definecolor{PRGn8B}{RGB}{118,42,131}

\definecolor{PRGn82}{RGB}{153,112,171}

\definecolor{PRGn8D}{RGB}{153,112,171}

\definecolor{PRGn83}{RGB}{194,165,207}

\definecolor{PRGn8F}{RGB}{194,165,207}

\definecolor{PRGn84}{RGB}{231,212,232}

\definecolor{PRGn8G}{RGB}{231,212,232}

\definecolor{PRGn85}{RGB}{217,240,211}

\definecolor{PRGn8I}{RGB}{217,240,211}

\definecolor{PRGn86}{RGB}{166,219,160}

\definecolor{PRGn8J}{RGB}{166,219,160}

\definecolor{PRGn87}{RGB}{90,174,97}

\definecolor{PRGn8L}{RGB}{90,174,97}

\definecolor{PRGn88}{RGB}{27,120,55}

\definecolor{PRGn8N}{RGB}{27,120,55}

\definecolor{PRGn91}{RGB}{118,42,131}

\definecolor{PRGn9B}{RGB}{118,42,131}

\definecolor{PRGn92}{RGB}{153,112,171}

\definecolor{PRGn9D}{RGB}{153,112,171}

\definecolor{PRGn93}{RGB}{194,165,207}

\definecolor{PRGn9F}{RGB}{194,165,207}

\definecolor{PRGn94}{RGB}{231,212,232}

\definecolor{PRGn9G}{RGB}{231,212,232}

\definecolor{PRGn95}{RGB}{247,247,247}

\definecolor{PRGn9H}{RGB}{247,247,247}

\definecolor{PRGn96}{RGB}{217,240,211}

\definecolor{PRGn9I}{RGB}{217,240,211}

\definecolor{PRGn97}{RGB}{166,219,160}

\definecolor{PRGn9J}{RGB}{166,219,160}

\definecolor{PRGn98}{RGB}{90,174,97}

\definecolor{PRGn9L}{RGB}{90,174,97}

\definecolor{PRGn99}{RGB}{27,120,55}

\definecolor{PRGn9N}{RGB}{27,120,55}

\definecolor{PRGn101}{RGB}{64,0,75}

\definecolor{PRGn10A}{RGB}{64,0,75}

\definecolor{PRGn102}{RGB}{118,42,131}

\definecolor{PRGn10B}{RGB}{118,42,131}

\definecolor{PRGn103}{RGB}{153,112,171}

\definecolor{PRGn10D}{RGB}{153,112,171}

\definecolor{PRGn104}{RGB}{194,165,207}

\definecolor{PRGn10F}{RGB}{194,165,207}

\definecolor{PRGn105}{RGB}{231,212,232}

\definecolor{PRGn10G}{RGB}{231,212,232}

\definecolor{PRGn106}{RGB}{217,240,211}

\definecolor{PRGn10I}{RGB}{217,240,211}

\definecolor{PRGn107}{RGB}{166,219,160}

\definecolor{PRGn10J}{RGB}{166,219,160}

\definecolor{PRGn108}{RGB}{90,174,97}

\definecolor{PRGn10L}{RGB}{90,174,97}

\definecolor{PRGn109}{RGB}{27,120,55}

\definecolor{PRGn10N}{RGB}{27,120,55}

\definecolor{PRGn1010}{RGB}{0,68,27}

\definecolor{PRGn10O}{RGB}{0,68,27}

\definecolor{PRGn111}{RGB}{64,0,75}

\definecolor{PRGn11A}{RGB}{64,0,75}

\definecolor{PRGn112}{RGB}{118,42,131}

\definecolor{PRGn11B}{RGB}{118,42,131}

\definecolor{PRGn113}{RGB}{153,112,171}

\definecolor{PRGn11D}{RGB}{153,112,171}

\definecolor{PRGn114}{RGB}{194,165,207}

\definecolor{PRGn11F}{RGB}{194,165,207}

\definecolor{PRGn115}{RGB}{231,212,232}

\definecolor{PRGn11G}{RGB}{231,212,232}

\definecolor{PRGn116}{RGB}{247,247,247}

\definecolor{PRGn11H}{RGB}{247,247,247}

\definecolor{PRGn117}{RGB}{217,240,211}

\definecolor{PRGn11I}{RGB}{217,240,211}

\definecolor{PRGn118}{RGB}{166,219,160}

\definecolor{PRGn11J}{RGB}{166,219,160}

\definecolor{PRGn119}{RGB}{90,174,97}

\definecolor{PRGn11L}{RGB}{90,174,97}

\definecolor{PRGn1110}{RGB}{27,120,55}

\definecolor{PRGn11N}{RGB}{27,120,55}

\definecolor{PRGn1111}{RGB}{0,68,27}

\definecolor{PRGn11O}{RGB}{0,68,27}

\definecolor{PiYG31}{RGB}{233,163,201}

\definecolor{PiYG3E}{RGB}{233,163,201}

\definecolor{PiYG32}{RGB}{247,247,247}

\definecolor{PiYG3H}{RGB}{247,247,247}

\definecolor{PiYG33}{RGB}{161,215,106}

\definecolor{PiYG3K}{RGB}{161,215,106}

\definecolor{PiYG41}{RGB}{208,28,139}

\definecolor{PiYG4C}{RGB}{208,28,139}

\definecolor{PiYG42}{RGB}{241,182,218}

\definecolor{PiYG4F}{RGB}{241,182,218}

\definecolor{PiYG43}{RGB}{184,225,134}

\definecolor{PiYG4J}{RGB}{184,225,134}

\definecolor{PiYG44}{RGB}{77,172,38}

\definecolor{PiYG4M}{RGB}{77,172,38}

\definecolor{PiYG51}{RGB}{208,28,139}

\definecolor{PiYG5C}{RGB}{208,28,139}

\definecolor{PiYG52}{RGB}{241,182,218}

\definecolor{PiYG5F}{RGB}{241,182,218}

\definecolor{PiYG53}{RGB}{247,247,247}

\definecolor{PiYG5H}{RGB}{247,247,247}

\definecolor{PiYG54}{RGB}{184,225,134}

\definecolor{PiYG5J}{RGB}{184,225,134}

\definecolor{PiYG55}{RGB}{77,172,38}

\definecolor{PiYG5M}{RGB}{77,172,38}

\definecolor{PiYG61}{RGB}{197,27,125}

\definecolor{PiYG6B}{RGB}{197,27,125}

\definecolor{PiYG62}{RGB}{233,163,201}

\definecolor{PiYG6E}{RGB}{233,163,201}

\definecolor{PiYG63}{RGB}{253,224,239}

\definecolor{PiYG6G}{RGB}{253,224,239}

\definecolor{PiYG64}{RGB}{230,245,208}

\definecolor{PiYG6I}{RGB}{230,245,208}

\definecolor{PiYG65}{RGB}{161,215,106}

\definecolor{PiYG6K}{RGB}{161,215,106}

\definecolor{PiYG66}{RGB}{77,146,33}

\definecolor{PiYG6N}{RGB}{77,146,33}

\definecolor{PiYG71}{RGB}{197,27,125}

\definecolor{PiYG7B}{RGB}{197,27,125}

\definecolor{PiYG72}{RGB}{233,163,201}

\definecolor{PiYG7E}{RGB}{233,163,201}

\definecolor{PiYG73}{RGB}{253,224,239}

\definecolor{PiYG7G}{RGB}{253,224,239}

\definecolor{PiYG74}{RGB}{247,247,247}

\definecolor{PiYG7H}{RGB}{247,247,247}

\definecolor{PiYG75}{RGB}{230,245,208}

\definecolor{PiYG7I}{RGB}{230,245,208}

\definecolor{PiYG76}{RGB}{161,215,106}

\definecolor{PiYG7K}{RGB}{161,215,106}

\definecolor{PiYG77}{RGB}{77,146,33}

\definecolor{PiYG7N}{RGB}{77,146,33}

\definecolor{PiYG81}{RGB}{197,27,125}

\definecolor{PiYG8B}{RGB}{197,27,125}

\definecolor{PiYG82}{RGB}{222,119,174}

\definecolor{PiYG8D}{RGB}{222,119,174}

\definecolor{PiYG83}{RGB}{241,182,218}

\definecolor{PiYG8F}{RGB}{241,182,218}

\definecolor{PiYG84}{RGB}{253,224,239}

\definecolor{PiYG8G}{RGB}{253,224,239}

\definecolor{PiYG85}{RGB}{230,245,208}

\definecolor{PiYG8I}{RGB}{230,245,208}

\definecolor{PiYG86}{RGB}{184,225,134}

\definecolor{PiYG8J}{RGB}{184,225,134}

\definecolor{PiYG87}{RGB}{127,188,65}

\definecolor{PiYG8L}{RGB}{127,188,65}

\definecolor{PiYG88}{RGB}{77,146,33}

\definecolor{PiYG8N}{RGB}{77,146,33}

\definecolor{PiYG91}{RGB}{197,27,125}

\definecolor{PiYG9B}{RGB}{197,27,125}

\definecolor{PiYG92}{RGB}{222,119,174}

\definecolor{PiYG9D}{RGB}{222,119,174}

\definecolor{PiYG93}{RGB}{241,182,218}

\definecolor{PiYG9F}{RGB}{241,182,218}

\definecolor{PiYG94}{RGB}{253,224,239}

\definecolor{PiYG9G}{RGB}{253,224,239}

\definecolor{PiYG95}{RGB}{247,247,247}

\definecolor{PiYG9H}{RGB}{247,247,247}

\definecolor{PiYG96}{RGB}{230,245,208}

\definecolor{PiYG9I}{RGB}{230,245,208}

\definecolor{PiYG97}{RGB}{184,225,134}

\definecolor{PiYG9J}{RGB}{184,225,134}

\definecolor{PiYG98}{RGB}{127,188,65}

\definecolor{PiYG9L}{RGB}{127,188,65}

\definecolor{PiYG99}{RGB}{77,146,33}

\definecolor{PiYG9N}{RGB}{77,146,33}

\definecolor{PiYG101}{RGB}{142,1,82}

\definecolor{PiYG10A}{RGB}{142,1,82}

\definecolor{PiYG102}{RGB}{197,27,125}

\definecolor{PiYG10B}{RGB}{197,27,125}

\definecolor{PiYG103}{RGB}{222,119,174}

\definecolor{PiYG10D}{RGB}{222,119,174}

\definecolor{PiYG104}{RGB}{241,182,218}

\definecolor{PiYG10F}{RGB}{241,182,218}

\definecolor{PiYG105}{RGB}{253,224,239}

\definecolor{PiYG10G}{RGB}{253,224,239}

\definecolor{PiYG106}{RGB}{230,245,208}

\definecolor{PiYG10I}{RGB}{230,245,208}

\definecolor{PiYG107}{RGB}{184,225,134}

\definecolor{PiYG10J}{RGB}{184,225,134}

\definecolor{PiYG108}{RGB}{127,188,65}

\definecolor{PiYG10L}{RGB}{127,188,65}

\definecolor{PiYG109}{RGB}{77,146,33}

\definecolor{PiYG10N}{RGB}{77,146,33}

\definecolor{PiYG1010}{RGB}{39,100,25}

\definecolor{PiYG10O}{RGB}{39,100,25}

\definecolor{PiYG111}{RGB}{142,1,82}

\definecolor{PiYG11A}{RGB}{142,1,82}

\definecolor{PiYG112}{RGB}{197,27,125}

\definecolor{PiYG11B}{RGB}{197,27,125}

\definecolor{PiYG113}{RGB}{222,119,174}

\definecolor{PiYG11D}{RGB}{222,119,174}

\definecolor{PiYG114}{RGB}{241,182,218}

\definecolor{PiYG11F}{RGB}{241,182,218}

\definecolor{PiYG115}{RGB}{253,224,239}

\definecolor{PiYG11G}{RGB}{253,224,239}

\definecolor{PiYG116}{RGB}{247,247,247}

\definecolor{PiYG11H}{RGB}{247,247,247}

\definecolor{PiYG117}{RGB}{230,245,208}

\definecolor{PiYG11I}{RGB}{230,245,208}

\definecolor{PiYG118}{RGB}{184,225,134}

\definecolor{PiYG11J}{RGB}{184,225,134}

\definecolor{PiYG119}{RGB}{127,188,65}

\definecolor{PiYG11L}{RGB}{127,188,65}

\definecolor{PiYG1110}{RGB}{77,146,33}

\definecolor{PiYG11N}{RGB}{77,146,33}

\definecolor{PiYG1111}{RGB}{39,100,25}

\definecolor{PiYG11O}{RGB}{39,100,25}

\definecolor{RdBu31}{RGB}{239,138,98}

\definecolor{RdBu3E}{RGB}{239,138,98}

\definecolor{RdBu32}{RGB}{247,247,247}

\definecolor{RdBu3H}{RGB}{247,247,247}

\definecolor{RdBu33}{RGB}{103,169,207}

\definecolor{RdBu3K}{RGB}{103,169,207}

\definecolor{RdBu41}{RGB}{202,0,32}

\definecolor{RdBu4C}{RGB}{202,0,32}

\definecolor{RdBu42}{RGB}{244,165,130}

\definecolor{RdBu4F}{RGB}{244,165,130}

\definecolor{RdBu43}{RGB}{146,197,222}

\definecolor{RdBu4J}{RGB}{146,197,222}

\definecolor{RdBu44}{RGB}{5,113,176}

\definecolor{RdBu4M}{RGB}{5,113,176}

\definecolor{RdBu51}{RGB}{202,0,32}

\definecolor{RdBu5C}{RGB}{202,0,32}

\definecolor{RdBu52}{RGB}{244,165,130}

\definecolor{RdBu5F}{RGB}{244,165,130}

\definecolor{RdBu53}{RGB}{247,247,247}

\definecolor{RdBu5H}{RGB}{247,247,247}

\definecolor{RdBu54}{RGB}{146,197,222}

\definecolor{RdBu5J}{RGB}{146,197,222}

\definecolor{RdBu55}{RGB}{5,113,176}

\definecolor{RdBu5M}{RGB}{5,113,176}

\definecolor{RdBu61}{RGB}{178,24,43}

\definecolor{RdBu6B}{RGB}{178,24,43}

\definecolor{RdBu62}{RGB}{239,138,98}

\definecolor{RdBu6E}{RGB}{239,138,98}

\definecolor{RdBu63}{RGB}{253,219,199}

\definecolor{RdBu6G}{RGB}{253,219,199}

\definecolor{RdBu64}{RGB}{209,229,240}

\definecolor{RdBu6I}{RGB}{209,229,240}

\definecolor{RdBu65}{RGB}{103,169,207}

\definecolor{RdBu6K}{RGB}{103,169,207}

\definecolor{RdBu66}{RGB}{33,102,172}

\definecolor{RdBu6N}{RGB}{33,102,172}

\definecolor{RdBu71}{RGB}{178,24,43}

\definecolor{RdBu7B}{RGB}{178,24,43}

\definecolor{RdBu72}{RGB}{239,138,98}

\definecolor{RdBu7E}{RGB}{239,138,98}

\definecolor{RdBu73}{RGB}{253,219,199}

\definecolor{RdBu7G}{RGB}{253,219,199}

\definecolor{RdBu74}{RGB}{247,247,247}

\definecolor{RdBu7H}{RGB}{247,247,247}

\definecolor{RdBu75}{RGB}{209,229,240}

\definecolor{RdBu7I}{RGB}{209,229,240}

\definecolor{RdBu76}{RGB}{103,169,207}

\definecolor{RdBu7K}{RGB}{103,169,207}

\definecolor{RdBu77}{RGB}{33,102,172}

\definecolor{RdBu7N}{RGB}{33,102,172}

\definecolor{RdBu81}{RGB}{178,24,43}

\definecolor{RdBu8B}{RGB}{178,24,43}

\definecolor{RdBu82}{RGB}{214,96,77}

\definecolor{RdBu8D}{RGB}{214,96,77}

\definecolor{RdBu83}{RGB}{244,165,130}

\definecolor{RdBu8F}{RGB}{244,165,130}

\definecolor{RdBu84}{RGB}{253,219,199}

\definecolor{RdBu8G}{RGB}{253,219,199}

\definecolor{RdBu85}{RGB}{209,229,240}

\definecolor{RdBu8I}{RGB}{209,229,240}

\definecolor{RdBu86}{RGB}{146,197,222}

\definecolor{RdBu8J}{RGB}{146,197,222}

\definecolor{RdBu87}{RGB}{67,147,195}

\definecolor{RdBu8L}{RGB}{67,147,195}

\definecolor{RdBu88}{RGB}{33,102,172}

\definecolor{RdBu8N}{RGB}{33,102,172}

\definecolor{RdBu91}{RGB}{178,24,43}

\definecolor{RdBu9B}{RGB}{178,24,43}

\definecolor{RdBu92}{RGB}{214,96,77}

\definecolor{RdBu9D}{RGB}{214,96,77}

\definecolor{RdBu93}{RGB}{244,165,130}

\definecolor{RdBu9F}{RGB}{244,165,130}

\definecolor{RdBu94}{RGB}{253,219,199}

\definecolor{RdBu9G}{RGB}{253,219,199}

\definecolor{RdBu95}{RGB}{247,247,247}

\definecolor{RdBu9H}{RGB}{247,247,247}

\definecolor{RdBu96}{RGB}{209,229,240}

\definecolor{RdBu9I}{RGB}{209,229,240}

\definecolor{RdBu97}{RGB}{146,197,222}

\definecolor{RdBu9J}{RGB}{146,197,222}

\definecolor{RdBu98}{RGB}{67,147,195}

\definecolor{RdBu9L}{RGB}{67,147,195}

\definecolor{RdBu99}{RGB}{33,102,172}

\definecolor{RdBu9N}{RGB}{33,102,172}

\definecolor{RdBu101}{RGB}{103,0,31}

\definecolor{RdBu10A}{RGB}{103,0,31}

\definecolor{RdBu102}{RGB}{178,24,43}

\definecolor{RdBu10B}{RGB}{178,24,43}

\definecolor{RdBu103}{RGB}{214,96,77}

\definecolor{RdBu10D}{RGB}{214,96,77}

\definecolor{RdBu104}{RGB}{244,165,130}

\definecolor{RdBu10F}{RGB}{244,165,130}

\definecolor{RdBu105}{RGB}{253,219,199}

\definecolor{RdBu10G}{RGB}{253,219,199}

\definecolor{RdBu106}{RGB}{209,229,240}

\definecolor{RdBu10I}{RGB}{209,229,240}

\definecolor{RdBu107}{RGB}{146,197,222}

\definecolor{RdBu10J}{RGB}{146,197,222}

\definecolor{RdBu108}{RGB}{67,147,195}

\definecolor{RdBu10L}{RGB}{67,147,195}

\definecolor{RdBu109}{RGB}{33,102,172}

\definecolor{RdBu10N}{RGB}{33,102,172}

\definecolor{RdBu1010}{RGB}{5,48,97}

\definecolor{RdBu10O}{RGB}{5,48,97}

\definecolor{RdBu111}{RGB}{103,0,31}

\definecolor{RdBu11A}{RGB}{103,0,31}

\definecolor{RdBu112}{RGB}{178,24,43}

\definecolor{RdBu11B}{RGB}{178,24,43}

\definecolor{RdBu113}{RGB}{214,96,77}

\definecolor{RdBu11D}{RGB}{214,96,77}

\definecolor{RdBu114}{RGB}{244,165,130}

\definecolor{RdBu11F}{RGB}{244,165,130}

\definecolor{RdBu115}{RGB}{253,219,199}

\definecolor{RdBu11G}{RGB}{253,219,199}

\definecolor{RdBu116}{RGB}{247,247,247}

\definecolor{RdBu11H}{RGB}{247,247,247}

\definecolor{RdBu117}{RGB}{209,229,240}

\definecolor{RdBu11I}{RGB}{209,229,240}

\definecolor{RdBu118}{RGB}{146,197,222}

\definecolor{RdBu11J}{RGB}{146,197,222}

\definecolor{RdBu119}{RGB}{67,147,195}

\definecolor{RdBu11L}{RGB}{67,147,195}

\definecolor{RdBu1110}{RGB}{33,102,172}

\definecolor{RdBu11N}{RGB}{33,102,172}

\definecolor{RdBu1111}{RGB}{5,48,97}

\definecolor{RdBu11O}{RGB}{5,48,97}

\definecolor{RdGy31}{RGB}{239,138,98}

\definecolor{RdGy3E}{RGB}{239,138,98}

\definecolor{RdGy32}{RGB}{255,255,255}

\definecolor{RdGy3H}{RGB}{255,255,255}

\definecolor{RdGy33}{RGB}{153,153,153}

\definecolor{RdGy3K}{RGB}{153,153,153}

\definecolor{RdGy41}{RGB}{202,0,32}

\definecolor{RdGy4C}{RGB}{202,0,32}

\definecolor{RdGy42}{RGB}{244,165,130}

\definecolor{RdGy4F}{RGB}{244,165,130}

\definecolor{RdGy43}{RGB}{186,186,186}

\definecolor{RdGy4J}{RGB}{186,186,186}

\definecolor{RdGy44}{RGB}{64,64,64}

\definecolor{RdGy4M}{RGB}{64,64,64}

\definecolor{RdGy51}{RGB}{202,0,32}

\definecolor{RdGy5C}{RGB}{202,0,32}

\definecolor{RdGy52}{RGB}{244,165,130}

\definecolor{RdGy5F}{RGB}{244,165,130}

\definecolor{RdGy53}{RGB}{255,255,255}

\definecolor{RdGy5H}{RGB}{255,255,255}

\definecolor{RdGy54}{RGB}{186,186,186}

\definecolor{RdGy5J}{RGB}{186,186,186}

\definecolor{RdGy55}{RGB}{64,64,64}

\definecolor{RdGy5M}{RGB}{64,64,64}

\definecolor{RdGy61}{RGB}{178,24,43}

\definecolor{RdGy6B}{RGB}{178,24,43}

\definecolor{RdGy62}{RGB}{239,138,98}

\definecolor{RdGy6E}{RGB}{239,138,98}

\definecolor{RdGy63}{RGB}{253,219,199}

\definecolor{RdGy6G}{RGB}{253,219,199}

\definecolor{RdGy64}{RGB}{224,224,224}

\definecolor{RdGy6I}{RGB}{224,224,224}

\definecolor{RdGy65}{RGB}{153,153,153}

\definecolor{RdGy6K}{RGB}{153,153,153}

\definecolor{RdGy66}{RGB}{77,77,77}

\definecolor{RdGy6N}{RGB}{77,77,77}

\definecolor{RdGy71}{RGB}{178,24,43}

\definecolor{RdGy7B}{RGB}{178,24,43}

\definecolor{RdGy72}{RGB}{239,138,98}

\definecolor{RdGy7E}{RGB}{239,138,98}

\definecolor{RdGy73}{RGB}{253,219,199}

\definecolor{RdGy7G}{RGB}{253,219,199}

\definecolor{RdGy74}{RGB}{255,255,255}

\definecolor{RdGy7H}{RGB}{255,255,255}

\definecolor{RdGy75}{RGB}{224,224,224}

\definecolor{RdGy7I}{RGB}{224,224,224}

\definecolor{RdGy76}{RGB}{153,153,153}

\definecolor{RdGy7K}{RGB}{153,153,153}

\definecolor{RdGy77}{RGB}{77,77,77}

\definecolor{RdGy7N}{RGB}{77,77,77}

\definecolor{RdGy81}{RGB}{178,24,43}

\definecolor{RdGy8B}{RGB}{178,24,43}

\definecolor{RdGy82}{RGB}{214,96,77}

\definecolor{RdGy8D}{RGB}{214,96,77}

\definecolor{RdGy83}{RGB}{244,165,130}

\definecolor{RdGy8F}{RGB}{244,165,130}

\definecolor{RdGy84}{RGB}{253,219,199}

\definecolor{RdGy8G}{RGB}{253,219,199}

\definecolor{RdGy85}{RGB}{224,224,224}

\definecolor{RdGy8I}{RGB}{224,224,224}

\definecolor{RdGy86}{RGB}{186,186,186}

\definecolor{RdGy8J}{RGB}{186,186,186}

\definecolor{RdGy87}{RGB}{135,135,135}

\definecolor{RdGy8L}{RGB}{135,135,135}

\definecolor{RdGy88}{RGB}{77,77,77}

\definecolor{RdGy8N}{RGB}{77,77,77}

\definecolor{RdGy91}{RGB}{178,24,43}

\definecolor{RdGy9B}{RGB}{178,24,43}

\definecolor{RdGy92}{RGB}{214,96,77}

\definecolor{RdGy9D}{RGB}{214,96,77}

\definecolor{RdGy93}{RGB}{244,165,130}

\definecolor{RdGy9F}{RGB}{244,165,130}

\definecolor{RdGy94}{RGB}{253,219,199}

\definecolor{RdGy9G}{RGB}{253,219,199}

\definecolor{RdGy95}{RGB}{255,255,255}

\definecolor{RdGy9H}{RGB}{255,255,255}

\definecolor{RdGy96}{RGB}{224,224,224}

\definecolor{RdGy9I}{RGB}{224,224,224}

\definecolor{RdGy97}{RGB}{186,186,186}

\definecolor{RdGy9J}{RGB}{186,186,186}

\definecolor{RdGy98}{RGB}{135,135,135}

\definecolor{RdGy9L}{RGB}{135,135,135}

\definecolor{RdGy99}{RGB}{77,77,77}

\definecolor{RdGy9N}{RGB}{77,77,77}

\definecolor{RdGy101}{RGB}{103,0,31}

\definecolor{RdGy10A}{RGB}{103,0,31}

\definecolor{RdGy102}{RGB}{178,24,43}

\definecolor{RdGy10B}{RGB}{178,24,43}

\definecolor{RdGy103}{RGB}{214,96,77}

\definecolor{RdGy10D}{RGB}{214,96,77}

\definecolor{RdGy104}{RGB}{244,165,130}

\definecolor{RdGy10F}{RGB}{244,165,130}

\definecolor{RdGy105}{RGB}{253,219,199}

\definecolor{RdGy10G}{RGB}{253,219,199}

\definecolor{RdGy106}{RGB}{224,224,224}

\definecolor{RdGy10I}{RGB}{224,224,224}

\definecolor{RdGy107}{RGB}{186,186,186}

\definecolor{RdGy10J}{RGB}{186,186,186}

\definecolor{RdGy108}{RGB}{135,135,135}

\definecolor{RdGy10L}{RGB}{135,135,135}

\definecolor{RdGy109}{RGB}{77,77,77}

\definecolor{RdGy10N}{RGB}{77,77,77}

\definecolor{RdGy1010}{RGB}{26,26,26}

\definecolor{RdGy10O}{RGB}{26,26,26}

\definecolor{RdGy111}{RGB}{103,0,31}

\definecolor{RdGy11A}{RGB}{103,0,31}

\definecolor{RdGy112}{RGB}{178,24,43}

\definecolor{RdGy11B}{RGB}{178,24,43}

\definecolor{RdGy113}{RGB}{214,96,77}

\definecolor{RdGy11D}{RGB}{214,96,77}

\definecolor{RdGy114}{RGB}{244,165,130}

\definecolor{RdGy11F}{RGB}{244,165,130}

\definecolor{RdGy115}{RGB}{253,219,199}

\definecolor{RdGy11G}{RGB}{253,219,199}

\definecolor{RdGy116}{RGB}{255,255,255}

\definecolor{RdGy11H}{RGB}{255,255,255}

\definecolor{RdGy117}{RGB}{224,224,224}

\definecolor{RdGy11I}{RGB}{224,224,224}

\definecolor{RdGy118}{RGB}{186,186,186}

\definecolor{RdGy11J}{RGB}{186,186,186}

\definecolor{RdGy119}{RGB}{135,135,135}

\definecolor{RdGy11L}{RGB}{135,135,135}

\definecolor{RdGy1110}{RGB}{77,77,77}

\definecolor{RdGy11N}{RGB}{77,77,77}

\definecolor{RdGy1111}{RGB}{26,26,26}

\definecolor{RdGy11O}{RGB}{26,26,26}

\definecolor{RdYlBu31}{RGB}{252,141,89}

\definecolor{RdYlBu3E}{RGB}{252,141,89}

\definecolor{RdYlBu32}{RGB}{255,255,191}

\definecolor{RdYlBu3H}{RGB}{255,255,191}

\definecolor{RdYlBu33}{RGB}{145,191,219}

\definecolor{RdYlBu3K}{RGB}{145,191,219}

\definecolor{RdYlBu41}{RGB}{215,25,28}

\definecolor{RdYlBu4C}{RGB}{215,25,28}

\definecolor{RdYlBu42}{RGB}{253,174,97}

\definecolor{RdYlBu4F}{RGB}{253,174,97}

\definecolor{RdYlBu43}{RGB}{171,217,233}

\definecolor{RdYlBu4J}{RGB}{171,217,233}

\definecolor{RdYlBu44}{RGB}{44,123,182}

\definecolor{RdYlBu4M}{RGB}{44,123,182}

\definecolor{RdYlBu51}{RGB}{215,25,28}

\definecolor{RdYlBu5C}{RGB}{215,25,28}

\definecolor{RdYlBu52}{RGB}{253,174,97}

\definecolor{RdYlBu5F}{RGB}{253,174,97}

\definecolor{RdYlBu53}{RGB}{255,255,191}

\definecolor{RdYlBu5H}{RGB}{255,255,191}

\definecolor{RdYlBu54}{RGB}{171,217,233}

\definecolor{RdYlBu5J}{RGB}{171,217,233}

\definecolor{RdYlBu55}{RGB}{44,123,182}

\definecolor{RdYlBu5M}{RGB}{44,123,182}

\definecolor{RdYlBu61}{RGB}{215,48,39}

\definecolor{RdYlBu6B}{RGB}{215,48,39}

\definecolor{RdYlBu62}{RGB}{252,141,89}

\definecolor{RdYlBu6E}{RGB}{252,141,89}

\definecolor{RdYlBu63}{RGB}{254,224,144}

\definecolor{RdYlBu6G}{RGB}{254,224,144}

\definecolor{RdYlBu64}{RGB}{224,243,248}

\definecolor{RdYlBu6I}{RGB}{224,243,248}

\definecolor{RdYlBu65}{RGB}{145,191,219}

\definecolor{RdYlBu6K}{RGB}{145,191,219}

\definecolor{RdYlBu66}{RGB}{69,117,180}

\definecolor{RdYlBu6N}{RGB}{69,117,180}

\definecolor{RdYlBu71}{RGB}{215,48,39}

\definecolor{RdYlBu7B}{RGB}{215,48,39}

\definecolor{RdYlBu72}{RGB}{252,141,89}

\definecolor{RdYlBu7E}{RGB}{252,141,89}

\definecolor{RdYlBu73}{RGB}{254,224,144}

\definecolor{RdYlBu7G}{RGB}{254,224,144}

\definecolor{RdYlBu74}{RGB}{255,255,191}

\definecolor{RdYlBu7H}{RGB}{255,255,191}

\definecolor{RdYlBu75}{RGB}{224,243,248}

\definecolor{RdYlBu7I}{RGB}{224,243,248}

\definecolor{RdYlBu76}{RGB}{145,191,219}

\definecolor{RdYlBu7K}{RGB}{145,191,219}

\definecolor{RdYlBu77}{RGB}{69,117,180}

\definecolor{RdYlBu7N}{RGB}{69,117,180}

\definecolor{RdYlBu81}{RGB}{215,48,39}

\definecolor{RdYlBu8B}{RGB}{215,48,39}

\definecolor{RdYlBu82}{RGB}{244,109,67}

\definecolor{RdYlBu8D}{RGB}{244,109,67}

\definecolor{RdYlBu83}{RGB}{253,174,97}

\definecolor{RdYlBu8F}{RGB}{253,174,97}

\definecolor{RdYlBu84}{RGB}{254,224,144}

\definecolor{RdYlBu8G}{RGB}{254,224,144}

\definecolor{RdYlBu85}{RGB}{224,243,248}

\definecolor{RdYlBu8I}{RGB}{224,243,248}

\definecolor{RdYlBu86}{RGB}{171,217,233}

\definecolor{RdYlBu8J}{RGB}{171,217,233}

\definecolor{RdYlBu87}{RGB}{116,173,209}

\definecolor{RdYlBu8L}{RGB}{116,173,209}

\definecolor{RdYlBu88}{RGB}{69,117,180}

\definecolor{RdYlBu8N}{RGB}{69,117,180}

\definecolor{RdYlBu91}{RGB}{215,48,39}

\definecolor{RdYlBu9B}{RGB}{215,48,39}

\definecolor{RdYlBu92}{RGB}{244,109,67}

\definecolor{RdYlBu9D}{RGB}{244,109,67}

\definecolor{RdYlBu93}{RGB}{253,174,97}

\definecolor{RdYlBu9F}{RGB}{253,174,97}

\definecolor{RdYlBu94}{RGB}{254,224,144}

\definecolor{RdYlBu9G}{RGB}{254,224,144}

\definecolor{RdYlBu95}{RGB}{255,255,191}

\definecolor{RdYlBu9H}{RGB}{255,255,191}

\definecolor{RdYlBu96}{RGB}{224,243,248}

\definecolor{RdYlBu9I}{RGB}{224,243,248}

\definecolor{RdYlBu97}{RGB}{171,217,233}

\definecolor{RdYlBu9J}{RGB}{171,217,233}

\definecolor{RdYlBu98}{RGB}{116,173,209}

\definecolor{RdYlBu9L}{RGB}{116,173,209}

\definecolor{RdYlBu99}{RGB}{69,117,180}

\definecolor{RdYlBu9N}{RGB}{69,117,180}

\definecolor{RdYlBu101}{RGB}{165,0,38}

\definecolor{RdYlBu10A}{RGB}{165,0,38}

\definecolor{RdYlBu102}{RGB}{215,48,39}

\definecolor{RdYlBu10B}{RGB}{215,48,39}

\definecolor{RdYlBu103}{RGB}{244,109,67}

\definecolor{RdYlBu10D}{RGB}{244,109,67}

\definecolor{RdYlBu104}{RGB}{253,174,97}

\definecolor{RdYlBu10F}{RGB}{253,174,97}

\definecolor{RdYlBu105}{RGB}{254,224,144}

\definecolor{RdYlBu10G}{RGB}{254,224,144}

\definecolor{RdYlBu106}{RGB}{224,243,248}

\definecolor{RdYlBu10I}{RGB}{224,243,248}

\definecolor{RdYlBu107}{RGB}{171,217,233}

\definecolor{RdYlBu10J}{RGB}{171,217,233}

\definecolor{RdYlBu108}{RGB}{116,173,209}

\definecolor{RdYlBu10L}{RGB}{116,173,209}

\definecolor{RdYlBu109}{RGB}{69,117,180}

\definecolor{RdYlBu10N}{RGB}{69,117,180}

\definecolor{RdYlBu1010}{RGB}{49,54,149}

\definecolor{RdYlBu10O}{RGB}{49,54,149}

\definecolor{RdYlBu111}{RGB}{165,0,38}

\definecolor{RdYlBu11A}{RGB}{165,0,38}

\definecolor{RdYlBu112}{RGB}{215,48,39}

\definecolor{RdYlBu11B}{RGB}{215,48,39}

\definecolor{RdYlBu113}{RGB}{244,109,67}

\definecolor{RdYlBu11D}{RGB}{244,109,67}

\definecolor{RdYlBu114}{RGB}{253,174,97}

\definecolor{RdYlBu11F}{RGB}{253,174,97}

\definecolor{RdYlBu115}{RGB}{254,224,144}

\definecolor{RdYlBu11G}{RGB}{254,224,144}

\definecolor{RdYlBu116}{RGB}{255,255,191}

\definecolor{RdYlBu11H}{RGB}{255,255,191}

\definecolor{RdYlBu117}{RGB}{224,243,248}

\definecolor{RdYlBu11I}{RGB}{224,243,248}

\definecolor{RdYlBu118}{RGB}{171,217,233}

\definecolor{RdYlBu11J}{RGB}{171,217,233}

\definecolor{RdYlBu119}{RGB}{116,173,209}

\definecolor{RdYlBu11L}{RGB}{116,173,209}

\definecolor{RdYlBu1110}{RGB}{69,117,180}

\definecolor{RdYlBu11N}{RGB}{69,117,180}

\definecolor{RdYlBu1111}{RGB}{49,54,149}

\definecolor{RdYlBu11O}{RGB}{49,54,149}

\definecolor{Spectral31}{RGB}{252,141,89}

\definecolor{Spectral3E}{RGB}{252,141,89}

\definecolor{Spectral32}{RGB}{255,255,191}

\definecolor{Spectral3H}{RGB}{255,255,191}

\definecolor{Spectral33}{RGB}{153,213,148}

\definecolor{Spectral3K}{RGB}{153,213,148}

\definecolor{Spectral41}{RGB}{215,25,28}

\definecolor{Spectral4C}{RGB}{215,25,28}

\definecolor{Spectral42}{RGB}{253,174,97}

\definecolor{Spectral4F}{RGB}{253,174,97}

\definecolor{Spectral43}{RGB}{171,221,164}

\definecolor{Spectral4J}{RGB}{171,221,164}

\definecolor{Spectral44}{RGB}{43,131,186}

\definecolor{Spectral4M}{RGB}{43,131,186}

\definecolor{Spectral51}{RGB}{215,25,28}

\definecolor{Spectral5C}{RGB}{215,25,28}

\definecolor{Spectral52}{RGB}{253,174,97}

\definecolor{Spectral5F}{RGB}{253,174,97}

\definecolor{Spectral53}{RGB}{255,255,191}

\definecolor{Spectral5H}{RGB}{255,255,191}

\definecolor{Spectral54}{RGB}{171,221,164}

\definecolor{Spectral5J}{RGB}{171,221,164}

\definecolor{Spectral55}{RGB}{43,131,186}

\definecolor{Spectral5M}{RGB}{43,131,186}

\definecolor{Spectral61}{RGB}{213,62,79}

\definecolor{Spectral6B}{RGB}{213,62,79}

\definecolor{Spectral62}{RGB}{252,141,89}

\definecolor{Spectral6E}{RGB}{252,141,89}

\definecolor{Spectral63}{RGB}{254,224,139}

\definecolor{Spectral6G}{RGB}{254,224,139}

\definecolor{Spectral64}{RGB}{230,245,152}

\definecolor{Spectral6I}{RGB}{230,245,152}

\definecolor{Spectral65}{RGB}{153,213,148}

\definecolor{Spectral6K}{RGB}{153,213,148}

\definecolor{Spectral66}{RGB}{50,136,189}

\definecolor{Spectral6N}{RGB}{50,136,189}

\definecolor{Spectral71}{RGB}{213,62,79}

\definecolor{Spectral7B}{RGB}{213,62,79}

\definecolor{Spectral72}{RGB}{252,141,89}

\definecolor{Spectral7E}{RGB}{252,141,89}

\definecolor{Spectral73}{RGB}{254,224,139}

\definecolor{Spectral7G}{RGB}{254,224,139}

\definecolor{Spectral74}{RGB}{255,255,191}

\definecolor{Spectral7H}{RGB}{255,255,191}

\definecolor{Spectral75}{RGB}{230,245,152}

\definecolor{Spectral7I}{RGB}{230,245,152}

\definecolor{Spectral76}{RGB}{153,213,148}

\definecolor{Spectral7K}{RGB}{153,213,148}

\definecolor{Spectral77}{RGB}{50,136,189}

\definecolor{Spectral7N}{RGB}{50,136,189}

\definecolor{Spectral81}{RGB}{213,62,79}

\definecolor{Spectral8B}{RGB}{213,62,79}

\definecolor{Spectral82}{RGB}{244,109,67}

\definecolor{Spectral8D}{RGB}{244,109,67}

\definecolor{Spectral83}{RGB}{253,174,97}

\definecolor{Spectral8F}{RGB}{253,174,97}

\definecolor{Spectral84}{RGB}{254,224,139}

\definecolor{Spectral8G}{RGB}{254,224,139}

\definecolor{Spectral85}{RGB}{230,245,152}

\definecolor{Spectral8I}{RGB}{230,245,152}

\definecolor{Spectral86}{RGB}{171,221,164}

\definecolor{Spectral8J}{RGB}{171,221,164}

\definecolor{Spectral87}{RGB}{102,194,165}

\definecolor{Spectral8L}{RGB}{102,194,165}

\definecolor{Spectral88}{RGB}{50,136,189}

\definecolor{Spectral8N}{RGB}{50,136,189}

\definecolor{Spectral91}{RGB}{213,62,79}

\definecolor{Spectral9B}{RGB}{213,62,79}

\definecolor{Spectral92}{RGB}{244,109,67}

\definecolor{Spectral9D}{RGB}{244,109,67}

\definecolor{Spectral93}{RGB}{253,174,97}

\definecolor{Spectral9F}{RGB}{253,174,97}

\definecolor{Spectral94}{RGB}{254,224,139}

\definecolor{Spectral9G}{RGB}{254,224,139}

\definecolor{Spectral95}{RGB}{255,255,191}

\definecolor{Spectral9H}{RGB}{255,255,191}

\definecolor{Spectral96}{RGB}{230,245,152}

\definecolor{Spectral9I}{RGB}{230,245,152}

\definecolor{Spectral97}{RGB}{171,221,164}

\definecolor{Spectral9J}{RGB}{171,221,164}

\definecolor{Spectral98}{RGB}{102,194,165}

\definecolor{Spectral9L}{RGB}{102,194,165}

\definecolor{Spectral99}{RGB}{50,136,189}

\definecolor{Spectral9N}{RGB}{50,136,189}

\definecolor{Spectral101}{RGB}{158,1,66}

\definecolor{Spectral10A}{RGB}{158,1,66}

\definecolor{Spectral102}{RGB}{213,62,79}

\definecolor{Spectral10B}{RGB}{213,62,79}

\definecolor{Spectral103}{RGB}{244,109,67}

\definecolor{Spectral10D}{RGB}{244,109,67}

\definecolor{Spectral104}{RGB}{253,174,97}

\definecolor{Spectral10F}{RGB}{253,174,97}

\definecolor{Spectral105}{RGB}{254,224,139}

\definecolor{Spectral10G}{RGB}{254,224,139}

\definecolor{Spectral106}{RGB}{230,245,152}

\definecolor{Spectral10I}{RGB}{230,245,152}

\definecolor{Spectral107}{RGB}{171,221,164}

\definecolor{Spectral10J}{RGB}{171,221,164}

\definecolor{Spectral108}{RGB}{102,194,165}

\definecolor{Spectral10L}{RGB}{102,194,165}

\definecolor{Spectral109}{RGB}{50,136,189}

\definecolor{Spectral10N}{RGB}{50,136,189}

\definecolor{Spectral1010}{RGB}{94,79,162}

\definecolor{Spectral10O}{RGB}{94,79,162}

\definecolor{Spectral111}{RGB}{158,1,66}

\definecolor{Spectral11A}{RGB}{158,1,66}

\definecolor{Spectral112}{RGB}{213,62,79}

\definecolor{Spectral11B}{RGB}{213,62,79}

\definecolor{Spectral113}{RGB}{244,109,67}

\definecolor{Spectral11D}{RGB}{244,109,67}

\definecolor{Spectral114}{RGB}{253,174,97}

\definecolor{Spectral11F}{RGB}{253,174,97}

\definecolor{Spectral115}{RGB}{254,224,139}

\definecolor{Spectral11G}{RGB}{254,224,139}

\definecolor{Spectral116}{RGB}{255,255,191}

\definecolor{Spectral11H}{RGB}{255,255,191}

\definecolor{Spectral117}{RGB}{230,245,152}

\definecolor{Spectral11I}{RGB}{230,245,152}

\definecolor{Spectral118}{RGB}{171,221,164}

\definecolor{Spectral11J}{RGB}{171,221,164}

\definecolor{Spectral119}{RGB}{102,194,165}

\definecolor{Spectral11L}{RGB}{102,194,165}

\definecolor{Spectral1110}{RGB}{50,136,189}

\definecolor{Spectral11N}{RGB}{50,136,189}

\definecolor{Spectral1111}{RGB}{94,79,162}

\definecolor{Spectral11O}{RGB}{94,79,162}

\definecolor{RdYlGn31}{RGB}{252,141,89}

\definecolor{RdYlGn3E}{RGB}{252,141,89}

\definecolor{RdYlGn32}{RGB}{255,255,191}

\definecolor{RdYlGn3H}{RGB}{255,255,191}

\definecolor{RdYlGn33}{RGB}{145,207,96}

\definecolor{RdYlGn3K}{RGB}{145,207,96}

\definecolor{RdYlGn41}{RGB}{215,25,28}

\definecolor{RdYlGn4C}{RGB}{215,25,28}

\definecolor{RdYlGn42}{RGB}{253,174,97}

\definecolor{RdYlGn4F}{RGB}{253,174,97}

\definecolor{RdYlGn43}{RGB}{166,217,106}

\definecolor{RdYlGn4J}{RGB}{166,217,106}

\definecolor{RdYlGn44}{RGB}{26,150,65}

\definecolor{RdYlGn4M}{RGB}{26,150,65}

\definecolor{RdYlGn51}{RGB}{215,25,28}

\definecolor{RdYlGn5C}{RGB}{215,25,28}

\definecolor{RdYlGn52}{RGB}{253,174,97}

\definecolor{RdYlGn5F}{RGB}{253,174,97}

\definecolor{RdYlGn53}{RGB}{255,255,191}

\definecolor{RdYlGn5H}{RGB}{255,255,191}

\definecolor{RdYlGn54}{RGB}{166,217,106}

\definecolor{RdYlGn5J}{RGB}{166,217,106}

\definecolor{RdYlGn55}{RGB}{26,150,65}

\definecolor{RdYlGn5M}{RGB}{26,150,65}

\definecolor{RdYlGn61}{RGB}{215,48,39}

\definecolor{RdYlGn6B}{RGB}{215,48,39}

\definecolor{RdYlGn62}{RGB}{252,141,89}

\definecolor{RdYlGn6E}{RGB}{252,141,89}

\definecolor{RdYlGn63}{RGB}{254,224,139}

\definecolor{RdYlGn6G}{RGB}{254,224,139}

\definecolor{RdYlGn64}{RGB}{217,239,139}

\definecolor{RdYlGn6I}{RGB}{217,239,139}

\definecolor{RdYlGn65}{RGB}{145,207,96}

\definecolor{RdYlGn6K}{RGB}{145,207,96}

\definecolor{RdYlGn66}{RGB}{26,152,80}

\definecolor{RdYlGn6N}{RGB}{26,152,80}

\definecolor{RdYlGn71}{RGB}{215,48,39}

\definecolor{RdYlGn7B}{RGB}{215,48,39}

\definecolor{RdYlGn72}{RGB}{252,141,89}

\definecolor{RdYlGn7E}{RGB}{252,141,89}

\definecolor{RdYlGn73}{RGB}{254,224,139}

\definecolor{RdYlGn7G}{RGB}{254,224,139}

\definecolor{RdYlGn74}{RGB}{255,255,191}

\definecolor{RdYlGn7H}{RGB}{255,255,191}

\definecolor{RdYlGn75}{RGB}{217,239,139}

\definecolor{RdYlGn7I}{RGB}{217,239,139}

\definecolor{RdYlGn76}{RGB}{145,207,96}

\definecolor{RdYlGn7K}{RGB}{145,207,96}

\definecolor{RdYlGn77}{RGB}{26,152,80}

\definecolor{RdYlGn7N}{RGB}{26,152,80}

\definecolor{RdYlGn81}{RGB}{215,48,39}

\definecolor{RdYlGn8B}{RGB}{215,48,39}

\definecolor{RdYlGn82}{RGB}{244,109,67}

\definecolor{RdYlGn8D}{RGB}{244,109,67}

\definecolor{RdYlGn83}{RGB}{253,174,97}

\definecolor{RdYlGn8F}{RGB}{253,174,97}

\definecolor{RdYlGn84}{RGB}{254,224,139}

\definecolor{RdYlGn8G}{RGB}{254,224,139}

\definecolor{RdYlGn85}{RGB}{217,239,139}

\definecolor{RdYlGn8I}{RGB}{217,239,139}

\definecolor{RdYlGn86}{RGB}{166,217,106}

\definecolor{RdYlGn8J}{RGB}{166,217,106}

\definecolor{RdYlGn87}{RGB}{102,189,99}

\definecolor{RdYlGn8L}{RGB}{102,189,99}

\definecolor{RdYlGn88}{RGB}{26,152,80}

\definecolor{RdYlGn8N}{RGB}{26,152,80}

\definecolor{RdYlGn91}{RGB}{215,48,39}

\definecolor{RdYlGn9B}{RGB}{215,48,39}

\definecolor{RdYlGn92}{RGB}{244,109,67}

\definecolor{RdYlGn9D}{RGB}{244,109,67}

\definecolor{RdYlGn93}{RGB}{253,174,97}

\definecolor{RdYlGn9F}{RGB}{253,174,97}

\definecolor{RdYlGn94}{RGB}{254,224,139}

\definecolor{RdYlGn9G}{RGB}{254,224,139}

\definecolor{RdYlGn95}{RGB}{255,255,191}

\definecolor{RdYlGn9H}{RGB}{255,255,191}

\definecolor{RdYlGn96}{RGB}{217,239,139}

\definecolor{RdYlGn9I}{RGB}{217,239,139}

\definecolor{RdYlGn97}{RGB}{166,217,106}

\definecolor{RdYlGn9J}{RGB}{166,217,106}

\definecolor{RdYlGn98}{RGB}{102,189,99}

\definecolor{RdYlGn9L}{RGB}{102,189,99}

\definecolor{RdYlGn99}{RGB}{26,152,80}

\definecolor{RdYlGn9N}{RGB}{26,152,80}

\definecolor{RdYlGn101}{RGB}{165,0,38}

\definecolor{RdYlGn10A}{RGB}{165,0,38}

\definecolor{RdYlGn102}{RGB}{215,48,39}

\definecolor{RdYlGn10B}{RGB}{215,48,39}

\definecolor{RdYlGn103}{RGB}{244,109,67}

\definecolor{RdYlGn10D}{RGB}{244,109,67}

\definecolor{RdYlGn104}{RGB}{253,174,97}

\definecolor{RdYlGn10F}{RGB}{253,174,97}

\definecolor{RdYlGn105}{RGB}{254,224,139}

\definecolor{RdYlGn10G}{RGB}{254,224,139}

\definecolor{RdYlGn106}{RGB}{217,239,139}

\definecolor{RdYlGn10I}{RGB}{217,239,139}

\definecolor{RdYlGn107}{RGB}{166,217,106}

\definecolor{RdYlGn10J}{RGB}{166,217,106}

\definecolor{RdYlGn108}{RGB}{102,189,99}

\definecolor{RdYlGn10L}{RGB}{102,189,99}

\definecolor{RdYlGn109}{RGB}{26,152,80}

\definecolor{RdYlGn10N}{RGB}{26,152,80}

\definecolor{RdYlGn1010}{RGB}{0,104,55}

\definecolor{RdYlGn10O}{RGB}{0,104,55}

\definecolor{RdYlGn111}{RGB}{165,0,38}

\definecolor{RdYlGn11A}{RGB}{165,0,38}

\definecolor{RdYlGn112}{RGB}{215,48,39}

\definecolor{RdYlGn11B}{RGB}{215,48,39}

\definecolor{RdYlGn113}{RGB}{244,109,67}

\definecolor{RdYlGn11D}{RGB}{244,109,67}

\definecolor{RdYlGn114}{RGB}{253,174,97}

\definecolor{RdYlGn11F}{RGB}{253,174,97}

\definecolor{RdYlGn115}{RGB}{254,224,139}

\definecolor{RdYlGn11G}{RGB}{254,224,139}

\definecolor{RdYlGn116}{RGB}{255,255,191}

\definecolor{RdYlGn11H}{RGB}{255,255,191}

\definecolor{RdYlGn117}{RGB}{217,239,139}

\definecolor{RdYlGn11I}{RGB}{217,239,139}

\definecolor{RdYlGn118}{RGB}{166,217,106}

\definecolor{RdYlGn11J}{RGB}{166,217,106}

\definecolor{RdYlGn119}{RGB}{102,189,99}

\definecolor{RdYlGn11L}{RGB}{102,189,99}

\definecolor{RdYlGn1110}{RGB}{26,152,80}

\definecolor{RdYlGn11N}{RGB}{26,152,80}

\definecolor{RdYlGn1111}{RGB}{0,104,55}

\definecolor{RdYlGn11O}{RGB}{0,104,55}

\definecolor{Set331}{RGB}{141,211,199}

\definecolor{Set33A}{RGB}{141,211,199}

\definecolor{Set332}{RGB}{255,255,179}

\definecolor{Set33B}{RGB}{255,255,179}

\definecolor{Set333}{RGB}{190,186,218}

\definecolor{Set33C}{RGB}{190,186,218}

\definecolor{Set341}{RGB}{141,211,199}

\definecolor{Set34A}{RGB}{141,211,199}

\definecolor{Set342}{RGB}{255,255,179}

\definecolor{Set34B}{RGB}{255,255,179}

\definecolor{Set343}{RGB}{190,186,218}

\definecolor{Set34C}{RGB}{190,186,218}

\definecolor{Set344}{RGB}{251,128,114}

\definecolor{Set34D}{RGB}{251,128,114}

\definecolor{Set351}{RGB}{141,211,199}

\definecolor{Set35A}{RGB}{141,211,199}

\definecolor{Set352}{RGB}{255,255,179}

\definecolor{Set35B}{RGB}{255,255,179}

\definecolor{Set353}{RGB}{190,186,218}

\definecolor{Set35C}{RGB}{190,186,218}

\definecolor{Set354}{RGB}{251,128,114}

\definecolor{Set35D}{RGB}{251,128,114}

\definecolor{Set355}{RGB}{128,177,211}

\definecolor{Set35E}{RGB}{128,177,211}

\definecolor{Set361}{RGB}{141,211,199}

\definecolor{Set36A}{RGB}{141,211,199}

\definecolor{Set362}{RGB}{255,255,179}

\definecolor{Set36B}{RGB}{255,255,179}

\definecolor{Set363}{RGB}{190,186,218}

\definecolor{Set36C}{RGB}{190,186,218}

\definecolor{Set364}{RGB}{251,128,114}

\definecolor{Set36D}{RGB}{251,128,114}

\definecolor{Set365}{RGB}{128,177,211}

\definecolor{Set36E}{RGB}{128,177,211}

\definecolor{Set366}{RGB}{253,180,98}

\definecolor{Set36F}{RGB}{253,180,98}

\definecolor{Set371}{RGB}{141,211,199}

\definecolor{Set37A}{RGB}{141,211,199}

\definecolor{Set372}{RGB}{255,255,179}

\definecolor{Set37B}{RGB}{255,255,179}

\definecolor{Set373}{RGB}{190,186,218}

\definecolor{Set37C}{RGB}{190,186,218}

\definecolor{Set374}{RGB}{251,128,114}

\definecolor{Set37D}{RGB}{251,128,114}

\definecolor{Set375}{RGB}{128,177,211}

\definecolor{Set37E}{RGB}{128,177,211}

\definecolor{Set376}{RGB}{253,180,98}

\definecolor{Set37F}{RGB}{253,180,98}

\definecolor{Set377}{RGB}{179,222,105}

\definecolor{Set37G}{RGB}{179,222,105}

\definecolor{Set381}{RGB}{141,211,199}

\definecolor{Set38A}{RGB}{141,211,199}

\definecolor{Set382}{RGB}{255,255,179}

\definecolor{Set38B}{RGB}{255,255,179}

\definecolor{Set383}{RGB}{190,186,218}

\definecolor{Set38C}{RGB}{190,186,218}

\definecolor{Set384}{RGB}{251,128,114}

\definecolor{Set38D}{RGB}{251,128,114}

\definecolor{Set385}{RGB}{128,177,211}

\definecolor{Set38E}{RGB}{128,177,211}

\definecolor{Set386}{RGB}{253,180,98}

\definecolor{Set38F}{RGB}{253,180,98}

\definecolor{Set387}{RGB}{179,222,105}

\definecolor{Set38G}{RGB}{179,222,105}

\definecolor{Set388}{RGB}{252,205,229}

\definecolor{Set38H}{RGB}{252,205,229}

\definecolor{Set391}{RGB}{141,211,199}

\definecolor{Set39A}{RGB}{141,211,199}

\definecolor{Set392}{RGB}{255,255,179}

\definecolor{Set39B}{RGB}{255,255,179}

\definecolor{Set393}{RGB}{190,186,218}

\definecolor{Set39C}{RGB}{190,186,218}

\definecolor{Set394}{RGB}{251,128,114}

\definecolor{Set39D}{RGB}{251,128,114}

\definecolor{Set395}{RGB}{128,177,211}

\definecolor{Set39E}{RGB}{128,177,211}

\definecolor{Set396}{RGB}{253,180,98}

\definecolor{Set39F}{RGB}{253,180,98}

\definecolor{Set397}{RGB}{179,222,105}

\definecolor{Set39G}{RGB}{179,222,105}

\definecolor{Set398}{RGB}{252,205,229}

\definecolor{Set39H}{RGB}{252,205,229}

\definecolor{Set399}{RGB}{217,217,217}

\definecolor{Set39I}{RGB}{217,217,217}

\definecolor{Set3101}{RGB}{141,211,199}

\definecolor{Set310A}{RGB}{141,211,199}

\definecolor{Set3102}{RGB}{255,255,179}

\definecolor{Set310B}{RGB}{255,255,179}

\definecolor{Set3103}{RGB}{190,186,218}

\definecolor{Set310C}{RGB}{190,186,218}

\definecolor{Set3104}{RGB}{251,128,114}

\definecolor{Set310D}{RGB}{251,128,114}

\definecolor{Set3105}{RGB}{128,177,211}

\definecolor{Set310E}{RGB}{128,177,211}

\definecolor{Set3106}{RGB}{253,180,98}

\definecolor{Set310F}{RGB}{253,180,98}

\definecolor{Set3107}{RGB}{179,222,105}

\definecolor{Set310G}{RGB}{179,222,105}

\definecolor{Set3108}{RGB}{252,205,229}

\definecolor{Set310H}{RGB}{252,205,229}

\definecolor{Set3109}{RGB}{217,217,217}

\definecolor{Set310I}{RGB}{217,217,217}

\definecolor{Set31010}{RGB}{188,128,189}

\definecolor{Set310J}{RGB}{188,128,189}

\definecolor{Set3111}{RGB}{141,211,199}

\definecolor{Set311A}{RGB}{141,211,199}

\definecolor{Set3112}{RGB}{255,255,179}

\definecolor{Set311B}{RGB}{255,255,179}

\definecolor{Set3113}{RGB}{190,186,218}

\definecolor{Set311C}{RGB}{190,186,218}

\definecolor{Set3114}{RGB}{251,128,114}

\definecolor{Set311D}{RGB}{251,128,114}

\definecolor{Set3115}{RGB}{128,177,211}

\definecolor{Set311E}{RGB}{128,177,211}

\definecolor{Set3116}{RGB}{253,180,98}

\definecolor{Set311F}{RGB}{253,180,98}

\definecolor{Set3117}{RGB}{179,222,105}

\definecolor{Set311G}{RGB}{179,222,105}

\definecolor{Set3118}{RGB}{252,205,229}

\definecolor{Set311H}{RGB}{252,205,229}

\definecolor{Set3119}{RGB}{217,217,217}

\definecolor{Set311I}{RGB}{217,217,217}

\definecolor{Set31110}{RGB}{188,128,189}

\definecolor{Set311J}{RGB}{188,128,189}

\definecolor{Set31111}{RGB}{204,235,197}

\definecolor{Set311K}{RGB}{204,235,197}

\definecolor{Set3121}{RGB}{141,211,199}

\definecolor{Set312A}{RGB}{141,211,199}

\definecolor{Set3122}{RGB}{255,255,179}

\definecolor{Set312B}{RGB}{255,255,179}

\definecolor{Set3123}{RGB}{190,186,218}

\definecolor{Set312C}{RGB}{190,186,218}

\definecolor{Set3124}{RGB}{251,128,114}

\definecolor{Set312D}{RGB}{251,128,114}

\definecolor{Set3125}{RGB}{128,177,211}

\definecolor{Set312E}{RGB}{128,177,211}

\definecolor{Set3126}{RGB}{253,180,98}

\definecolor{Set312F}{RGB}{253,180,98}

\definecolor{Set3127}{RGB}{179,222,105}

\definecolor{Set312G}{RGB}{179,222,105}

\definecolor{Set3128}{RGB}{252,205,229}

\definecolor{Set312H}{RGB}{252,205,229}

\definecolor{Set3129}{RGB}{217,217,217}

\definecolor{Set312I}{RGB}{217,217,217}

\definecolor{Set31210}{RGB}{188,128,189}

\definecolor{Set312J}{RGB}{188,128,189}

\definecolor{Set31211}{RGB}{204,235,197}

\definecolor{Set312K}{RGB}{204,235,197}

\definecolor{Set31212}{RGB}{255,237,111}

\definecolor{Set312L}{RGB}{255,237,111}

\definecolor{Pastel131}{RGB}{251,180,174}

\definecolor{Pastel13A}{RGB}{251,180,174}

\definecolor{Pastel132}{RGB}{179,205,227}

\definecolor{Pastel13B}{RGB}{179,205,227}

\definecolor{Pastel133}{RGB}{204,235,197}

\definecolor{Pastel13C}{RGB}{204,235,197}

\definecolor{Pastel141}{RGB}{251,180,174}

\definecolor{Pastel14A}{RGB}{251,180,174}

\definecolor{Pastel142}{RGB}{179,205,227}

\definecolor{Pastel14B}{RGB}{179,205,227}

\definecolor{Pastel143}{RGB}{204,235,197}

\definecolor{Pastel14C}{RGB}{204,235,197}

\definecolor{Pastel144}{RGB}{222,203,228}

\definecolor{Pastel14D}{RGB}{222,203,228}

\definecolor{Pastel151}{RGB}{251,180,174}

\definecolor{Pastel15A}{RGB}{251,180,174}

\definecolor{Pastel152}{RGB}{179,205,227}

\definecolor{Pastel15B}{RGB}{179,205,227}

\definecolor{Pastel153}{RGB}{204,235,197}

\definecolor{Pastel15C}{RGB}{204,235,197}

\definecolor{Pastel154}{RGB}{222,203,228}

\definecolor{Pastel15D}{RGB}{222,203,228}

\definecolor{Pastel155}{RGB}{254,217,166}

\definecolor{Pastel15E}{RGB}{254,217,166}

\definecolor{Pastel161}{RGB}{251,180,174}

\definecolor{Pastel16A}{RGB}{251,180,174}

\definecolor{Pastel162}{RGB}{179,205,227}

\definecolor{Pastel16B}{RGB}{179,205,227}

\definecolor{Pastel163}{RGB}{204,235,197}

\definecolor{Pastel16C}{RGB}{204,235,197}

\definecolor{Pastel164}{RGB}{222,203,228}

\definecolor{Pastel16D}{RGB}{222,203,228}

\definecolor{Pastel165}{RGB}{254,217,166}

\definecolor{Pastel16E}{RGB}{254,217,166}

\definecolor{Pastel166}{RGB}{255,255,204}

\definecolor{Pastel16F}{RGB}{255,255,204}

\definecolor{Pastel171}{RGB}{251,180,174}

\definecolor{Pastel17A}{RGB}{251,180,174}

\definecolor{Pastel172}{RGB}{179,205,227}

\definecolor{Pastel17B}{RGB}{179,205,227}

\definecolor{Pastel173}{RGB}{204,235,197}

\definecolor{Pastel17C}{RGB}{204,235,197}

\definecolor{Pastel174}{RGB}{222,203,228}

\definecolor{Pastel17D}{RGB}{222,203,228}

\definecolor{Pastel175}{RGB}{254,217,166}

\definecolor{Pastel17E}{RGB}{254,217,166}

\definecolor{Pastel176}{RGB}{255,255,204}

\definecolor{Pastel17F}{RGB}{255,255,204}

\definecolor{Pastel177}{RGB}{229,216,189}

\definecolor{Pastel17G}{RGB}{229,216,189}

\definecolor{Pastel181}{RGB}{251,180,174}

\definecolor{Pastel18A}{RGB}{251,180,174}

\definecolor{Pastel182}{RGB}{179,205,227}

\definecolor{Pastel18B}{RGB}{179,205,227}

\definecolor{Pastel183}{RGB}{204,235,197}

\definecolor{Pastel18C}{RGB}{204,235,197}

\definecolor{Pastel184}{RGB}{222,203,228}

\definecolor{Pastel18D}{RGB}{222,203,228}

\definecolor{Pastel185}{RGB}{254,217,166}

\definecolor{Pastel18E}{RGB}{254,217,166}

\definecolor{Pastel186}{RGB}{255,255,204}

\definecolor{Pastel18F}{RGB}{255,255,204}

\definecolor{Pastel187}{RGB}{229,216,189}

\definecolor{Pastel18G}{RGB}{229,216,189}

\definecolor{Pastel188}{RGB}{253,218,236}

\definecolor{Pastel18H}{RGB}{253,218,236}

\definecolor{Pastel191}{RGB}{251,180,174}

\definecolor{Pastel19A}{RGB}{251,180,174}

\definecolor{Pastel192}{RGB}{179,205,227}

\definecolor{Pastel19B}{RGB}{179,205,227}

\definecolor{Pastel193}{RGB}{204,235,197}

\definecolor{Pastel19C}{RGB}{204,235,197}

\definecolor{Pastel194}{RGB}{222,203,228}

\definecolor{Pastel19D}{RGB}{222,203,228}

\definecolor{Pastel195}{RGB}{254,217,166}

\definecolor{Pastel19E}{RGB}{254,217,166}

\definecolor{Pastel196}{RGB}{255,255,204}

\definecolor{Pastel19F}{RGB}{255,255,204}

\definecolor{Pastel197}{RGB}{229,216,189}

\definecolor{Pastel19G}{RGB}{229,216,189}

\definecolor{Pastel198}{RGB}{253,218,236}

\definecolor{Pastel19H}{RGB}{253,218,236}

\definecolor{Pastel199}{RGB}{242,242,242}

\definecolor{Pastel19I}{RGB}{242,242,242}

\definecolor{Set131}{RGB}{228,26,28}

\definecolor{Set13A}{RGB}{228,26,28}

\definecolor{Set132}{RGB}{55,126,184}

\definecolor{Set13B}{RGB}{55,126,184}

\definecolor{Set133}{RGB}{77,175,74}

\definecolor{Set13C}{RGB}{77,175,74}

\definecolor{Set141}{RGB}{228,26,28}

\definecolor{Set14A}{RGB}{228,26,28}

\definecolor{Set142}{RGB}{55,126,184}

\definecolor{Set14B}{RGB}{55,126,184}

\definecolor{Set143}{RGB}{77,175,74}

\definecolor{Set14C}{RGB}{77,175,74}

\definecolor{Set144}{RGB}{152,78,163}

\definecolor{Set14D}{RGB}{152,78,163}

\definecolor{Set151}{RGB}{228,26,28}

\definecolor{Set15A}{RGB}{228,26,28}

\definecolor{Set152}{RGB}{55,126,184}

\definecolor{Set15B}{RGB}{55,126,184}

\definecolor{Set153}{RGB}{77,175,74}

\definecolor{Set15C}{RGB}{77,175,74}

\definecolor{Set154}{RGB}{152,78,163}

\definecolor{Set15D}{RGB}{152,78,163}

\definecolor{Set155}{RGB}{255,127,0}

\definecolor{Set15E}{RGB}{255,127,0}

\definecolor{Set161}{RGB}{228,26,28}

\definecolor{Set16A}{RGB}{228,26,28}

\definecolor{Set162}{RGB}{55,126,184}

\definecolor{Set16B}{RGB}{55,126,184}

\definecolor{Set163}{RGB}{77,175,74}

\definecolor{Set16C}{RGB}{77,175,74}

\definecolor{Set164}{RGB}{152,78,163}

\definecolor{Set16D}{RGB}{152,78,163}

\definecolor{Set165}{RGB}{255,127,0}

\definecolor{Set16E}{RGB}{255,127,0}

\definecolor{Set166}{RGB}{255,255,51}

\definecolor{Set16F}{RGB}{255,255,51}

\definecolor{Set171}{RGB}{228,26,28}

\definecolor{Set17A}{RGB}{228,26,28}

\definecolor{Set172}{RGB}{55,126,184}

\definecolor{Set17B}{RGB}{55,126,184}

\definecolor{Set173}{RGB}{77,175,74}

\definecolor{Set17C}{RGB}{77,175,74}

\definecolor{Set174}{RGB}{152,78,163}

\definecolor{Set17D}{RGB}{152,78,163}

\definecolor{Set175}{RGB}{255,127,0}

\definecolor{Set17E}{RGB}{255,127,0}

\definecolor{Set176}{RGB}{255,255,51}

\definecolor{Set17F}{RGB}{255,255,51}

\definecolor{Set177}{RGB}{166,86,40}

\definecolor{Set17G}{RGB}{166,86,40}

\definecolor{Set181}{RGB}{228,26,28}

\definecolor{Set18A}{RGB}{228,26,28}

\definecolor{Set182}{RGB}{55,126,184}

\definecolor{Set18B}{RGB}{55,126,184}

\definecolor{Set183}{RGB}{77,175,74}

\definecolor{Set18C}{RGB}{77,175,74}

\definecolor{Set184}{RGB}{152,78,163}

\definecolor{Set18D}{RGB}{152,78,163}

\definecolor{Set185}{RGB}{255,127,0}

\definecolor{Set18E}{RGB}{255,127,0}

\definecolor{Set186}{RGB}{255,255,51}

\definecolor{Set18F}{RGB}{255,255,51}

\definecolor{Set187}{RGB}{166,86,40}

\definecolor{Set18G}{RGB}{166,86,40}

\definecolor{Set188}{RGB}{247,129,191}

\definecolor{Set18H}{RGB}{247,129,191}

\definecolor{Set191}{RGB}{228,26,28}

\definecolor{Set19A}{RGB}{228,26,28}

\definecolor{Set192}{RGB}{55,126,184}

\definecolor{Set19B}{RGB}{55,126,184}

\definecolor{Set193}{RGB}{77,175,74}

\definecolor{Set19C}{RGB}{77,175,74}

\definecolor{Set194}{RGB}{152,78,163}

\definecolor{Set19D}{RGB}{152,78,163}

\definecolor{Set195}{RGB}{255,127,0}

\definecolor{Set19E}{RGB}{255,127,0}

\definecolor{Set196}{RGB}{255,255,51}

\definecolor{Set19F}{RGB}{255,255,51}

\definecolor{Set197}{RGB}{166,86,40}

\definecolor{Set19G}{RGB}{166,86,40}

\definecolor{Set198}{RGB}{247,129,191}

\definecolor{Set19H}{RGB}{247,129,191}

\definecolor{Set199}{RGB}{153,153,153}

\definecolor{Set19I}{RGB}{153,153,153}

\definecolor{Pastel231}{RGB}{179,226,205}

\definecolor{Pastel23A}{RGB}{179,226,205}

\definecolor{Pastel232}{RGB}{253,205,172}

\definecolor{Pastel23B}{RGB}{253,205,172}

\definecolor{Pastel233}{RGB}{203,213,232}

\definecolor{Pastel23C}{RGB}{203,213,232}

\definecolor{Pastel241}{RGB}{179,226,205}

\definecolor{Pastel24A}{RGB}{179,226,205}

\definecolor{Pastel242}{RGB}{253,205,172}

\definecolor{Pastel24B}{RGB}{253,205,172}

\definecolor{Pastel243}{RGB}{203,213,232}

\definecolor{Pastel24C}{RGB}{203,213,232}

\definecolor{Pastel244}{RGB}{244,202,228}

\definecolor{Pastel24D}{RGB}{244,202,228}

\definecolor{Pastel251}{RGB}{179,226,205}

\definecolor{Pastel25A}{RGB}{179,226,205}

\definecolor{Pastel252}{RGB}{253,205,172}

\definecolor{Pastel25B}{RGB}{253,205,172}

\definecolor{Pastel253}{RGB}{203,213,232}

\definecolor{Pastel25C}{RGB}{203,213,232}

\definecolor{Pastel254}{RGB}{244,202,228}

\definecolor{Pastel25D}{RGB}{244,202,228}

\definecolor{Pastel255}{RGB}{230,245,201}

\definecolor{Pastel25E}{RGB}{230,245,201}

\definecolor{Pastel261}{RGB}{179,226,205}

\definecolor{Pastel26A}{RGB}{179,226,205}

\definecolor{Pastel262}{RGB}{253,205,172}

\definecolor{Pastel26B}{RGB}{253,205,172}

\definecolor{Pastel263}{RGB}{203,213,232}

\definecolor{Pastel26C}{RGB}{203,213,232}

\definecolor{Pastel264}{RGB}{244,202,228}

\definecolor{Pastel26D}{RGB}{244,202,228}

\definecolor{Pastel265}{RGB}{230,245,201}

\definecolor{Pastel26E}{RGB}{230,245,201}

\definecolor{Pastel266}{RGB}{255,242,174}

\definecolor{Pastel26F}{RGB}{255,242,174}

\definecolor{Pastel271}{RGB}{179,226,205}

\definecolor{Pastel27A}{RGB}{179,226,205}

\definecolor{Pastel272}{RGB}{253,205,172}

\definecolor{Pastel27B}{RGB}{253,205,172}

\definecolor{Pastel273}{RGB}{203,213,232}

\definecolor{Pastel27C}{RGB}{203,213,232}

\definecolor{Pastel274}{RGB}{244,202,228}

\definecolor{Pastel27D}{RGB}{244,202,228}

\definecolor{Pastel275}{RGB}{230,245,201}

\definecolor{Pastel27E}{RGB}{230,245,201}

\definecolor{Pastel276}{RGB}{255,242,174}

\definecolor{Pastel27F}{RGB}{255,242,174}

\definecolor{Pastel277}{RGB}{241,226,204}

\definecolor{Pastel27G}{RGB}{241,226,204}

\definecolor{Pastel281}{RGB}{179,226,205}

\definecolor{Pastel28A}{RGB}{179,226,205}

\definecolor{Pastel282}{RGB}{253,205,172}

\definecolor{Pastel28B}{RGB}{253,205,172}

\definecolor{Pastel283}{RGB}{203,213,232}

\definecolor{Pastel28C}{RGB}{203,213,232}

\definecolor{Pastel284}{RGB}{244,202,228}

\definecolor{Pastel28D}{RGB}{244,202,228}

\definecolor{Pastel285}{RGB}{230,245,201}

\definecolor{Pastel28E}{RGB}{230,245,201}

\definecolor{Pastel286}{RGB}{255,242,174}

\definecolor{Pastel28F}{RGB}{255,242,174}

\definecolor{Pastel287}{RGB}{241,226,204}

\definecolor{Pastel28G}{RGB}{241,226,204}

\definecolor{Pastel288}{RGB}{204,204,204}

\definecolor{Pastel28H}{RGB}{204,204,204}

\definecolor{Set231}{RGB}{102,194,165}

\definecolor{Set23A}{RGB}{102,194,165}

\definecolor{Set232}{RGB}{252,141,98}

\definecolor{Set23B}{RGB}{252,141,98}

\definecolor{Set233}{RGB}{141,160,203}

\definecolor{Set23C}{RGB}{141,160,203}

\definecolor{Set241}{RGB}{102,194,165}

\definecolor{Set24A}{RGB}{102,194,165}

\definecolor{Set242}{RGB}{252,141,98}

\definecolor{Set24B}{RGB}{252,141,98}

\definecolor{Set243}{RGB}{141,160,203}

\definecolor{Set24C}{RGB}{141,160,203}

\definecolor{Set244}{RGB}{231,138,195}

\definecolor{Set24D}{RGB}{231,138,195}

\definecolor{Set251}{RGB}{102,194,165}

\definecolor{Set25A}{RGB}{102,194,165}

\definecolor{Set252}{RGB}{252,141,98}

\definecolor{Set25B}{RGB}{252,141,98}

\definecolor{Set253}{RGB}{141,160,203}

\definecolor{Set25C}{RGB}{141,160,203}

\definecolor{Set254}{RGB}{231,138,195}

\definecolor{Set25D}{RGB}{231,138,195}

\definecolor{Set255}{RGB}{166,216,84}

\definecolor{Set25E}{RGB}{166,216,84}

\definecolor{Set261}{RGB}{102,194,165}

\definecolor{Set26A}{RGB}{102,194,165}

\definecolor{Set262}{RGB}{252,141,98}

\definecolor{Set26B}{RGB}{252,141,98}

\definecolor{Set263}{RGB}{141,160,203}

\definecolor{Set26C}{RGB}{141,160,203}

\definecolor{Set264}{RGB}{231,138,195}

\definecolor{Set26D}{RGB}{231,138,195}

\definecolor{Set265}{RGB}{166,216,84}

\definecolor{Set26E}{RGB}{166,216,84}

\definecolor{Set266}{RGB}{255,217,47}

\definecolor{Set26F}{RGB}{255,217,47}

\definecolor{Set271}{RGB}{102,194,165}

\definecolor{Set27A}{RGB}{102,194,165}

\definecolor{Set272}{RGB}{252,141,98}

\definecolor{Set27B}{RGB}{252,141,98}

\definecolor{Set273}{RGB}{141,160,203}

\definecolor{Set27C}{RGB}{141,160,203}

\definecolor{Set274}{RGB}{231,138,195}

\definecolor{Set27D}{RGB}{231,138,195}

\definecolor{Set275}{RGB}{166,216,84}

\definecolor{Set27E}{RGB}{166,216,84}

\definecolor{Set276}{RGB}{255,217,47}

\definecolor{Set27F}{RGB}{255,217,47}

\definecolor{Set277}{RGB}{229,196,148}

\definecolor{Set27G}{RGB}{229,196,148}

\definecolor{Set281}{RGB}{102,194,165}

\definecolor{Set28A}{RGB}{102,194,165}

\definecolor{Set282}{RGB}{252,141,98}

\definecolor{Set28B}{RGB}{252,141,98}

\definecolor{Set283}{RGB}{141,160,203}

\definecolor{Set28C}{RGB}{141,160,203}

\definecolor{Set284}{RGB}{231,138,195}

\definecolor{Set28D}{RGB}{231,138,195}

\definecolor{Set285}{RGB}{166,216,84}

\definecolor{Set28E}{RGB}{166,216,84}

\definecolor{Set286}{RGB}{255,217,47}

\definecolor{Set28F}{RGB}{255,217,47}

\definecolor{Set287}{RGB}{229,196,148}

\definecolor{Set28G}{RGB}{229,196,148}

\definecolor{Set288}{RGB}{179,179,179}

\definecolor{Set28H}{RGB}{179,179,179}

\definecolor{Dark231}{RGB}{27,158,119}

\definecolor{Dark23A}{RGB}{27,158,119}

\definecolor{Dark232}{RGB}{217,95,2}

\definecolor{Dark23B}{RGB}{217,95,2}

\definecolor{Dark233}{RGB}{117,112,179}

\definecolor{Dark23C}{RGB}{117,112,179}

\definecolor{Dark241}{RGB}{27,158,119}

\definecolor{Dark24A}{RGB}{27,158,119}

\definecolor{Dark242}{RGB}{217,95,2}

\definecolor{Dark24B}{RGB}{217,95,2}

\definecolor{Dark243}{RGB}{117,112,179}

\definecolor{Dark24C}{RGB}{117,112,179}

\definecolor{Dark244}{RGB}{231,41,138}

\definecolor{Dark24D}{RGB}{231,41,138}

\definecolor{Dark251}{RGB}{27,158,119}

\definecolor{Dark25A}{RGB}{27,158,119}

\definecolor{Dark252}{RGB}{217,95,2}

\definecolor{Dark25B}{RGB}{217,95,2}

\definecolor{Dark253}{RGB}{117,112,179}

\definecolor{Dark25C}{RGB}{117,112,179}

\definecolor{Dark254}{RGB}{231,41,138}

\definecolor{Dark25D}{RGB}{231,41,138}

\definecolor{Dark255}{RGB}{102,166,30}

\definecolor{Dark25E}{RGB}{102,166,30}

\definecolor{Dark261}{RGB}{27,158,119}

\definecolor{Dark26A}{RGB}{27,158,119}

\definecolor{Dark262}{RGB}{217,95,2}

\definecolor{Dark26B}{RGB}{217,95,2}

\definecolor{Dark263}{RGB}{117,112,179}

\definecolor{Dark26C}{RGB}{117,112,179}

\definecolor{Dark264}{RGB}{231,41,138}

\definecolor{Dark26D}{RGB}{231,41,138}

\definecolor{Dark265}{RGB}{102,166,30}

\definecolor{Dark26E}{RGB}{102,166,30}

\definecolor{Dark266}{RGB}{230,171,2}

\definecolor{Dark26F}{RGB}{230,171,2}

\definecolor{Dark271}{RGB}{27,158,119}

\definecolor{Dark27A}{RGB}{27,158,119}

\definecolor{Dark272}{RGB}{217,95,2}

\definecolor{Dark27B}{RGB}{217,95,2}

\definecolor{Dark273}{RGB}{117,112,179}

\definecolor{Dark27C}{RGB}{117,112,179}

\definecolor{Dark274}{RGB}{231,41,138}

\definecolor{Dark27D}{RGB}{231,41,138}

\definecolor{Dark275}{RGB}{102,166,30}

\definecolor{Dark27E}{RGB}{102,166,30}

\definecolor{Dark276}{RGB}{230,171,2}

\definecolor{Dark27F}{RGB}{230,171,2}

\definecolor{Dark277}{RGB}{166,118,29}

\definecolor{Dark27G}{RGB}{166,118,29}

\definecolor{Dark281}{RGB}{27,158,119}

\definecolor{Dark28A}{RGB}{27,158,119}

\definecolor{Dark282}{RGB}{217,95,2}

\definecolor{Dark28B}{RGB}{217,95,2}

\definecolor{Dark283}{RGB}{117,112,179}

\definecolor{Dark28C}{RGB}{117,112,179}

\definecolor{Dark284}{RGB}{231,41,138}

\definecolor{Dark28D}{RGB}{231,41,138}

\definecolor{Dark285}{RGB}{102,166,30}

\definecolor{Dark28E}{RGB}{102,166,30}

\definecolor{Dark286}{RGB}{230,171,2}

\definecolor{Dark28F}{RGB}{230,171,2}

\definecolor{Dark287}{RGB}{166,118,29}

\definecolor{Dark28G}{RGB}{166,118,29}

\definecolor{Dark288}{RGB}{102,102,102}

\definecolor{Dark28H}{RGB}{102,102,102}

\definecolor{Paired31}{RGB}{166,206,227}

\definecolor{Paired3A}{RGB}{166,206,227}

\definecolor{Paired32}{RGB}{31,120,180}

\definecolor{Paired3B}{RGB}{31,120,180}

\definecolor{Paired33}{RGB}{178,223,138}

\definecolor{Paired3C}{RGB}{178,223,138}

\definecolor{Paired41}{RGB}{166,206,227}

\definecolor{Paired4A}{RGB}{166,206,227}

\definecolor{Paired42}{RGB}{31,120,180}

\definecolor{Paired4B}{RGB}{31,120,180}

\definecolor{Paired43}{RGB}{178,223,138}

\definecolor{Paired4C}{RGB}{178,223,138}

\definecolor{Paired44}{RGB}{51,160,44}

\definecolor{Paired4D}{RGB}{51,160,44}

\definecolor{Paired51}{RGB}{166,206,227}

\definecolor{Paired5A}{RGB}{166,206,227}

\definecolor{Paired52}{RGB}{31,120,180}

\definecolor{Paired5B}{RGB}{31,120,180}

\definecolor{Paired53}{RGB}{178,223,138}

\definecolor{Paired5C}{RGB}{178,223,138}

\definecolor{Paired54}{RGB}{51,160,44}

\definecolor{Paired5D}{RGB}{51,160,44}

\definecolor{Paired55}{RGB}{251,154,153}

\definecolor{Paired5E}{RGB}{251,154,153}

\definecolor{Paired61}{RGB}{166,206,227}

\definecolor{Paired6A}{RGB}{166,206,227}

\definecolor{Paired62}{RGB}{31,120,180}

\definecolor{Paired6B}{RGB}{31,120,180}

\definecolor{Paired63}{RGB}{178,223,138}

\definecolor{Paired6C}{RGB}{178,223,138}

\definecolor{Paired64}{RGB}{51,160,44}

\definecolor{Paired6D}{RGB}{51,160,44}

\definecolor{Paired65}{RGB}{251,154,153}

\definecolor{Paired6E}{RGB}{251,154,153}

\definecolor{Paired66}{RGB}{227,26,28}

\definecolor{Paired6F}{RGB}{227,26,28}

\definecolor{Paired71}{RGB}{166,206,227}

\definecolor{Paired7A}{RGB}{166,206,227}

\definecolor{Paired72}{RGB}{31,120,180}

\definecolor{Paired7B}{RGB}{31,120,180}

\definecolor{Paired73}{RGB}{178,223,138}

\definecolor{Paired7C}{RGB}{178,223,138}

\definecolor{Paired74}{RGB}{51,160,44}

\definecolor{Paired7D}{RGB}{51,160,44}

\definecolor{Paired75}{RGB}{251,154,153}

\definecolor{Paired7E}{RGB}{251,154,153}

\definecolor{Paired76}{RGB}{227,26,28}

\definecolor{Paired7F}{RGB}{227,26,28}

\definecolor{Paired77}{RGB}{253,191,111}

\definecolor{Paired7G}{RGB}{253,191,111}

\definecolor{Paired81}{RGB}{166,206,227}

\definecolor{Paired8A}{RGB}{166,206,227}

\definecolor{Paired82}{RGB}{31,120,180}

\definecolor{Paired8B}{RGB}{31,120,180}

\definecolor{Paired83}{RGB}{178,223,138}

\definecolor{Paired8C}{RGB}{178,223,138}

\definecolor{Paired84}{RGB}{51,160,44}

\definecolor{Paired8D}{RGB}{51,160,44}

\definecolor{Paired85}{RGB}{251,154,153}

\definecolor{Paired8E}{RGB}{251,154,153}

\definecolor{Paired86}{RGB}{227,26,28}

\definecolor{Paired8F}{RGB}{227,26,28}

\definecolor{Paired87}{RGB}{253,191,111}

\definecolor{Paired8G}{RGB}{253,191,111}

\definecolor{Paired88}{RGB}{255,127,0}

\definecolor{Paired8H}{RGB}{255,127,0}

\definecolor{Paired91}{RGB}{166,206,227}

\definecolor{Paired9A}{RGB}{166,206,227}

\definecolor{Paired92}{RGB}{31,120,180}

\definecolor{Paired9B}{RGB}{31,120,180}

\definecolor{Paired93}{RGB}{178,223,138}

\definecolor{Paired9C}{RGB}{178,223,138}

\definecolor{Paired94}{RGB}{51,160,44}

\definecolor{Paired9D}{RGB}{51,160,44}

\definecolor{Paired95}{RGB}{251,154,153}

\definecolor{Paired9E}{RGB}{251,154,153}

\definecolor{Paired96}{RGB}{227,26,28}

\definecolor{Paired9F}{RGB}{227,26,28}

\definecolor{Paired97}{RGB}{253,191,111}

\definecolor{Paired9G}{RGB}{253,191,111}

\definecolor{Paired98}{RGB}{255,127,0}

\definecolor{Paired9H}{RGB}{255,127,0}

\definecolor{Paired99}{RGB}{202,178,214}

\definecolor{Paired9I}{RGB}{202,178,214}

\definecolor{Paired101}{RGB}{166,206,227}

\definecolor{Paired10A}{RGB}{166,206,227}

\definecolor{Paired102}{RGB}{31,120,180}

\definecolor{Paired10B}{RGB}{31,120,180}

\definecolor{Paired103}{RGB}{178,223,138}

\definecolor{Paired10C}{RGB}{178,223,138}

\definecolor{Paired104}{RGB}{51,160,44}

\definecolor{Paired10D}{RGB}{51,160,44}

\definecolor{Paired105}{RGB}{251,154,153}

\definecolor{Paired10E}{RGB}{251,154,153}

\definecolor{Paired106}{RGB}{227,26,28}

\definecolor{Paired10F}{RGB}{227,26,28}

\definecolor{Paired107}{RGB}{253,191,111}

\definecolor{Paired10G}{RGB}{253,191,111}

\definecolor{Paired108}{RGB}{255,127,0}

\definecolor{Paired10H}{RGB}{255,127,0}

\definecolor{Paired109}{RGB}{202,178,214}

\definecolor{Paired10I}{RGB}{202,178,214}

\definecolor{Paired1010}{RGB}{106,61,154}

\definecolor{Paired10J}{RGB}{106,61,154}

\definecolor{Paired111}{RGB}{166,206,227}

\definecolor{Paired11A}{RGB}{166,206,227}

\definecolor{Paired112}{RGB}{31,120,180}

\definecolor{Paired11B}{RGB}{31,120,180}

\definecolor{Paired113}{RGB}{178,223,138}

\definecolor{Paired11C}{RGB}{178,223,138}

\definecolor{Paired114}{RGB}{51,160,44}

\definecolor{Paired11D}{RGB}{51,160,44}

\definecolor{Paired115}{RGB}{251,154,153}

\definecolor{Paired11E}{RGB}{251,154,153}

\definecolor{Paired116}{RGB}{227,26,28}

\definecolor{Paired11F}{RGB}{227,26,28}

\definecolor{Paired117}{RGB}{253,191,111}

\definecolor{Paired11G}{RGB}{253,191,111}

\definecolor{Paired118}{RGB}{255,127,0}

\definecolor{Paired11H}{RGB}{255,127,0}

\definecolor{Paired119}{RGB}{202,178,214}

\definecolor{Paired11I}{RGB}{202,178,214}

\definecolor{Paired1110}{RGB}{106,61,154}

\definecolor{Paired11J}{RGB}{106,61,154}

\definecolor{Paired1111}{RGB}{255,255,153}

\definecolor{Paired11K}{RGB}{255,255,153}

\definecolor{Paired121}{RGB}{166,206,227}

\definecolor{Paired12A}{RGB}{166,206,227}

\definecolor{Paired122}{RGB}{31,120,180}

\definecolor{Paired12B}{RGB}{31,120,180}

\definecolor{Paired123}{RGB}{178,223,138}

\definecolor{Paired12C}{RGB}{178,223,138}

\definecolor{Paired124}{RGB}{51,160,44}

\definecolor{Paired12D}{RGB}{51,160,44}

\definecolor{Paired125}{RGB}{251,154,153}

\definecolor{Paired12E}{RGB}{251,154,153}

\definecolor{Paired126}{RGB}{227,26,28}

\definecolor{Paired12F}{RGB}{227,26,28}

\definecolor{Paired127}{RGB}{253,191,111}

\definecolor{Paired12G}{RGB}{253,191,111}

\definecolor{Paired128}{RGB}{255,127,0}

\definecolor{Paired12H}{RGB}{255,127,0}

\definecolor{Paired129}{RGB}{202,178,214}

\definecolor{Paired12I}{RGB}{202,178,214}

\definecolor{Paired1210}{RGB}{106,61,154}

\definecolor{Paired12J}{RGB}{106,61,154}

\definecolor{Paired1211}{RGB}{255,255,153}

\definecolor{Paired12K}{RGB}{255,255,153}

\definecolor{Paired1212}{RGB}{177,89,40}

\definecolor{Paired12L}{RGB}{177,89,40}

\definecolor{Accent31}{RGB}{127,201,127}

\definecolor{Accent3A}{RGB}{127,201,127}

\definecolor{Accent32}{RGB}{190,174,212}

\definecolor{Accent3B}{RGB}{190,174,212}

\definecolor{Accent33}{RGB}{253,192,134}

\definecolor{Accent3C}{RGB}{253,192,134}

\definecolor{Accent41}{RGB}{127,201,127}

\definecolor{Accent4A}{RGB}{127,201,127}

\definecolor{Accent42}{RGB}{190,174,212}

\definecolor{Accent4B}{RGB}{190,174,212}

\definecolor{Accent43}{RGB}{253,192,134}

\definecolor{Accent4C}{RGB}{253,192,134}

\definecolor{Accent44}{RGB}{255,255,153}

\definecolor{Accent4D}{RGB}{255,255,153}

\definecolor{Accent51}{RGB}{127,201,127}

\definecolor{Accent5A}{RGB}{127,201,127}

\definecolor{Accent52}{RGB}{190,174,212}

\definecolor{Accent5B}{RGB}{190,174,212}

\definecolor{Accent53}{RGB}{253,192,134}

\definecolor{Accent5C}{RGB}{253,192,134}

\definecolor{Accent54}{RGB}{255,255,153}

\definecolor{Accent5D}{RGB}{255,255,153}

\definecolor{Accent55}{RGB}{56,108,176}

\definecolor{Accent5E}{RGB}{56,108,176}

\definecolor{Accent61}{RGB}{127,201,127}

\definecolor{Accent6A}{RGB}{127,201,127}

\definecolor{Accent62}{RGB}{190,174,212}

\definecolor{Accent6B}{RGB}{190,174,212}

\definecolor{Accent63}{RGB}{253,192,134}

\definecolor{Accent6C}{RGB}{253,192,134}

\definecolor{Accent64}{RGB}{255,255,153}

\definecolor{Accent6D}{RGB}{255,255,153}

\definecolor{Accent65}{RGB}{56,108,176}

\definecolor{Accent6E}{RGB}{56,108,176}

\definecolor{Accent66}{RGB}{240,2,127}

\definecolor{Accent6F}{RGB}{240,2,127}

\definecolor{Accent71}{RGB}{127,201,127}

\definecolor{Accent7A}{RGB}{127,201,127}

\definecolor{Accent72}{RGB}{190,174,212}

\definecolor{Accent7B}{RGB}{190,174,212}

\definecolor{Accent73}{RGB}{253,192,134}

\definecolor{Accent7C}{RGB}{253,192,134}

\definecolor{Accent74}{RGB}{255,255,153}

\definecolor{Accent7D}{RGB}{255,255,153}

\definecolor{Accent75}{RGB}{56,108,176}

\definecolor{Accent7E}{RGB}{56,108,176}

\definecolor{Accent76}{RGB}{240,2,127}

\definecolor{Accent7F}{RGB}{240,2,127}

\definecolor{Accent77}{RGB}{191,91,23}

\definecolor{Accent7G}{RGB}{191,91,23}

\definecolor{Accent81}{RGB}{127,201,127}

\definecolor{Accent8A}{RGB}{127,201,127}

\definecolor{Accent82}{RGB}{190,174,212}

\definecolor{Accent8B}{RGB}{190,174,212}

\definecolor{Accent83}{RGB}{253,192,134}

\definecolor{Accent8C}{RGB}{253,192,134}

\definecolor{Accent84}{RGB}{255,255,153}

\definecolor{Accent8D}{RGB}{255,255,153}

\definecolor{Accent85}{RGB}{56,108,176}

\definecolor{Accent8E}{RGB}{56,108,176}

\definecolor{Accent86}{RGB}{240,2,127}

\definecolor{Accent8F}{RGB}{240,2,127}

\definecolor{Accent87}{RGB}{191,91,23}

\definecolor{Accent8G}{RGB}{191,91,23}

\definecolor{Accent88}{RGB}{102,102,102}

\definecolor{Accent8H}{RGB}{102,102,102}

%% file: figures/Results_OdomPaths.tikz
\tikzmath{
    \wh001 = 100.8; 
    \hh001 = 18.2;
    \dswallbarrier = 2.35;
    \dfwallbarrier = 4.17;
    \dwbarrier = 0.3;
    \xcenterh001 = 65.47;
    \ycenterh001 = \hh001/2;
    \rbarrier = (\hh001-2*\dswallbarrier)/2;
    \xstartbow = \wh001 - \xcenterh001 - \dfwallbarrier - \rbarrier;
} 
\begin{tikzpicture}
    \begin{axis}[
        unit vector ratio*=1 1 1,
        xmin=-8,
        ymin=-\ycenterh001,
        xmax=15,
        ymax=5,
        legend style={at={(0.5,-0.25)},
                      anchor=north,
                      draw=none, 
                      /tikz/every even column/.append style={column sep=0.5cm}, 
                      },
        legend columns=2,
        no markers,
        xlabel={$x$\ [m]},
        ylabel={$y$\ [m]},
    ]
        \addplot[Set133, thick] table[x=x, y=y, col sep=comma, mark=none, each nth point=10]
            {Data/VinsFusion_path.csv};
        \addlegendentry{VINS-Fusion};
        \addplot[Set144, thick] table[x=x, y=y, col sep=comma, mark=none, each nth point=10]
            {Data/OpenVins_path.csv};
        \addlegendentry{OpenVINS};
        \addplot[Set131, thick] table[x=x, y=y, col sep=comma, mark=none, each nth point=10]
            {Data/Rtabmap_path.csv};
        \addlegendentry{RTAB-Map};
        \addplot[Set132, thick] table[x=x, y=y, col sep=comma, mark=none, each nth point=10]
            {Data/Racelogic_path.csv};
        \addlegendentry{Ground Truth};
            
        \begin{pgfonlayer}{background}
            \clip \pgfextra{\pgfplotspathaxisoutline};
            \draw[fill, gray!30] (-\xcenterh001,-\ycenterh001) rectangle +(\wh001, \hh001);
            
            \draw[fill, white, line width=2] (-\xcenterh001 + 34, 0) -- (\xstartbow, 0);
            \draw[dash pattern=on 24 off 48, white, line width=1] (-\xcenterh001 + 34, -\ycenterh001/2+\dswallbarrier/2) -- (\xstartbow, -\ycenterh001/2+\dswallbarrier/2);
            \draw[dash pattern=on 24 off 48, white, line width=1] (-\xcenterh001 + 34, \ycenterh001/2-\dswallbarrier/2) -- (\xstartbow, \ycenterh001/2-\dswallbarrier/2);
            
            \draw[pattern=north east lines] (-\xcenterh001, \hh001-\ycenterh001) rectangle +(16.8, -6.2);
        
            \draw[fill, gray] (-\xcenterh001, \dswallbarrier- \ycenterh001) rectangle (\xstartbow, \dswallbarrier- \ycenterh001-\dwbarrier);
            \draw[fill, gray] (-\xcenterh001 + 30, \hh001- \ycenterh001-\dswallbarrier) rectangle (\xstartbow, \hh001- \ycenterh001-\dswallbarrier+\dwbarrier);
            \draw[fill, gray] (\xstartbow, \dswallbarrier- \ycenterh001) arc (-90:90:\rbarrier)
                              --
                              ($(\xstartbow, \dswallbarrier- \ycenterh001) + (90:2*\rbarrier+\dwbarrier)$) arc (90:-90:\rbarrier+\dwbarrier)
                              --
                              cycle; 
            \draw[fill, gray] (-\xcenterh001 + 30, \hh001 - \ycenterh001 -\dswallbarrier+\dwbarrier) arc (-90:0:-2.5-\dwbarrier) -- 
                              (-\xcenterh001 + 27.5, \hh001 - \ycenterh001 -\dswallbarrier-2.5) arc (0:-90:-2.5) -- cycle;
            \draw[fill, gray] (-\xcenterh001 + 27.5, \hh001 - \ycenterh001 -\dswallbarrier-2.5) arc (0:-90:2.5+\dwbarrier) -- 
                              (-\xcenterh001 + 25-\dwbarrier, \hh001 - \ycenterh001 -\dswallbarrier-2.5-2.5) arc (-90:0:2.5) -- cycle;
        \end{pgfonlayer}
    \end{axis}
\end{tikzpicture}

%% file: figures/Result_v_omega.tikz
\begin{tikzpicture}
    \begin{axis}[
        no markers,
        grid=both,
        grid style={line width=.1pt, draw=gray!10},
        major grid style={line width=.2pt,draw=gray!50},
        minor tick num = 1,
        ylabel={$v$\ [ms$^{-1}$] / $\omega$\ [rads$^{-1}$]},
        xlabel={$t$\ [s]},
        xmin=0,
        xmax=130,
        ymax=2,
        ymin=-1,
        legend style={at={(0.5, -0.175)},
                      anchor=north,
                      draw=none, 
                      /tikz/every even column/.append style={column sep=0.5cm}, 
                      },
        legend columns=2,
    ]
    \addplot[Set133, thick] table[x=t, y=v, col sep=comma, mark=none, each nth point=20]
            {Data/rsvr_speed_rotationrate.csv};
    \addlegendentry{speed [ms$^{-1}$]}
    \addplot[Set131, thick] table[x=t, y=w, col sep=comma, mark=none, each nth point=20]
            {Data/rsvr_speed_rotationrate.csv};
    \addlegendentry{yaw rate [rads$^{-1}$]}
    \end{axis}
\end{tikzpicture}

%% file: figures/new_figures.tex
\textbf{(a)}\\
\begin{tikzpicture}
	\begin{semilogyaxis} [
	width=\linewidth,
	boxplot/draw direction = y,
	boxplot/box extend = {1/17},
	grid=major,
	xmin = 0,
	xmax = 3,
	xtick = {0, 1, 2, 3},
	x tick label as interval,
	xticklabels = { %
		{$APE_\mathrm{full}$ {[-]}},
		{$APE_\mathrm{rot}$ {[-]}},
		{$APE_\mathrm{trans}$ {[m]}}
	},
	x tick label style={
		text width=2.5cm,
		align=center
	},
	legend to name={legend_ape},
	legend style={
		draw=none,
	},
	legend columns = 3,
	name=border,
	]
	\addlegendimage{line width=1pt, densely dotted, color=black}
	\addlegendentry{Complex Urban};
	\addlegendimage{line width=1pt, densely dashed, color=black}
	\addlegendentry{Kitti};
	\addlegendimage{line width=1pt, densely dashdotted, color=black}
	\addlegendentry{EuRoC};
	\addlegendimage{empty legend};
	\addlegendentry{};
	\addlegendimage{line width=1pt, solid, color=black}
	\addlegendentry{Self-recorded};
	\addlegendimage{empty legend};
	\addlegendentry{};
	\addplot+[boxplot, densely dotted, Set133, fill=Set133!20!white, mark=*, mark options={fill=Set133},     boxplot/draw position=0.1        ] table[y index=1, col sep=comma]{Data/data_boxplot/results_kaist_vinsfusion.txt};
	\addlegendentry{VINS-Fusion}
	\addplot+[boxplot, densely dotted, Set144, fill=Set144!20!white, mark=*, mark options={fill=Set144},     boxplot/draw position=0.171428571] table[y index=1, col sep=comma]{Data/data_boxplot/results_kaist_openvins.txt};
	\addlegendentry{OpenVins}
	\addplot+[boxplot, densely dotted, Set131, fill=Set131!20!white, mark=*, mark options={fill=Set131},     boxplot/draw position=0.242857143] table[y index=1, col sep=comma]{Data/data_boxplot/results_kaist_rtabmap.txt};
	\addlegendentry{RTAB-Map}
	\addplot+[boxplot, densely dashed, Set133, fill=Set133!20!white, mark=*, mark options={fill=Set133},     boxplot/draw position=0.314285714] table[y index=1, col sep=comma]{Data/data_boxplot/results_kitti_vinsfusion.txt};
	\addplot+[boxplot, densely dashed, Set131, fill=Set131!20!white, mark=*, mark options={fill=Set131},     boxplot/draw position=0.457142857] table[y index=1, col sep=comma]{Data/data_boxplot/results_kitti_rtabmap.txt};
	\addplot+[boxplot, densely dashdotted, Set133, fill=Set133!20!white, mark=*, mark options={fill=Set133}, boxplot/draw position=0.528571429] table[y index=1, col sep=comma]{Data/data_boxplot/results_euroc_vinsfusion.txt};
	\addplot+[boxplot, densely dashdotted, Set144, fill=Set144!20!white, mark=*, mark options={fill=Set144}, boxplot/draw position=0.6        ] table[y index=1, col sep=comma]{Data/data_boxplot/results_euroc_openvins.txt};
	\addplot+[boxplot, densely dashdotted, Set131, fill=Set131!20!white, mark=*, mark options={fill=Set131}, boxplot/draw position=0.671428571] table[y index=1, col sep=comma]{Data/data_boxplot/results_euroc_rtabmap.txt};
	\addplot+[boxplot, solid, Set133, fill=Set133!20!white, mark=*, mark options={fill=Set133},              boxplot/draw position=0.742857143] table[y index=1, col sep=comma]{Data/data_boxplot/results_one2five_vinsfusion.txt};
	\addplot+[boxplot, solid, Set144, fill=Set144!20!white, mark=*, mark options={fill=Set144},              boxplot/draw position=0.814285714] table[y index=1, col sep=comma]{Data/data_boxplot/results_one2five_openvins.txt};
	\addplot+[boxplot, solid, Set131, fill=Set131!20!white, mark=*, mark options={fill=Set131},              boxplot/draw position=0.885714286] table[y index=1, col sep=comma]{Data/data_boxplot/results_one2five_rtabmap.txt};
	
	\addplot+[boxplot, dotted, Set133, fill=Set133!20!white, mark=*, mark options={fill=Set133}    , boxplot/draw position=1.1        ] table[y index=2, col sep=comma]{Data/data_boxplot/results_kaist_vinsfusion.txt};
	\addplot+[boxplot, dotted, Set144, fill=Set144!20!white, mark=*, mark options={fill=Set144}    , boxplot/draw position=1.171428571] table[y index=2, col sep=comma]{Data/data_boxplot/results_kaist_openvins.txt};
	\addplot+[boxplot, dotted, Set131, fill=Set131!20!white, mark=*, mark options={fill=Set131}    , boxplot/draw position=1.242857143] table[y index=2, col sep=comma]{Data/data_boxplot/results_kaist_rtabmap.txt};
	\addplot+[boxplot, dashed, Set133, fill=Set133!20!white, mark=*, mark options={fill=Set133}    , boxplot/draw position=1.314285714] table[y index=2, col sep=comma]{Data/data_boxplot/results_kitti_vinsfusion.txt};
	\addplot+[boxplot, dashed, Set131, fill=Set131!20!white, mark=*, mark options={fill=Set131}    , boxplot/draw position=1.457142857] table[y index=2, col sep=comma]{Data/data_boxplot/results_kitti_rtabmap.txt};
	\addplot+[boxplot, dashdotted, Set133, fill=Set133!20!white, mark=*, mark options={fill=Set133}, boxplot/draw position=1.528571429] table[y index=2, col sep=comma]{Data/data_boxplot/results_euroc_vinsfusion.txt};
	\addplot+[boxplot, dashdotted, Set144, fill=Set144!20!white, mark=*, mark options={fill=Set144}, boxplot/draw position=1.6        ] table[y index=2, col sep=comma]{Data/data_boxplot/results_euroc_openvins.txt};
	\addplot+[boxplot, dashdotted, Set131, fill=Set131!20!white, mark=*, mark options={fill=Set131}, boxplot/draw position=1.671428571] table[y index=2, col sep=comma]{Data/data_boxplot/results_euroc_rtabmap.txt};
	\addplot+[boxplot, solid, Set133, fill=Set133!20!white, mark=*, mark options={fill=Set133}     , boxplot/draw position=1.742857143] table[y index=2, col sep=comma]{Data/data_boxplot/results_one2five_vinsfusion.txt};
	\addplot+[boxplot, solid, Set144, fill=Set144!20!white, mark=*, mark options={fill=Set144}     , boxplot/draw position=1.814285714] table[y index=2, col sep=comma]{Data/data_boxplot/results_one2five_openvins.txt};
	\addplot+[boxplot, solid, Set131, fill=Set131!20!white, mark=*, mark options={fill=Set131}     , boxplot/draw position=1.885714286] table[y index=2, col sep=comma]{Data/data_boxplot/results_one2five_rtabmap.txt};
	
	\addplot+[boxplot, dotted, Set133, fill=Set133!20!white, mark=*, mark options={fill=Set133}    , boxplot/draw position=2.1        ] table[y index=3, col sep=comma]{Data/data_boxplot/results_kaist_vinsfusion.txt};
	\addplot+[boxplot, dotted, Set144, fill=Set144!20!white, mark=*, mark options={fill=Set144}    , boxplot/draw position=2.171428571] table[y index=3, col sep=comma]{Data/data_boxplot/results_kaist_openvins.txt};
	\addplot+[boxplot, dotted, Set131, fill=Set131!20!white, mark=*, mark options={fill=Set131}    , boxplot/draw position=2.242857143] table[y index=3, col sep=comma]{Data/data_boxplot/results_kaist_rtabmap.txt};
	\addplot+[boxplot, dashed, Set133, fill=Set133!20!white, mark=*, mark options={fill=Set133}    , boxplot/draw position=2.314285714] table[y index=3, col sep=comma]{Data/data_boxplot/results_kitti_vinsfusion.txt};
	\addplot+[boxplot, dashed, Set131, fill=Set131!20!white, mark=*, mark options={fill=Set131}    , boxplot/draw position=2.457142857] table[y index=3, col sep=comma]{Data/data_boxplot/results_kitti_rtabmap.txt};
	\addplot+[boxplot, dashdotted, Set133, fill=Set133!20!white,mark=*,  mark options={fill=Set133}, boxplot/draw position=2.528571429] table[y index=3, col sep=comma]{Data/data_boxplot/results_euroc_vinsfusion.txt};
	\addplot+[boxplot, dashdotted, Set144, fill=Set144!20!white, mark=*, mark options={fill=Set144}, boxplot/draw position=2.6        ] table[y index=3, col sep=comma]{Data/data_boxplot/results_euroc_openvins.txt};
	\addplot+[boxplot, dashdotted, Set131, fill=Set131!20!white, mark=*, mark options={fill=Set131}, boxplot/draw position=2.671428571] table[y index=3, col sep=comma]{Data/data_boxplot/results_euroc_rtabmap.txt};
	\addplot+[boxplot, solid, Set133, fill=Set133!20!white, mark=*, mark options={fill=Set133}     , boxplot/draw position=2.742857143] table[y index=3, col sep=comma]{Data/data_boxplot/results_one2five_vinsfusion.txt};
	\addplot+[boxplot, solid, Set144, fill=Set144!20!white, mark=*, mark options={fill=Set144}     , boxplot/draw position=2.814285714] table[y index=3, col sep=comma]{Data/data_boxplot/results_one2five_openvins.txt};
	\addplot+[boxplot, solid, Set131, fill=Set131!20!white, mark=*, mark options={fill=Set131}     , boxplot/draw position=2.885714286] table[y index=3, col sep=comma]{Data/data_boxplot/results_one2five_rtabmap.txt};
\end{semilogyaxis}
\end{tikzpicture}
\textbf{(b)}\\
\begin{tikzpicture}
\begin{semilogyaxis} [
	width=\linewidth,
	boxplot/draw direction = y,
	boxplot/draw position = {0.1 + floor(\plotnumofactualtype/12) + 1/14*mod(\plotnumofactualtype, 12)},
	boxplot/box extend = {1/17},
	grid=major,
	xmin = 0,
	xmax = 3,
	xtick = {0, 1, 2, 3, 4},
	x tick label as interval,
	xticklabels = { %
		{$RPE_\mathrm{full}$ {[-]}},
		{$RPE_\mathrm{rot}$ {[-]}},
		{$RPE_\mathrm{trans}$ {[m]}},
		{$RPE_\mathrm{normed}$}
	},
	x tick label style={
		text width=2.5cm,
		align=center
	},
	legend to name={legend1},
	legend style={
		draw=none,
	},
	legend columns = 3,
	name=border,
	]
	\addlegendimage{line width=1pt, dotted, color=black}
	\addlegendentry{Complex Urban};
	\addlegendimage{line width=1pt, dashed, color=black}
	\addlegendentry{Kitti};
	\addlegendimage{line width=1pt, dashdotted, color=black}
	\addlegendentry{EuRoC};
	\addlegendimage{empty legend};
	\addlegendentry{};
	\addlegendimage{line width=1pt, solid, color=black}
	\addlegendentry{self-recorded};
	\addlegendimage{empty legend};
	\addlegendentry{};
	\addplot+[boxplot, densely dotted, Set133, fill=Set133!20!white, mark=*, mark options={fill=Set133}    , boxplot/draw position=0.1        ] table[y index=4, col sep=comma]{Data/data_boxplot/results_kaist_vinsfusion.txt};
	\addlegendentry{VINS-Fusion}
	\addplot+[boxplot, densely dotted, Set144, fill=Set144!20!white, mark=*, mark options={fill=Set144}    , boxplot/draw position=0.171428571] table[y index=4, col sep=comma]{Data/data_boxplot/results_kaist_openvins.txt};
	\addlegendentry{OpenVins}
	\addplot+[boxplot, densely dotted, Set131, fill=Set131!20!white, mark=*, mark options={fill=Set131}    , boxplot/draw position=0.242857143] table[y index=4, col sep=comma]{Data/data_boxplot/results_kaist_rtabmap.txt};
	\addlegendentry{RTAB-Map}
	\addplot+[boxplot, densely dashed, Set133, fill=Set133!20!white, mark=*, mark options={fill=Set133}    , boxplot/draw position=0.314285714] table[y index=4, col sep=comma]{Data/data_boxplot/results_kitti_vinsfusion.txt};
	\addplot+[boxplot, densely dashed, Set131, fill=Set131!20!white, mark=*, mark options={fill=Set131}    , boxplot/draw position=0.457142857] table[y index=4, col sep=comma]{Data/data_boxplot/results_kitti_rtabmap.txt};
	\addplot+[boxplot, densely dashdotted, Set133, fill=Set133!20!white, mark=*, mark options={fill=Set133}, boxplot/draw position=0.528571429] table[y index=4, col sep=comma]{Data/data_boxplot/results_euroc_vinsfusion.txt};
	\addplot+[boxplot, densely dashdotted, Set144, fill=Set144!20!white, mark=*, mark options={fill=Set144}, boxplot/draw position=0.6        ] table[y index=4, col sep=comma]{Data/data_boxplot/results_euroc_openvins.txt};
	\addplot+[boxplot, densely dashdotted, Set131, fill=Set131!20!white, mark=*, mark options={fill=Set131}, boxplot/draw position=0.671428571] table[y index=4, col sep=comma]{Data/data_boxplot/results_euroc_rtabmap.txt};
	\addplot+[boxplot, solid, Set133, fill=Set133!20!white, mark=*, mark options={fill=Set133}             , boxplot/draw position=0.742857143] table[y index=4, col sep=comma]{Data/data_boxplot/results_one2five_vinsfusion.txt};
	\addplot+[boxplot, solid, Set144, fill=Set144!20!white,mark=*,  mark options={fill=Set144}             , boxplot/draw position=0.814285714] table[y index=4, col sep=comma]{Data/data_boxplot/results_one2five_openvins.txt};
	\addplot+[boxplot, solid, Set131, fill=Set131!20!white, mark=*, mark options={fill=Set131}             , boxplot/draw position=0.885714286] table[y index=4, col sep=comma]{Data/data_boxplot/results_one2five_rtabmap.txt};
	
	\addplot+[boxplot, densely dotted, Set133, fill=Set133!20!white, mark=*, mark options={fill=Set133}    , boxplot/draw position=1.1        ] table[y index=5, col sep=comma]{Data/data_boxplot/results_kaist_vinsfusion.txt};
	\addplot+[boxplot, densely dotted, Set144, fill=Set144!20!white, mark=*, mark options={fill=Set144}    , boxplot/draw position=1.171428571] table[y index=5, col sep=comma]{Data/data_boxplot/results_kaist_openvins.txt};
	\addplot+[boxplot, densely dotted, Set131, fill=Set131!20!white, mark=*, mark options={fill=Set131}    , boxplot/draw position=1.242857143] table[y index=5, col sep=comma]{Data/data_boxplot/results_kaist_rtabmap.txt};
	\addplot+[boxplot, densely dashed, Set133, fill=Set133!20!white, mark=*, mark options={fill=Set133}    , boxplot/draw position=1.314285714] table[y index=5, col sep=comma]{Data/data_boxplot/results_kitti_vinsfusion.txt};
	\addplot+[boxplot, densely dashed, Set131, fill=Set131!20!white, mark=*, mark options={fill=Set131}    , boxplot/draw position=1.457142857] table[y index=5, col sep=comma]{Data/data_boxplot/results_kitti_rtabmap.txt};
	\addplot+[boxplot, densely dashdotted, Set133, fill=Set133!20!white, mark=*, mark options={fill=Set133}, boxplot/draw position=1.528571429] table[y index=5, col sep=comma]{Data/data_boxplot/results_euroc_vinsfusion.txt};
	\addplot+[boxplot, densely dashdotted, Set144, fill=Set144!20!white, mark=*, mark options={fill=Set144}, boxplot/draw position=1.6        ] table[y index=5, col sep=comma]{Data/data_boxplot/results_euroc_openvins.txt};
	\addplot+[boxplot, densely dashdotted, Set131, fill=Set131!20!white, mark=*, mark options={fill=Set131}, boxplot/draw position=1.671428571] table[y index=5, col sep=comma]{Data/data_boxplot/results_euroc_rtabmap.txt};
	\addplot+[boxplot, solid, Set133, fill=Set133!20!white, mark=*, mark options={fill=Set133}             , boxplot/draw position=1.742857143] table[y index=5, col sep=comma]{Data/data_boxplot/results_one2five_vinsfusion.txt};
	\addplot+[boxplot, solid, Set144, fill=Set144!20!white, mark=*, mark options={fill=Set144}             , boxplot/draw position=1.814285714] table[y index=5, col sep=comma]{Data/data_boxplot/results_one2five_openvins.txt};
	\addplot+[boxplot, solid, Set131, fill=Set131!20!white, mark=*, mark options={fill=Set131}             , boxplot/draw position=1.885714286] table[y index=5, col sep=comma]{Data/data_boxplot/results_one2five_rtabmap.txt};
	
	\addplot+[boxplot, densely dotted, Set133, fill=Set133!20!white, mark=*, mark options={fill=Set133}    , boxplot/draw position=2.1        ] table[y index=6, col sep=comma]{Data/data_boxplot/results_kaist_vinsfusion.txt};
	\addplot+[boxplot, densely dotted, Set144, fill=Set144!20!white, mark=*, mark options={fill=Set144}    , boxplot/draw position=2.171428571] table[y index=6, col sep=comma]{Data/data_boxplot/results_kaist_openvins.txt};
	\addplot+[boxplot, densely dotted, Set131, fill=Set131!20!white, mark=*, mark options={fill=Set131}    , boxplot/draw position=2.242857143] table[y index=6, col sep=comma]{Data/data_boxplot/results_kaist_rtabmap.txt};
	\addplot+[boxplot, densely dashed, Set133, fill=Set133!20!white, mark=*, mark options={fill=Set133}    , boxplot/draw position=2.314285714] table[y index=6, col sep=comma]{Data/data_boxplot/results_kitti_vinsfusion.txt};
	\addplot+[boxplot, densely dashed, Set131, fill=Set131!20!white, mark=*, mark options={fill=Set131}    , boxplot/draw position=2.457142857] table[y index=6, col sep=comma]{Data/data_boxplot/results_kitti_rtabmap.txt};
	\addplot+[boxplot, densely dashdotted, Set133, fill=Set133!20!white, mark=*, mark options={fill=Set133}, boxplot/draw position=2.528571429] table[y index=6, col sep=comma]{Data/data_boxplot/results_euroc_vinsfusion.txt};
	\addplot+[boxplot, densely dashdotted, Set144, fill=Set144!20!white, mark=*, mark options={fill=Set144}, boxplot/draw position=2.6        ] table[y index=6, col sep=comma]{Data/data_boxplot/results_euroc_openvins.txt};
	\addplot+[boxplot, densely dashdotted, Set131, fill=Set131!20!white, mark=*, mark options={fill=Set131}, boxplot/draw position=2.671428571] table[y index=6, col sep=comma]{Data/data_boxplot/results_euroc_rtabmap.txt};
	\addplot+[boxplot, solid, Set133, fill=Set133!20!white, mark=*, mark options={fill=Set133}             , boxplot/draw position=2.742857143] table[y index=6, col sep=comma]{Data/data_boxplot/results_one2five_vinsfusion.txt};
	\addplot+[boxplot, solid, Set144, fill=Set144!20!white, mark=*, mark options={fill=Set144}             , boxplot/draw position=2.814285714] table[y index=6, col sep=comma]{Data/data_boxplot/results_one2five_openvins.txt};
	\addplot+[boxplot, solid, Set131, fill=Set131!20!white, mark=*, mark options={fill=Set131}             , boxplot/draw position=2.885714286] table[y index=6, col sep=comma]{Data/data_boxplot/results_one2five_rtabmap.txt};
\end{semilogyaxis}
\node[below] at ($(border.south) + (0,-1.2)$) {\pgfplotslegendfromname{legend1}};
\end{tikzpicture}